%% file: manuscript.tex
\documentclass{article}

\usepackage{arxiv}

\usepackage[toc,page]{appendix}

\usepackage{color,soul}
\usepackage{comment}
\usepackage{amsmath}
\usepackage{amssymb}
\usepackage{natbib}
\usepackage{dblfloatfix}  
\setcitestyle{square,comma,numbers,sort}
\usepackage{float}
\usepackage[font={footnotesize},labelfont={footnotesize}]{caption}
\usepackage[font={footnotesize},labelfont={footnotesize}]{subcaption}
\usepackage{tikz,pgfplots,adjustbox}
\tikzset{>=latex}
\pgfplotsset{compat=1.3}
\usetikzlibrary{shapes.geometric, shapes.symbols, fit, calc, arrows, backgrounds}
\usetikzlibrary{patterns}
\usetikzlibrary{pgfplots.groupplots}
\usepackage{bbold}
\usepgfplotslibrary{fillbetween}

\usepackage[utf8]{inputenc} 
\usepackage[T1]{fontenc}    
\usepackage{hyperref}       
\usepackage{url}            
\usepackage{booktabs}       
\usepackage{amsfonts}       
\usepackage{nicefrac}       
\usepackage{microtype}      
\usepackage{lipsum}
\usepackage{graphicx}
\graphicspath{ {./images/} }
\newmuskip\pFqmuskip
\newcommand*\pFq[6][8]{%
  \begingroup 
  \pFqmuskip=#1mu\relax
  \mathchardef\normalcomma=\mathcode`,
  \mathcode`\,=\string"8000
  \begingroup\lccode`\~=`\,
  \lowercase{\endgroup\let~}\pFqcomma
  {}_{#2}F_{#3}{\left[\genfrac..{0pt}{}{#4}{#5};#6\right]}%
  \endgroup
}
\newcommand{\pFqcomma}{{\normalcomma}\mskip\pFqmuskip}

\newmuskip\pFqmuskip
\newcommand*\pFqtilde[6][8]{%
  \begingroup 
  \pFqmuskip=#1mu\relax
  \mathchardef\normalcomma=\mathcode`,
  \mathcode`\,=\string"8000
  \begingroup\lccode`\~=`\,
  \lowercase{\endgroup\let~}\pFqcomma
  {}_{#2}\tilde{F}_{#3}{\left[\genfrac..{0pt}{}{#4}{#5};#6\right]}%
  \endgroup
}

\newmuskip\pFqmuskip
\newcommand*\pFqhat[6][8]{%
  \begingroup 
  \pFqmuskip=#1mu\relax
  \mathchardef\normalcomma=\mathcode`,
  \mathcode`\,=\string"8000
  \begingroup\lccode`\~=`\,
  \lowercase{\endgroup\let~}\pFqcomma
  {}_{#2}\hat{F}_{#3}{\left[\genfrac..{0pt}{}{#4}{#5};#6\right]}%
  \endgroup
}

\usepackage{times,amsmath,epsfig,cite,bm,float,epstopdf,graphicx,graphics,amsthm}
\usepackage{relsize}
\usepackage{dblfloatfix}
\usepackage{balance}
\usepackage[stable]{footmisc}
\usepackage{lettrine}
\usepackage{algorithm,algpseudocode}
\usepackage{tabularx,multirow,booktabs}
\usepackage{rotating,afterpage,lscape}

\usepackage{color,soul}

\usepackage{enumitem}
\usepackage{afterpage}
\usepackage{chronology}
\usepackage{blindtext}

\newcolumntype{C}{>{\centering\arraybackslash}X} 
\usepackage[font={footnotesize},labelfont={footnotesize}]{caption}
\usepackage[font={footnotesize},labelfont={footnotesize}]{subcaption}
\usepackage{tikz,pgfplots,adjustbox}
\tikzset{>=latex}
\pgfplotsset{compat=newest}
\usetikzlibrary{patterns}
\usetikzlibrary{pgfplots.groupplots}
\usetikzlibrary{plotmarks}
\usetikzlibrary{spy}

\pgfplotscreateplotcyclelist{colors_qda}{%
smooth,blue!80!black,mark=*,mark size=1.6pt,semithick\\%
smooth,red,mark=asterisk,every mark/.append style={solid,fill=red!80!black},semithick\\%
smooth,green!60!black,mark=square*,mark size=1.2pt,semithick\\%
smooth,orange,mark=diamond*,semithick\\%
smooth,black,mark=pentagon*,mark size=1.7pt,semithick\\%
smooth,magenta!80!white,mark=triangle*,semithick\\%
}
\pgfplotscreateplotcyclelist{colors_obtl}{%
smooth,blue!80!black,solid,semithick\\%
smooth,red!80!black,densely dashdotted,semithick\\%
smooth,green!60!black,densely dashed,semithick\\%
smooth,orange,densely dashdotdotted,semithick\\%
smooth,black,densely dotted, thick\\%
smooth,black,densely dotted, thick\\%
}

\pgfplotscreateplotcyclelist{colors_scz}{%
green!60!black,densely dashdotted,  thick\\%
red!80!black,densely dotted,  thick\\%
 blue!80!black ,densely dashed,  thick\\%
smooth,orange,densely dashdotdotted,semithick\\%
smooth,black,densely dotted,  thick\\%
smooth,black,densely dotted, thick\\%
}
\title{Robust Importance Sampling for Error Estimation in the Context of Optimal Bayesian Transfer Learning}

\author{
 Omar Maddouri \\
  Department of Electrical and Computer Engineering\\
  Texas A\&M University\\
  College Station, TX 77843, USA \\
  \texttt{omar.maddouri@tamu.edu} \\
   \And
 Xiaoning Qian \\
  Department of Electrical and Computer Engineering\\
  Texas A\&M University\\
  College Station, TX 77843, USA \\
  \texttt{xqian@ece.tamu.edu} \\
  \And
  Francis~J.~Alexander \\
  Brookhaven National Laboratory\\
  Upton, NY 11973, USA \\
  \texttt{falexander@bnl.gov} \\
  \And
  Edward~R.~Dougherty \\
  Department of Electrical and Computer Engineering\\
  Texas A\&M University\\
  College Station, TX 77843, USA \\
  \texttt{edward@ece.tamu.edu} \\
  \And
 Byung-Jun Yoon \\
  Department of Electrical and Computer Engineering\\
  Texas A\&M University\\
  College Station, TX 77843, USA \\
  \texttt{bjyoon@ece.tamu.edu} \\
}

\begin{document}
\maketitle
\begin{abstract}
 Classification has been a major task for building intelligent systems as it enables decision-making under uncertainty. Classifier design aims at building models from training data for representing feature-label distributions--either explicitly or implicitly. In many scientific or clinical settings, training data are typically limited, which makes designing accurate classifiers and evaluating their classification error extremely challenging. 
While transfer learning (TL) can alleviate this issue by incorporating data from relevant source domains to improve learning in a different target domain, it has received little attention for performance assessment, notably in error estimation. In this paper, we fill this gap by investigating knowledge transferability in the context of classification error estimation within a Bayesian paradigm. We introduce a novel class of Bayesian minimum mean-square error (MMSE) estimators for optimal Bayesian transfer learning (OBTL), which enables rigorous evaluation of classification error under uncertainty in a small-sample setting. Using Monte Carlo importance sampling, we employ the proposed estimator to evaluate the classification accuracy of a broad family of classifiers that span diverse learning capabilities. Experimental results based on both synthetic data as well as real-world RNA sequencing (RNA-seq) data show that our proposed OBTL error estimation scheme clearly outperforms standard error estimators, especially in a small-sample setting, by tapping into the data from other relevant domains.
\end{abstract}


\section{Introduction}
Transfer learning provides  promising means to repurpose the data and/or scientific knowledge available in other relevant domains for new applications in a given domain. The ability to transfer relevant data/knowledge across different domains practically enables learning effective models in target domains with limited data. Classifier design can take advantage of transfer learning (TL) to address small-sample challenges we often face in various scientific applications. However, rigorous error estimators that can leverage such transferred data/knowledge for better estimation of classification error have been missing to date, which makes the design framework epistemologically incomplete~\citep{Dougherty2006}.
Generally, the scientific validity of any predictive model is assessed by the ability to generalize outside the observed training sample. However, the available sample is often too small in many scientific applications (e.g., bio-marker discovery) to hold out sufficient data just for testing purpose, which makes the reuse of training data for both classifier design and error estimation inevitable.
While various error estimation schemes exist to date, their accuracy and reliability in a small-sample setting are often questioned~\citep{Diamandis2010}. For instance, in~\citep{Dalton2011_A} many classification studies of cancer gene expression data have been listed where the performance was assessed by cross-validation (CV) based on small-size training datasets. Analyses in~\citep{Neto2004_A} have shown that CV error estimators derived based on small-size samples show large variance, which explains the controversy across many biological studies that relied on data-driven CV~\citep{Song2016}. Model-based error estimation also faces practical challenges as non-informative modeling assumptions may mislead the error estimators in case of model mismatch.

The ability for accurate error estimation based on small samples is also critical in other contexts, an example being continual learning~\citep{Schlimmer1986}, where a series of labeled datasets are sequentially fed to the learner as in realistic learning scenarios.
In recent years, continual learning regained attention as a promising strategy for avoiding ``catastrophic forgetting'' that may arise when the training data are split for a series of small learning operations called tasks~\citep{Goodfellow2013}. Such a continual learning setting is getting prevalent these days, where retaining the observed training data is either undesirable (confidentiality) or intractable (high-throughput systems), and developing reliable task-specific error estimators is indispensable. For instance, an intuitive approach to continual learning from a Bayesian perspective is to leverage the posterior of the current task to update the prior of the next task~\citep{farquhar_unifying_2018}. However, analysis in~\citep{farquhar_towards_2018} has shown that evaluation approaches for this prior-focused set-up suffer from severe bias in realistic scenarios, particularly for finely partitioned data. Recent work in~\citep{Gossmann2021} provided a solution for test data scarcity by reusing the same test set in the context of a continuously evolving classification problem. To avoid overfitting the test data, the authors employed a reusable holdout mechanism based on the AUC (area under the receiver operating characteristic curve) metric. Nevertheless, this approach remains contingent on the availability of an independent test set.
For these reasons, there is a pressing need to develop novel error estimators that can effectively overcome data-scarcity limitations. For assessing different classification models in the context of small-size training datasets, having an accurate error estimator with TL capabilities that can take advantage of relevant datasets in other domains would be highly beneficial. Such an estimator would be readily applicable to continual learning as cross-task datasets can be seen as related source-target samples. In the remainder of this introductory section, we provide a brief review of the standard error estimation techniques along with prevalent transfer learning scenarios.

\subsection{Relevant prior studies}

\subsubsection{Error Estimation}

For unknown feature-label distributions, the classification error of a given classifier is typically estimated by leveraging a large sample collected from the true distribution. However, limiting factors, such as the excessive cost of large-scale data acquisition, make it often infeasible to collect and hold out large test sets. Consequently, the available small-size sample may have to be used for both training and assessing the classifier, and researchers have striven to devise practical methods for accurate error estimation.

Existing error estimation schemes can be broadly categorized into parametric and non-parametric methods. Non-parametric estimators compute the error rate by counting the misclassified points, where widely used estimators include the resubstitution, cross-validation (CV), and bootstrap estimators. Resubstitution assesses the error committed on the training data, as a result of which it has an optimistic bias and frequently raises overfitting issues~\citep{Neto2009}.
In $k$-fold CV~\citep{Lunts1967,Stone1974}, the data is randomly split into $k$-folds and the classification error is estimated by the average error of the $k$ distinct classifiers, each trained on $k$-1 folds and evaluated on the left-out fold. Leave-one-out CV (LOO) is a non-randomized special case of CV where the number of folds $k$ is equal to the sample size. Unlike resubstitution, CV and LOO tend to have small bias but they suffer from large variance~\citep{Neto2004_A}.
The bootstrap estimator~\citep{Efron1979,Efron1983} takes a different randomized approach. Instead of partitioning the data into $k$-folds, bootstrap generates $k$ sets of equal size by sampling (with replacement) from the original set.
In this sampling approach, bootstrap sets are formed on average by 63.2\% of unique data points from the original training data. This randomization, while reducing the variance, results in a pessimistic bias as it excludes parts of the data from model training. To reconcile this deviation, the 0.632-bootstrap uses a weighted average of the standard bootstrap and the optimistically biased resubstitution with 63.2\% of the approximation being a bootstrap estimate.
Bolstered error estimation~\citep{Neto2004_B} employs density kernels centered at validation data points to smoothly approximate the true error. Instead of counting the misclassified points, bolstered estimator quantifies the overflow of bolstering kernels through the classifier decision boundaries. This technique, when associated with optimal kernel variances, has shown to result in a significant improvement over resubstitution and CV in terms of variance and bias.

Parametric methods include the popular plug-in estimator that naively estimates the true error from an empirical model. One major limitation for this technique is the strong dependence on the estimated parameters, which may fail drastically due to poor parameter estimation.
The Bayesian minimum mean square error estimator (BEE) proposed in~\citep{Dalton2011_A, Dalton2011_B} significantly enhances the robustness by computing the expected true error with respect to the posterior of the model parameters. The BEE has shown notable improvements over standard estimators as it effectively handles the uncertainty about the feature-label distributions~\citep{Dalton2011_A, Dalton2011_B}.

\subsubsection{Transfer Learning}

Transfer Learning (TL) was originally proposed to provide remedies for pitfalls caused by training-data scarcity in a target domain by utilizing available data from different yet relevant source domains~\citep{Pan2010}. Based on the properties of source and target domains, two scenarios of TL are available. The first one, commonly known as ``homogeneous TL'', occurs when the source and target domains share the same feature space. The second scenario is called ``heterogeneous TL'' and is considered when differences exist between domains in terms of feature space or data dimensionality.

Practically, TL is applied across domains that share identical category structures, where the sample and label spaces remain unchanged while only the probability distributions differ across domains. This specific case of TL is also known as domain adaptation (DA)~\citep{Patel2015,Csurka2017}. In order to maximize the similarity between feature-label distributions in the source and target domains, the vast majority of existent DA methods proceed by adapting three domain aspects: the data, the model parameters, or a cross-domain latent space.
For data adaptation, methods such as instance re-weighting~\citep{Jiang2007} and transfer adaptive boosting~\citep{Dai2007} are mostly suited for homogeneous TL as they act by assigning optimized weights to source and target samples to build an augmented dataset. In~\citep{Karbalayghareh2018}, a Bayesian framework was proposed for TL, in which the source and target domains are related through the joint prior density of the model parameters.
For parameter adaptation, popular algorithms redesign classifier models built in source domains to make them effective in target domains. This is enabled by leveraging the available target data to update the classifier parameters~\citep{Hoffman2013,Duan2009,Bruzzone2010}.
Another approach for TL is to project the source and target data into an intermediate invariant space that unifies the two domains~\citep{Herath2017}. This approach showed promising results in heterogeneous TL~\citep{Duan2012}, taking advantage of deep learning models that can map complex feature spaces in the respective domains into an invariant latent domain~\citep{Long2015,Liu2016}.

\subsection{Main contributions of this paper}

In this paper, we propose a novel transfer learning framework for robust estimation of classification error based on a rigorous Bayesian paradigm. To the best of our knowledge, this study is the first work on TL-based BEE, which can significantly enhance our understanding of transferability across domains in the context of error estimation. Building on the Bayesian transfer learning framework proposed in~\citep{Karbalayghareh2018}, we introduce a novel TL-based BEE estimator that can enhance the error estimation accuracy in the target domain by utilizing the data available in a relevant source domain. We present a rigorous study of error estimation in the context of Bayesian TL and show that our proposed TL-based BEE effectively represents and exploits the relatedness (or dependency) between different domains to improve error estimates in a challenging small-sample setting.
For applicability of the proposed TL-based BEE estimator in real-world problems for arbitrary classifiers, we introduce an efficient and robust importance sampling set-up with control variates where the importance density and the control variates function are carefully defined to reduce the variance of the estimator while keeping the overall sampling process computationally feasible and scalable. For this purpose, we utilize Laplace approximations for fast evaluation of matrix-variate confluent and Gauss hypergeometric functions.
The performance of the TL-based BEE estimator is extensively evaluated using both synthetic datasets as well as real-world biological datasets. A wide range of classifiers with different levels of learning capabilities are considered, in order to demonstrate the general applicability of our TL-based BEE estimation scheme. We also show the outstanding performance of the proposed estimator with respect to standard error estimation techniques that are commonly used.

The remainder of this paper is organized as follows. In Sec.~\ref{sec:tl} we present the Bayesian transfer learning validation framework. Section~\ref{sec:bee} defines the novel TL-based Bayesian MMSE estimator and derives its expression for arbitrary classifiers. In Sec.~\ref{sec:sampling}, details of the robust importance sampling set-up is provided with the definition of the control variates. Section~\ref{sec:experiment} elaborates the experimental set-up for both synthetic and real-world datasets. The performance analysis results are presented in Sec.~\ref{sec:result} and we conclude the paper in Sec.~\ref{sec:conclusion} with potential future research directions.
Appendix A reviews the binary classification setting. Appendix B states the definition of the standard Bayesian MMSE estimator. Appendix C recalls the definition of matrix-variate hypergeometric functions and provides the Laplace approximations of confluent and Gauss hypergeometric functions of matrix argument. Appendix D states some useful results for importance sampling and provides the derivation of the control variates function. Finally, Appendix E provides additional results for linear classifiers.


\section{Bayesian Transfer Learning Framework for Binary Classification}
\label{sec:tl}

We consider a binary classification problem in the context of supervised transfer learning where there are two common classes in each domain.
Let $\mathcal{D}_{s}$ and $\mathcal{D}_{t}$ be two labeled datasets from the source and target domains with sizes $N_{s}$ and $N_{t}$, respectively.
We are interested in the scenario where $N_t \ll N_{s}$.
Let $\mathcal{D}_{s}^{y}=\left\{ \mathbf{x}_{s,1}^{y}, \mathbf{x}_{s,2}^{y}, \cdots, \mathbf{x}_{s,n_{s}}^{y}\right \}$, $y \in \left \{0,1 \right \}$, where $n_{s}^{y}$ denotes the size of source data in class $y$.
Likewise, let $\mathcal{D}_{t}^{y}=\left\{ \mathbf{x}_{t,1}^{y}, \mathbf{x}_{t,2}^{y}, \cdots, \mathbf{x}_{t,n_{t}}^{y}\right \}$, $y \in \left \{0,1 \right \}$, where $n_{t}^{y}$ denotes the size of target data in class $y$.
We consider a $d$-dimensional homogeneous transfer learning scenario where $\mathcal{D}_{s}$ and $\mathcal{D}_{t}$ are normally distributed and separately sampled from the source and target domains, respectively.
\begin{equation}
    \mathbf{x}_{z}^{y}  \sim \mathcal{N}\left ( \mathbf{\mu}_{z}^{y},\left( \mathbf{\Lambda}_{z}^{y}\right)^{-1}\right ),~y \in \left\{0,1\right\},
\end{equation}
where $z\in\left\{s,t\right\}$, $\mathbf{\mu}_{z}^{y}$ is a $\left(d\times 1\right)$ mean vector in domain $z$ for class $y$, and $\mathbf{\Lambda}_{z}^{y}$ is a $\left(d \times d\right)$ matrix that denotes the precision matrix (inverse of covariance) in domain $z$ for label $y$.
An augmented feature vector $\mathbf{x}^{y}=\begin{bmatrix}\mathbf{x}_{t}^{y}\\\mathbf{x}_{s}^{y}\end{bmatrix}$ is a joint sample point from two related source and target domains given by
\begin{equation}
    \mathbf{x}^{y}\sim \mathcal{N}\left ( \mathbf{\mu}^{y},\left( \mathbf{\Lambda}^{y}\right)^{-1}\right ),~y \in \left\{0,1\right\},
\end{equation}
with
\begin{equation}
    \mathbf{\mu}^{y}=\begin{bmatrix}\mathbf{\mu}_{t}^{y}\\\mathbf{\mu}_{s}^{y}\end{bmatrix}, \mathbf{\Lambda}^{y}=\begin{bmatrix}
\mathbf{\Lambda}_{t}^{y} & \mathbf{\Lambda}_{ts}^{y}\\ 
{\mathbf{\Lambda}_{ts}^{y}}^{T} & \mathbf{\Lambda}_{s}^{y}
\end{bmatrix}
\end{equation}
where $X^{T}$ denotes the transpose of matrix $X$.
This sampling is enabled through a joint prior distribution for $\mathbf{\Lambda}_{s}^{y}$ and $\mathbf{\Lambda}_{t}^{y}$ that marginalizes out the off-diagonal block matrix $\mathbf{\Lambda}_{ts}^{y}$.
Using a Gaussian-Wishart distribution as the joint prior for mean and precision matrices, the joint model factorizes as
\begin{equation}\label{joint_prior}
    p\left(\mathbf{\mu}_{s}^{y}, \mathbf{\mu}_{t}^{y}, \mathbf{\Lambda}_{s}^{y}, \mathbf{\Lambda}_{t}^{y}\right)=p\left(\mathbf{\mu}_{s}^{y}, \mathbf{\mu}_{t}^{y}|\mathbf{\Lambda}_{s}^{y}, \mathbf{\Lambda}_{t}^{y}\right)~p\left(\mathbf{\Lambda}_{s}^{y}, \mathbf{\Lambda}_{t}^{y}\right).
\end{equation}
For conditionally independent mean vectors given the covariances, the joint prior in~\eqref{joint_prior} further expands to
\begin{equation}
    p\left(\mathbf{\mu}_{s}^{y}, \mathbf{\mu}_{t}^{y}, \mathbf{\Lambda}_{s}^{y}, \mathbf{\Lambda}_{t}^{y}\right)=p\left(\mathbf{\mu}_{s}^{y}|\mathbf{\Lambda}_{s}^{y}\right)~p\left(\mathbf{\mu}_{t}^{y}|\mathbf{\Lambda}_{t}^{y}\right)~p\left(\mathbf{\Lambda}_{s}^{y}, \mathbf{\Lambda}_{t}^{y}\right). 
\end{equation}
The block diagonal precision matrices $\mathbf{\Lambda}_{z}^{y}$ for $z \in \left \{t,s \right \}$ are obtained after sampling $\mathbf{\Lambda}^{y}$ from a predefined joint Wishart distribution as defined in~\citep{Karbalayghareh2018} such that
$\mathbf{\Lambda}^{y} \sim W_{2d}\left(\mathbf{M}^{y},\nu^{y} \right)$, where $\nu^{y}$ is a hyperparameter for the degrees of freedom that satisfies $\nu^{y} \geq 2d$ and $\mathbf{M}^{y}$ is a $\left(2d \times 2d\right)$ positive definite scale matrix of the form $\mathbf{M}^{y}=\begin{pmatrix}
\mathbf{M}_{t}^{y} & \mathbf{M}_{ts}^{y}\\ 
{\mathbf{M}_{ts}^{y}}^{T} & \mathbf{M}_{s}^{y}
\end{pmatrix}$.
$\mathbf{M}_{t}^{y}$ and $\mathbf{M}_{s}^{y}$ are also positive definite scale matrices and $\mathbf{M}_{ts}$ denotes the off-diagonal component that models the interaction between source and target domains.
Given $\mathbf{\Lambda}_{z}^{y}$, and assuming normally distributed mean vectors we get
\begin{equation}
    \mathbf{\mu}_{z}^{y}\sim \mathcal{N}\left ( \mathbf{m}_{z}^{y},\left( \kappa_{z}^{y}~\mathbf{\Lambda}_{z}^{y}\right)^{-1}\right ),~z \in \left\{s,t\right\} \mbox{~and~} y \in \left\{0,1\right\},
\end{equation}
where $\mathbf{m}_{z}^{y}$ is the $\left(d \times 1\right)$ mean vector of the mean parameter $\mathbf{\mu}_{z}^{y}$ and $\kappa_{z}^{y}$ is a positive scalar hyperparameter.
The joint prior distribution $p\left(\mathbf{\Lambda}_{s}^{y},\mathbf{\Lambda}_{t}^{y}\right)$ as derived in~\citep{Karbalayghareh2018} acts like a bridge through which the useful knowledge transfers from the source to the target domain, making the posterior of the target parameters of the underlying feature-label distribution distributed more narrowly  around the true values.


\section{Bayesian MMSE Estimation via Transfer Learning}
\label{sec:bee}

In this section we propose a novel class of Bayesian MMSE error estimators for transfer learning where the observed sample is a mixture of source and target data. The basic  classification setting and a brief review of the standard BEE estimator are presented in the Appendices A and B.

Rooted in signal estimation, the BEE has been motivated by optimal filtering for functions of random variables~\citep{Dalton2011_A}. For a function of two random variables $g\left(X,Y\right)$, the optimal estimator $\hat{g}\left(Y\right)$ of a filter $g\left(Y\right)$ after observing only $Y$ in the mean-square sense is given by
\begin{equation}
    \hat{g}\left(Y\right)=E_{X}\left[g\left(X,Y\right)|Y\right].
\end{equation}
Replacing $X$ with the parameter vector $\theta$ of the feature-label distribution and $Y$ by the sample $S_{n}$ (of size $n$), leads to the standard BEE that has been introduced in~\citep{Dalton2011_A} as
\begin{equation}
    \hat{\varepsilon}\left (S_{n} \right )=E_{\mathbf{\theta}}\left[\varepsilon_{n}\left( \mathbf{\theta},S_{n}~|~S_{n}\right) \right].
\end{equation}
In TL, the sample $S_{n}$ is a mixture of source and target data such that $S_{n}=\left(\mathcal{D}_{s}\cup \mathcal{D}_{t}\right)_{n}$ with $n=N_{s}+N_{t}$ and the classifier $\psi_{n}$ is designed either on $\mathcal{D}_{t}$, $\mathcal{D}_{s}$, or $\mathcal{D}_{s}\cup \mathcal{D}_{t}$. This requires a close attention as the TL-based BEE is valid only for fixed classifiers given the sample. This assumption carries limitations. For instance, classifiers that are only fixed given $\mathcal{D}_{t}$ but not $\mathcal{D}_{s}$ are not deterministic for every set of parameters estimated based on $\mathcal{D}_{s}\cup \mathcal{D}_{t}$. We introduce in this paper the novel TL-based BEE for TL defined as
\begin{equation}\label{BEE_TL_long}
    \hat{\varepsilon}\left (\left(\mathcal{D}_{s}\cup \mathcal{D}_{t}\right)_{n} \right )=E_{\mathbf{\theta}}\left[\varepsilon_{n}\left( \mathbf{\theta},\left(\mathcal{D}_{s}\cup \mathcal{D}_{t}\right)_{n}\right)~|~\left(\mathcal{D}_{s}\cup \mathcal{D}_{t}\right)_{n} \right],
\end{equation}
where $\theta=\left[\theta_{t}, \theta_{s}\right]$ denotes the parameter vector of the joint model formed by the target parameters $\theta_{t}$ and source parameters $\theta_{s}$. For a fixed classifier given $\left(\mathcal{D}_{s}\cup \mathcal{D}_{t}\right)_{n}$, this estimator is optimal on average in the mean-square sense and unbiased when averaged over all parameters and samples.
For classification in the target domain, the posterior density $\pi^{*}\left(\theta\right)$ reduces to the posterior of the target parameters after observing the target and source data and takes the form
\begin{equation}
    \pi^{*}\left(\theta_{t}\right)=\pi^{*}\left(\theta_{t}~|~\mathcal{D}_{s}, \mathcal{D}_{t}\right),
\end{equation}
where $\pi^{*}\left(\theta_{t}~|~\mathcal{D}_{s}, \mathcal{D}_{t}\right)$ is obtained by marginalizing out the source domain parameters.
Ultimately, the BEE for TL takes the form
\begin{align}
    \begin{split}
        \hat{\varepsilon}\left (\left(\mathcal{D}_{s}\cup \mathcal{D}_{t}\right)_{n} \right )&=E_{\mathbf{\theta}_{t}}\left[\varepsilon_{n}\left( \mathbf{\theta}_{t},\left(\mathcal{D}_{s}\cup \mathcal{D}_{t}\right)_{n}\right)~|~\left(\mathcal{D}_{s}\cup \mathcal{D}_{t}\right)_{n} \right]\\
        &=E_{\pi^{*}\left(\mathbf{\theta}_{t}\right)}\left[\varepsilon_{n}\left( \mathbf{\theta}_{t},\left(\mathcal{D}_{s}\cup \mathcal{D}_{t}\right)_{n}\right)\right].
    \end{split}
\end{align}
For the sake of simplicity we write
\begin{equation}
    \hat{\varepsilon}=E_{\pi^{*}}\left[\varepsilon_{n}\right],
\end{equation}
where $\pi^{*}=\pi^{*}\left(\mathbf{\theta}_{t}~|~\mathcal{D}_{t},\mathcal{D}_{s}\right)$ denotes the posterior of the target parameters after observing the hybrid sample $\mathcal{D}_{t} \cup \mathcal{D}_{s}$.

\subsection{Bayesian transfer learning for error estimation}

The advantage of the mathematical formulation that underlies the proposed TL-based BEE (and also the original TL Bayesian framework in~\citep{Karbalayghareh2018}) is that it articulates a unified Bayesian inference model that assumes a specified prior distribution governing the parameter vector $\theta_{t}$ and acting like a bridge to help update $\pi^{*}\left(\theta_{t}\right)$ after observing $\mathcal{D}_{t}$ and $\mathcal{D}_{s}$. From this standpoint, the derivation of the TL-based BEE for TL depends on determining $\pi^{*}\left(\theta_{t}\right)$.
To determine the TL-based BEE in the context of the presented Bayesian transfer learning framework we evoke the following theorem.

\emph{Theorem 1~\citep{Karbalayghareh2018}:} Given the target $\mathcal{D}_{t}$ and source $\mathcal{D}_{s}$ data, the posterior distribution of target mean $\mu_{t}^{y}$ and the target precision matrix $\mathbf{\Lambda}_{t}^{y}$ for the classes $y\in\left\{0,1\right\}$ has Gaussian-hypergeometric function distribution given by
\begin{align}
    \begin{split}
        &p\left(\mu_{t}^{y},\mathbf{\Lambda}_{t}^{y}~|~\mathcal{D}_{t}^{y},\mathcal{D}_{s}^{y}\right)=A^{y}\left|\mathbf{\Lambda}_{t}^{y}\right|^{\frac{1}{2}}\\
        &\times~\mbox{exp}\left(-\frac{\kappa_{t,n}^{y}}{2}\left(\mu_{t}^{y}-\mathbf{m}_{t,n}^{y}\right)^{T}\mathbf{\Lambda}_{t}^{y}\left(\mu_{t}^{y}-\mathbf{m}_{t,n}^{y}\right)\right)\\
        &\times~\left|\mathbf{\Lambda}_{t}^{y}\right|^{\frac{\nu^{y}+n_{t}^{y}-d-1}{2}}~\mbox{etr}\left(-\frac{1}{2}\left(\mathbf{T}_{t}^{y}\right)^{-1}\mathbf{\Lambda}_{t}^{y}\right)\\
        &\times~\pFq{1}{1}{\frac{\nu^{y}+n_{s}^{y}}{2}}{\frac{\nu^{y}}{2}}{\frac{1}{2}~\mathbf{F}^{y}~\mathbf{\Lambda}_{t}^{y}~{\mathbf{F}^{y}}^{T}~\mathbf{T}_{s}^{y}},
    \end{split}
\end{align}
where $A^{y}$ is a constant of proportionality given by
\begin{align}
    \begin{split}
        \left(A^{y}\right)^{-1}&=\left(\frac{2\pi}{\kappa_{t,n}^{y}}\right)^{\frac{d}{2}}2^{\frac{d\left(\nu^{y}+n_{t}^{y}\right)}{2}}\Gamma_{d}\left(\frac{\left(\nu^{y}+n_{t}^{y}\right)}{2}\right)\left|\mathbf{T}_{t}^{y}\right|^{\frac{\left(\nu^{y}+n_{t}^{y}\right)}{2}}\\
        &\times~\pFq{2}{1}{\frac{\nu^{y}+n_{s}^{y}}{2}, \frac{\nu^{y}+n_{t}^{y}}{2}}{\frac{\nu^{y}}{2}}{\mathbf{T}_{s}^{y}~\mathbf{F}^{y}~\mathbf{T}_{t}^{y}~{\mathbf{F}^{y}}^{T}},
    \end{split}
\end{align}
and
\begin{align}
    \begin{split}
        \kappa_{t,n}^{y}&=\kappa_{t}^{y}+n_{t}^{y},\\
        \mathbf{m}_{t,n}^{y}&=\frac{\kappa_{t}^{y}~\mathbf{m}_{t}^{y}+n_{t}^{y}~\overline{\mathbf{x}}_{t}^{y}}{\kappa_{t}^{y}+n_{t}^{y}},\\
        \left(\mathbf{T}_{t}^{y}\right)^{-1}&=\left(\mathbf{M}_{t}^{y}\right)^{-1}+{\mathbf{F}^{y}}^{T}~\mathbf{C}^{y}~\mathbf{F}^{y}+\mathbf{S}_{t}^{y}\\
        &+\frac{\kappa_{t}^{y}~n_{t}^{y}}{\kappa_{t}^{y}+n_{t}^{y}}\left(\mathbf{m}_{t}^{y}-\overline{\mathbf{x}}_{t}^{y}\right)\left(\mathbf{m}_{t}^{y}-\overline{\mathbf{x}}_{t}^{y}\right)^{T},\\
        \left(\mathbf{T}_{s}^{y}\right)^{-1}&=\left(\mathbf{C}^{y}\right)^{-1}+\mathbf{S}_{s}^{y}\\
        &+\frac{\kappa_{s}^{y}~n_{s}^{y}}{\kappa_{s}^{y}+n_{s}^{y}}\left(\mathbf{m}_{s}^{y}-\overline{\mathbf{x}}_{s}^{y}\right)\left(\mathbf{m}_{s}^{y}-\overline{\mathbf{x}}_{s}^{y}\right)^{T},
    \end{split}
\end{align}
with sample means and covariances for $z\in\left\{s,t\right\}$ given by
\begin{align*}
    \begin{split}
    \overline{\mathbf{x}}_{z}^{y}&=\frac{1}{n_{z}^{y}}\sum_{i=1}^{n_{z}^{y}}\mathbf{x}_{z,i}^{y}\\
    \mathbf{S}_{z}^{y}&=\sum_{i=1}^{n_{z}^{y}}\left(\mathbf{x}_{z,i}^{y}-\overline{\mathbf{x}}_{z}^{y}\right)\left(\mathbf{x}_{z,i}^{y}-\overline{\mathbf{x}}_{z}^{y}\right)^{T}.
    \end{split}
\end{align*}
$\pFq{1}{1}{a}{b}{\mathbf{X}}$ and $\pFq{2}{1}{a,b}{c}{\mathbf{X}}$ are respectively the confluent and Gauss matrix-variate hypergeometric functions reviewed in Appendix C.
Now, using Theorem~1 and assuming that $c$, $\theta_{t}^{0}$, and $\theta_{t}^{1}$ are independent prior to observing $\mathcal{D}_{t}$ and $\mathcal{D}_{s}$, the BEE for TL is given by
\begin{equation}
    \hat{\varepsilon}=E_{\pi^{*}}\left[c \right]~E_{\pi^{*}}\left[\varepsilon_{n}^{0} \right]+\left(1-E_{\pi^{*}}\left[c \right]\right)~E_{\pi^{*}}\left[\varepsilon_{n}^{1} \right],
\end{equation}
where
\begin{equation}\label{BEE_final}
    E_{\pi^{*}}\left[\varepsilon_{n}^{y}\right]=\int_{\Theta_{t}^{y}}\varepsilon_{n}^{y}\left(\theta_{t}^{y}\right)~\pi^{*}\left(\theta_{t}^{y}\right)~d\theta_{t}^{y}
\end{equation}
with $\Theta_{t}^{y}$ being the parameter space that contains all possible values for $\theta_{t}^{y}$.

\subsection{Computing TL-based BEE for arbitrary classifiers}

Computing the TL-based BEE for an arbitrary classifier $\psi_{n}$ involves the evaluation of the integral in~\eqref{BEE_final}. 
Even when we have an analytic expression for the true error of the studied classifier, the closed-form expression for the TL-based BEE cannot be easily derived due to the complex expression of the target posterior in the presence of the matrix-variate hypergeometric functions. With non-linear classifiers, this becomes practically impossible as no closed-form expression exists for the true error itself. The standard way to approximate the true error in this case is to consider the test error. For a specified parameter $\theta_{t}$, a large test set is generated from $f_{\theta_{t}}\left(\mathbf{x},y\right)$ and the performance of $\psi_{n}$ is evaluated on that test set. This requires sampling from $\pi^{*}\left(\theta_{t}^{y}\right)$ so that the integral in~\eqref{BEE_final} can be approximated by a finite sum. Suppose we have $N$ posterior sample points $\theta_{t,i}^{y}\sim \pi^{*}\left(\theta_{t}^{y}\right),~i=1\cdots N$. Then the approximation is given by
\begin{equation}
    E_{\pi^{*}}\left[\varepsilon_{n}^{y}\right]\approx\frac{1}{N}\sum_{i=1}^{N}\varepsilon_{n}^{y}\left(\theta_{t,i}^{y}\right).
\end{equation}
Because of the generalized confluent and Gauss hypergeometric functions in the expression of $\pi^{*}$, sampling directly from the posterior is very laborious and the computational cost of applying Markov Chain Monte Carlo (MCMC) methods is exorbitant as the execution may take several weeks even on high performance computing clusters. To address this issue, we propose in the next section an efficient self-normalized importance sampling set-up with control variates that provides accurate estimates for the TL-based BEE and significantly reduces the computation time to make the proposed TL-based BEE feasible.

\section{Self-normalized Importance Sampling with Control Variates}
\label{sec:sampling}

\subsection{Importance sampling}

Importance sampling (IS) is a variance reduction technique that provides a remedy to sampling from complex distributions~\citep{Robert2004}. To estimate $E_{\pi^{*}}\left[\varepsilon_{n}^{y}\right]$, IS makes a multiplicative adjustment to $\varepsilon_{n}^{y}$ to compensate for sampling from an alternative importance distribution $\Phi^{*}$ instead of $\pi^{*}$. If $\Phi^{*}$ is a positive probability density function on $\Theta_{t}^{y}$, we can write
\begin{align}\label{importance_expectation}
    \begin{split}
        E_{\pi^{*}}\left[\varepsilon_{n}^{y}\right]&=\int_{\Theta_{t}^{y}}\varepsilon_{n}^{y}\left(\theta_{t}^{y}\right)~\pi^{*}\left(\theta_{t}^{y}\right)~d\theta_{t}^{y}\\
        &=\int_{\Theta_{t}^{y}}\frac{\varepsilon_{n}^{y}\left(\theta_{t}^{y}\right)~\pi^{*}\left(\theta_{t}^{y}\right)}{\Phi^{*}\left(\theta_{t}^{y}\right)}~\Phi^{*}\left(\theta_{t}^{y}\right)~d\theta_{t}^{y}\\
        &=E_{\Phi^{*}}\left[\frac{\varepsilon_{n}^{y}\left(\theta_{t}^{y}\right)~\pi^{*}\left(\theta_{t}^{y}\right)}{\Phi^{*}\left(\theta_{t}^{y}\right)}\right].
    \end{split}
\end{align}
Achieving an accurate IS estimation is contingent on selecting an appropriate importance density that is nearly proportional to $\varepsilon_{n}^{y}\left(\theta_{t}^{y}\right)~\pi^{*}\left(\theta_{t}^{y}\right)$. By analogy to~\citep{Gordon1993,Ackerberg2001}, a plausible and cogent candidate for $\Phi^{*}$ emanates as the posterior of target parameters upon observation of target-only data. Obviously, both distributions are tracking the same model parameters in the target domain upon observation of data. To determine $\Phi^{*}\left(\theta_{t}^{y}\right)= p\left(\mu_{t}^{y},\mathbf{\Lambda}_{t}^{y}~|~\mathcal{D}_{t}^{y}\right)$ we require the following lemma:

\emph{Lemma 1~\citep{Muirhead2009}:} If $\mathcal{D}=\left\{\mathbf{x}_{1},\cdots,\mathbf{x}_{n}\right\}$ where $\mathbf{x}_{i}$ is a $d\times 1$ vector and $\mathbf{x}_{i}\sim\mathcal{N}\left(\mu, \left(\mathbf{\Lambda}\right)^{-1}\right)$, for $i=1,\cdots,n$, and $\left(\mu, \mathbf{\Lambda}\right)$ has a Gaussian-Wishart prior, such that $\mu|\mathbf{\Lambda}\sim \mathcal{N}\left(\mathbf{m},\left(\kappa\mathbf{\Lambda}\right)^{-1}\right)$ and $\mathbf{\Lambda\sim W_{d}\left(\mathbf{M},\nu\right)}$, then the posterior of $\left(\mu,\mathbf{\Lambda}\right)$ upon observing $\mathcal{D}$ is also a Gaussian-Wishart distribution such that
\begin{align}
    \begin{split}
        \mu|\mathbf{\Lambda},\mathcal{D}&\sim\mathcal{N}\left(\mathbf{m}_{n},\left(\kappa_{n}\mathbf{\Lambda}\right)^{-1}\right);\,\,
        \mathbf{\Lambda}|\mathcal{D}\sim W_{d}\left(\mathbf{M}_{n},\nu_{n}\right),
    \end{split}
\end{align}
where
\begin{align}
    \begin{split}
        \kappa_{n}&=\kappa+n, \quad
        \nu_{n}=\nu+n,\\
        \mathbf{m}_{n}&=\frac{\kappa \mathbf{m}+n\overline{\mathbf{x}}}{\kappa+n}, \mbox{ and} \\
        \mathbf{M}_{n}^{-1}&= \mathbf{M}^{-1}+\mathbf{S}+\frac{\kappa n}{\kappa+n}\left(\mathbf{m}-\overline{\mathbf{x}}\right)\left(\mathbf{m}-\overline{\mathbf{x}}\right)^{T},
    \end{split}
\end{align}
depending on the sample mean and covariance matrix
\begin{align}
    \begin{split}
        \overline{\mathbf{x}}&=\frac{1}{n}\sum_{i=1}^{n}\mathbf{x}_{i}, \quad
        \mathbf{S}=\sum_{i=1}^{n}\left(\mathbf{x}_{i}-\overline{\mathbf{x}}\right)\left(\mathbf{x}_{i}-\overline{\mathbf{x}}\right)^{T}.
    \end{split}
\end{align}

Using \emph{Lemma 1} we now get the expression of the importance density $\Phi^{*}$ given by
\begin{align}
    \begin{split}
        &p\left(\mu_{t}^{y},\mathbf{\Lambda}_{t}^{y}~|~\mathcal{D}_{t}^{y}\right)=\left(\frac{2\pi}{\kappa_{t,n}^{y}}\right)^{-\frac{d}{2}}2^{-\frac{d\left(\nu^{y}+n_{t}^{y}\right)}{2}}\Gamma_{d}^{-1}\left(\frac{\left(\nu^{y}+n_{t}^{y}\right)}{2}\right),\\
        &\times~\left|\mathbf{M}_{t,n}^{y}\right|^{-\frac{\left(\nu^{y}+n_{t}^{y}\right)}{2}}\left|\mathbf{\Lambda}_{t}^{y}\right|^{\frac{1}{2}}\\
        &\times~\mbox{exp}\left(-\frac{\kappa_{t,n}^{y}}{2}\left(\mu_{t}^{y}-\mathbf{m}_{t,n}^{y}\right)^{T}\mathbf{\Lambda}_{t}^{y}\left(\mu_{t}^{y}-\mathbf{m}_{t,n}^{y}\right)\right)\\
        &\times~\left|\mathbf{\Lambda}_{t}^{y}\right|^{\frac{\nu^{y}+n_{t}^{y}-d-1}{2}}~\mbox{etr}\left(-\frac{1}{2}\left(\mathbf{M}_{t,n}^{y}\right)^{-1}\mathbf{\Lambda}_{t}^{y}\right),
    \end{split}
\end{align}
where
\begin{align}\label{empirical_estimates}
    \begin{split}
        \kappa_{t,n}^{y}&=\kappa_{t}^{y}+n_{t}^{y},\\
        \mathbf{m}_{t,n}^{y}&=\frac{\kappa_{t}^{y}~\mathbf{m}_{t}^{y}+n_{t}^{y}~\overline{\mathbf{x}}_{t}^{y}}{\kappa_{t}^{y}+n_{t}^{y}},\\
        \left(\mathbf{M}_{t,n}^{y}\right)^{-1}&=\left(\mathbf{M}_{t}^{y}\right)^{-1}+\mathbf{S}_{t}^{y}\\
        &+\frac{\kappa_{t}^{y}~n_{t}^{y}}{\kappa_{t}^{y}+n_{t}^{y}}\left(\mathbf{m}_{t}^{y}-\overline{\mathbf{x}}_{t}^{y}\right)\left(\mathbf{m}_{t}^{y}-\overline{\mathbf{x}}_{t}^{y}\right)^{T},
    \end{split}
\end{align}
with sample mean and covariance given by
\begin{align*}
    \begin{split}
    \overline{\mathbf{x}}_{t}^{y}&=\frac{1}{n_{t}^{y}}\sum_{i=1}^{n_{t}^{y}}\mathbf{x}_{t,i}^{y}\\
    \mathbf{S}_{t}^{y}&=\sum_{i=1}^{n_{t}^{y}}\left(\mathbf{x}_{t,i}^{y}-\overline{\mathbf{x}}_{t}^{y}\right)\left(\mathbf{x}_{t,i}^{y}-\overline{\mathbf{x}}_{t}^{y}\right)^{T}.
    \end{split}
\end{align*}
After simplifications, the expression of the TL-based BEE in~\eqref{importance_expectation} takes the form
\begin{equation}
    E_{\pi^{*}}\left[\varepsilon_{n}^{y}\right]=E_{\Phi^{*}}\left[\varepsilon_{n}^{y}\left(\theta_{t}^{y}\right)\mathcal{L}\left(\theta_{t}^{y}\right)\right],
\end{equation}
where $\theta_{t}^{y}=\left(\mu_{t}^{y}, \mathbf{\Lambda_{t}^{y}}\right)$ and $\mathcal{L}\left(\theta_{t}^{y}\right)$ is the likelihood ratio given by
\begin{align}
    \begin{split}
        &\mathcal{L}\left(\mu_{t}^{y}, \mathbf{\Lambda_{t}^{y}}\right)=\mbox{etr}\left(-\frac{1}{2}\left[\left(\mathbf{T}_{t}^{y}\right)^{-1}-\left(\mathbf{M}_{t,n}^{y}\right)^{-1} \right ]\mathbf{\Lambda}_{t}^{y}\right)\\
        &\times~\left|\frac{\mathbf{M}_{t,n}^{y}}{\mathbf{T}_{t}^{y}}\right|^{\frac{\left(\nu^{y}+n_{t}^{y}\right)}{2}}\frac{\pFq{1}{1}{\frac{\nu^{y}+n_{s}^{y}}{2}}{\frac{\nu^{y}}{2}}{\frac{1}{2}~\mathbf{F}^{y}~\mathbf{\Lambda}_{t}^{y}~{\mathbf{F}^{y}}^{T}~\mathbf{T}_{s}^{y}}}{\pFq{2}{1}{\frac{\nu^{y}+n_{s}^{y}}{2}, \frac{\nu^{y}+n_{t}^{y}}{2}}{\frac{\nu^{y}}{2}}{\mathbf{T}_{s}^{y}~\mathbf{F}^{y}~\mathbf{T}_{t}^{y}~{\mathbf{F}^{y}}^{T}}}.
    \end{split}
\end{align}
Although the likelihood ratio has a simplified expression, computing the hypergeometric functions involves the computation of series of zonal polynomials, which is computationally expensive and not scalable to high dimensions. To mitigate this limitation, we use the Laplace
approximations of these functions (see Appendix C). To rectify possible disproportionalities in likelihood ratios due to approximations, we consider the self-normalized importance sampling estimate given by
\begin{equation}
    \widehat{E}_{\Phi^{*}}\left[\varepsilon_{n}^{y}\right]\approx\frac{\sum_{i=1}^{N}\varepsilon_{n}^{y}\left(\theta_{t,i}^{y}\right)~\mathcal{L}\left(\theta_{t,i}^{y}\right)}{\sum_{i=1}^{N}~\mathcal{L}\left(\theta_{t,i}^{y}\right)}
\end{equation}
with $\theta_{t,i}^{y}\sim \Phi^{*}\left(\theta_{t}^{y}\right),~i=1\cdots N$.


\subsection{Control variates}

For more stable and efficient estimates, we further combine importance sampling with control variates.
Using control variates in conjunction with importance sampling is a variance reduction technique, in particular when a significant portion of a model for estimating the expectation can be solved explicitly.
In our case, a useful control variates function (CVF) $\mathcal{V}\left(\theta_{t}^{y}\right)$ satisfies
\begin{align}
\begin{split}
    E_{\Phi^{*}}\left[\mathcal{V}\left(\theta_{t}^{y}\right)\right]&=\int_{\Theta_{t}^{y}}\mathcal{V}\left(\theta_{t}^{y}\right)\Phi^{*}\left(\theta_{t}^{y}\right)~d\theta_{t}^{y}=\delta,
\end{split}
\end{align}
where $\delta$ is a constant. Under such circumstances, a more stable estimate for the TL-based BEE can be derived as
\begin{equation}
    \widetilde{E}_{\Phi^{*}}\left[\varepsilon_{n}^{y}\right]\approx\frac{\sum_{i=1}^{N}\varepsilon_{n}^{y}\left(\theta_{t,i}^{y}\right)~\mathcal{L}\left(\theta_{t,i}^{y}\right)}{\sum_{i=1}^{N}~\mathcal{L}\left(\theta_{t,i}^{y}\right)}-\frac{1}{N}\sum_{i=1}^{N}\frac{\beta\mathcal{V}\left(\theta_{t,i}^{y}\right)}{\Phi^{*}\left(\theta_{t,i}^{y}\right)}+\beta\delta,
\end{equation}
where $\theta_{t,i}^{y}\sim \Phi^{*}\left(\theta_{t}^{y}\right),~i=1\cdots N$ and $\beta$ is a weighting coefficient tuned to reduce the variance of the estimate. The optimal value of $\beta$ is given by
\begin{equation}
    \beta_{opt}=\frac{\mbox{cov}\left[\zeta_{n}^{y}\left(\theta_{t}^{y}\right),\mathcal{V}\left(\theta_{t}^{y}\right)\right]}{\mbox{var}\left[\mathcal{V}\left(\theta_{t}^{y}\right)\right]},
\end{equation}
with
\begin{equation}
    \zeta_{n}^{y}\left(\theta_{t}^{y}\right)=\frac{\varepsilon_{n}^{y}\left(\theta_{t}^{y}\right)~\mathcal{L}\left(\theta_{t}^{y}\right)}{\frac{1}{N}\sum_{i=1}^{N}~\mathcal{L}\left(\theta_{t,i}^{y}\right)}
\end{equation}
and $\mbox{cov}\left[\cdot, \cdot\right]$ and $\mbox{var}\left[\cdot\right]$ denote covariance and variance, respectively (see Appendix D for more details). In practice, it is not likely that we know $\beta_{opt}$ beforehand, but it is estimated from the Monte Carlo sample. It turns out that $\widetilde{E}_{\Phi^{*}}$ has lower variance than $\widehat{E}_{\Phi^{*}}$ by a factor of $\left(1-\mbox{corr}\left[\zeta_{n}^{y}\left(\theta_{t}^{y}\right),\mathcal{V}\left(\theta_{t}^{y}\right)\right]\right)$, where $\mbox{corr}\left[\mathbf{a}, \mathbf{b}\right]$ denotes the correlation coefficient between $\mathbf{a}$ and $\mathbf{b}$ and given by
\begin{equation}
    \mbox{corr}\left[\mathbf{a}, \mathbf{b}\right]=\frac{\mbox{cov}\left[\mathbf{a}, \mathbf{b}\right]}{\sqrt{\mbox{var}\left[\mathbf{a}\right]}~\sqrt{\mbox{var}\left[\mathbf{b}\right]}}.
\end{equation}
To select an appropriate CVF we need to consider two criteria. First, its expectation with respect to $\Phi^{*}$ should have an exact evaluation. Second, it has to be correlated with the estimated error. A favorable candidate is the analytic true error of linear classifiers. We consider in this study a CVF given by the true error of an LDA classifier defined by
$g_{N_{t}}\left(\mathbf{x}\right)=\mathbf{a}_{N_{t}}^{T}\mathbf{x}+b_{N_{t}}$ where $\mathbf{a}_{N_{t}}=\mathbf{S}_{t}^{-1}\left(\overline{\mathbf{x}}_{t}^{1}-\overline{\mathbf{x}}_{t}^{0}\right)$, $b_{N_{t}}=-\frac{1}{2}\mathbf{a}^{T}\left(\overline{\mathbf{x}}_{t}^{1}+\overline{\mathbf{x}}_{t}^{0}\right)+\ln{\frac{n_{t}^{1}}{n_{t}^{0}}}$ and the pooled covariance $\mathbf{S}_{t}$ is given by 
\begin{equation}
    \mathbf{S}_{t}=\frac{\left(n_{t}^{0}-1\right)\mathbf{S}_{t}^{0}+\left(n_{t}^{1}-1\right)\mathbf{S}_{t}^{1}}{N_{t}-2}.
\end{equation}
$\overline{\mathbf{x}}_{t}^{y}$ and $\mathbf{S}_{t}^{y}$ are the empirical estimates utilized in~\eqref{empirical_estimates}. Thus, the CVF is given by
\begin{equation}
    \mathcal{V}\left(\mu_{t}^{y}, \mathbf{\Lambda}_{t}^{y}\right)=\mathbf{\Phi}\left(\frac{\left(-1\right)^{y}g_{N_{t}}\left(\mu_{t}^{y}\right)}{\sqrt{\mathbf{a}_{N_{t}}^{T}~\left(\mathbf{\Lambda}_{t}^{y}\right)^{-1}~\mathbf{a}_{N_{t}}}}\right)
\end{equation}
with $\mathbf{\Phi}$ denoting the standard normal Gaussian cumulative distribution function (CDF).
Now it remains only to determine $E_{\Phi^{*}}\left[\mathcal{V}\left(\mu_{t}^{y}, \mathbf{\Lambda}_{t}^{y}\right)\right]$ in closed-form to fully define the estimation set-up. We can show after simplifications and using results from~\citep{Dalton2011_B} that
\small
\begin{align}
    \begin{split}
        &E_{\Phi^{*}}\left[\mathcal{V}\left(\mu_{t}^{y}, \mathbf{\Lambda}_{t}^{y}\right)\right]=\frac{1}{2}\\
        &+\frac{\mbox{sgn}\left(A\right)}{2}\mathcal{I}\left(\frac{A^{2}}{A^{2}+{\mathbf{a}_{N_{t}}}^{T}\left[\mathbf{M}_{t,n}^{y}\right]^{-1}\mathbf{a}_{N_{t}}};\frac{1}{2},\frac{\nu^{y}+n_{t}^{y}-d+1}{2}\right),
    \end{split}
\end{align}
\normalsize
where $\mbox{sgn}\left(\cdot\right)$ is the sign function,
\begin{equation}
    A=\left(-1\right)^{y}g_{N_{t}}\left(\mathbf{m}_{t,n}^{y}\right)\sqrt{\frac{\kappa_{t,n}^{y}}{1+\kappa_{t,n}^{y}}},
\end{equation}
and $\mathcal{I}\left(\cdot;\cdot,\cdot\right)$ denotes the regularized incomplete beta function given by
\begin{equation}
    \mathcal{I}\left(x;a,b\right)=\frac{\Gamma\left(a+b\right)}{\Gamma\left(a\right)\Gamma\left(b\right)}\int_{0}^{x}t^{a-1}\left(1-t\right)^{b-1}dt,
\end{equation}
with $\Gamma\left(\cdot\right)$ being the regular univariate gamma function. Details for simplifying $E_{\Phi^{*}}\left[\mathcal{V}\left(\mu_{t}^{y}, \mathbf{\Lambda}_{t}^{y}\right)\right]$ are covered in Appendix D.

The complete specification of the CVF concludes our importance sampling set-up. We enumerate some advantages of the proposed set-up over direct sampling methods. First, the importance density $\Phi^{*}$ is much simpler than the nominal density $\pi^{*}$ that involves matrix-variate hypergeometric functions. Second, our set-up successfully combines two variance reduction techniques that enable accurate estimation. Last, and most importantly, the independence of the generated Monte Carlo samples w.r.t source data permits the reuse of the sampled parameters with various source datasets for fixed models. This reusability significantly reduces the computational cost of sampling from $\Phi^{*}$ and makes the utilization of advanced MCMC methods amenable as the whole process could be accelerated by a factor of 10$\sim$20, that also scales up with the dimensionality and the number of used source datasets. For efficient sampling from $\Phi^{*}$, we use Hamiltonian Monte Carlo (HMC), proven to have a superior performance to standard MCMC samplers~\citep{Carpenter2017}. For this purpose, we utilize the STAN software that offers a full Bayesian statistical inference framework with HMC~\citep{Carpenter2017}.

 \section{Experiments and Datasets}
 \label{sec:experiment}
 
 To evaluate the performance of the proposed error estimator, we consider the mean-square error (MSE) as a performance measure to understand the joint behaviour of the classification error $\varepsilon_{n}$ and its estimate $\hat{\varepsilon}$. For the random vector $\left(\varepsilon_{n}, \hat{\varepsilon}\right)$, the MSE is defined as
 \begin{equation}
     \mbox{MSE}\left(\hat{\varepsilon}\right)=\mbox{E}\left[\left|\hat{\varepsilon}-\varepsilon_{n}\right|^{2}\right].
 \end{equation}
 In what follows,  we present the details of the experimental set-up for demonstrating the performance of the proposed TL-based BEE based on three different types of classifiers applied to both synthetic data as well as real biological datasets.
 
 \subsection{Synthetic datasets}
 
 To simulate and verify the extent of knowledge transferability across domains, we consider a wide range of joint prior densities that model the different levels of relatedness between the source and target domains. The proposed set-up is as follows. We consider a binary classification problem in the context of homogeneous TL with dimensions 2, 3, and 5. In the simulated datasets, the number of source data points per class varies between 10 and 500 and between 5 and 50 for target datasets. This mimics realistic settings of small-size sample conditions (especially, in the target domain) as reported in the literature~\citep{Dalton2011_A}. We set up the data distributions as follows. $\nu=\nu^{y}=d+20$, $\kappa_{t}=\kappa_{t}^{y}=100$, $\kappa_{s}=\kappa_{s}^{y}=100$, $\mathbf{m}_{t}^{0}=\mathbf{0}_{d}$, $\mathbf{m}_{t}^{1}=\vartheta \times\mathbf{1}_{d}$, $\mathbf{m}_{s}^{0}=\mathbf{m}_{t}^{0}+10\times\mathbf{1}_{d}$, $\mathbf{m}_{s}^{1}=\mathbf{m}_{t}^{1}+10\times\mathbf{1}_{d}$, where $\vartheta$ is an adjustable scalar used to control the Bayes error in the target domain, and $\mathbf{0}_{d}$ and $\mathbf{1}_{d}$ are $d\times 1$ all-zero and all-one vectors, respectively. For the scale matrices of Wishart distributions we set $\mathbf{M}_{t}^{y}=k_{t}\mathbf{I}_{d}$, $\mathbf{M}_{s}^{y}=k_{s}\mathbf{I}_{d}$, and $\mathbf{M}_{ts}^{y}=k_{ts}\mathbf{I}_{d}$ where $\mathbf{I}_{d}$ is the identity matrix of rank $d$.
To ensure that the joint scale matrix $\mathbf{M}^{y}=\begin{pmatrix}
\mathbf{M}_{t}^{y} & \mathbf{M}_{ts}^{y}\\ 
{\mathbf{M}_{ts}^{y}}^{T} & \mathbf{M}_{s}^{y}
\end{pmatrix}$ is positive definite $\forall y\in\left\{0,1\right\}$, we set $k_{ts}=\alpha \sqrt{k_{t}k_{s}}$ with $k_{t}>0$, $k_{s}>0$, and $\left|\alpha\right|<1$. As in~\citep{Karbalayghareh2018}, the value of $\left|\alpha\right|$ controls the amount of relatedness between the source and target domains.
To control the level of relatedness by adjusting only $\left|\alpha\right|$ without involving other confounding factors, we set $k_{t}=k_{s}=1$.
To sample from the joint prior, we first sample from a non-singular Wishart distribution $W_{2d}\left(\mathbf{M}^{y}, \nu\right)$ to get a block partitioned sample of the form $\mathbf{\Lambda}^{y}=\begin{pmatrix}
\mathbf{\Lambda}_{t}^{y} & \mathbf{\Lambda}_{ts}^{y}\\ 
{\mathbf{\Lambda}_{ts}^{y}}^{T} & \mathbf{\Lambda}_{s}^{y}
\end{pmatrix}$ from which we extract $\left(\mathbf{\Lambda}_{t}^{y}, \mathbf{\Lambda}_{s}^{y}\right)$. Afterwards, we sample $\mu_{z}^{y} \sim \mathcal{N}\left(\mathbf{m}_{z}^{y}, \left(\kappa_{z}^{y}\mathbf{\Lambda}_{z}^{y}\right)^{-1}\right)$ for $z\in \left\{s,t\right\}$ and $y\in \left\{0,1\right\}$. We use in our simulations two types of datasets. Training datasets that contain samples from both domains and testing datasets that contain only samples from the target domain. In all the simulations we consider testing datasets of 1,000 data points per class and we assume equal prior probabilities for the classes.

\subsection{RNA sequencing (RNA-seq) datasets}

To evaluate the performance of the TL-based BEE on real-world data, we consider classifying patients diagnosed with Schizophrenia using transcriptomic profiles collected from psychiatric disorder studies~\citep{Gandal2018}. Based on two RNA-seq datasets listed in Table~\ref{Table1}, we selected the transcriptomic profiles of 3 genes, based on a stringent feature selection procedure comprising the analysis of differential gene expression, clustering of gene-gene interactions, and statistical testing for multivariate normality. More specifically, we focus on analyzing the astrocyte-related cluster of differentiation 4 (CD4), found to be significantly up-regulated in subjects with Schizophrenia~\citep{Gandal2018}. We select the top three hub genes that collectively satisfy the Royston's multivariate normality test applied to the full datasets for both classes at a significance level of 99\%. The identified genes satisfying all the aforementioned criteria include SOX9, AHCYL1, and CLDN10 with an average module centrality of $0.86$ measured by genes' module membership (kME)~\citep{Gandal2018}.
In addition to normalization and quality control performed in~\citep{Gandal2018}, the selected features in both datasets have been further standardized to zero means and unit variances across both classes as in~\citep{Hoffman2013,Karbalayghareh2018}.
\begin{table}[h!]
\centering
\begin{tabular}{|c|c|c|cc}
\hline
\multirow{2}{*}{\textbf{Disease}} & \multicolumn{2}{c|}{\textbf{Number of samples}} & \multicolumn{1}{c|}{\multirow{2}{*}{\textbf{Brain region}}} & \multicolumn{1}{c|}{\multirow{2}{*}{\textbf{Dataset}}} \\ \cline{2-3}
 & \textbf{Case} & \textbf{Control} & \multicolumn{1}{c|}{} & \multicolumn{1}{c|}{} \\ \hline
\multirow{2}{*}{\textbf{Schizophrenia}} & 53 & 53 & \multicolumn{1}{c|}{Frontal cortex} & \multicolumn{1}{c|}{syn4590909~\citep{Gandal2018}} \\ \cline{2-5} 
 & 262 & 293 & \multicolumn{1}{c|}{DLPFC} & \multicolumn{1}{c|}{syn2759792~\citep{Fromer2016}} \\ \hline
\textbf{TOTAL} & 315 & 346 &  &  \\ \cline{1-3}
\end{tabular}%
\caption{Independent datasets sampled from two different brain tissues.}
\label{Table1}
\end{table}
We consider the dataset {\tt syn2759792}, sampled from the brain dorsolateral prefrontal cortex area (DLPFC), as a target dataset and {\tt syn4590909}, sampled from the frontal cortex region (FC), as a source dataset. Among 555 postmortem brain samples in {\tt syn2759792}, we randomly draw 5 samples per class as training data and we use the remaining samples to evaluate the classification error.
This process is repeated 10,000 times to estimate the average MSE deviation of the TL-based BEE from the true error.
To tune the model hyperparameters, we assume shared values for case and control samples in source and target domains. We set $\nu=10d=30$, $n_{t}=5$, and we consider source datasets of different sizes ($n_{s}\in\left\{10,~30,~50\right\}$) to conduct a greedy search to identify the optimal relatedness coefficient $\left|\alpha\right|$ across domains. At each iteration, we randomly permute the source samples for statistical significance. The remaining parameters are set as follows: $\kappa_{t}=n_{t}$, $\kappa_{s}=n_{s}$, and $k_{t}=k_{s}=\frac{1}{\nu}$ such that the mean of the Wishart precision matrices will be equal to the identity matrix which matches the normal standardization. For mean vectors $\mathbf{m}_{t}$ and $\mathbf{m}_{s}$, we pool all case and control samples in each domain and consider their means, respectively.

\subsection{Classifier design}

For comprehensive evaluation of our novel TL-based error estimator, we design and perform a set of experiments. The proposed TL-based estimator is applied to a collection of classifiers with different levels of learning capacities and tested under various scenarios.
To separate error estimation from classifier design, we start by analyzing the performance of the TL-based BEE estimator for fixed classifiers that do not depend on training data. This set-up distinctly reveals the major characteristics of the TL-based BEE, excluding any confounding factors that may stem from classifier design and the performance of the resulting classifier.

Next, we also conduct a comparative study of the TL-based BEE performance with respect to other widely-used error estimators, which include resubstitution, CV, LOO, and the 0.632-bootstrap estimators. As these popular data-driven estimators involve  classifier design on the training data, we will also consider a TL-based classifier designed on target and source data that operates in the target domain for comparison. For this, we employ the optimal Bayesian transfer learning (OBTL) classifier introduced in~\citep{Karbalayghareh2018}, which shares the same Bayesian framework on which our TL-based BEE is developed. In what follows, we recall the definition of each classifier considered in our evaluations and also present the details of the evaluation experiments performed in this study.

In the first set of experiments, we employ a fixed quadratic classifier assuming we know beforehand the true target parameters. For normally distributed data, this quadratic classifier corresponds also to the Bayes classifier that is optimal for the given feature-label distributions. Using quadratic discriminant analysis (QDA), we define
$\Psi_{QDA}\left(\mathbf{x}\right)=\mathbf{x}^{T}\mathbf{A}\mathbf{x}+\mathbf{b}^{T}\mathbf{x}+c$, where
\begin{align}
    \begin{split}
        &\mathbf{A}=-\frac{1}{2}\left(\mathbf{\Lambda_{t}^{1}}-\mathbf{\Lambda_{t}^{0}}\right), \quad
        \mathbf{b}=\mathbf{\Lambda_{t}^{1}}\mu_{t}^{1}-\mathbf{\Lambda_{t}^{0}}\mu_{t}^{0},\\
        &c=-\frac{1}{2}\left({\mu_{t}^{1}}^T\mathbf{\Lambda_{t}^{1}}\mu_{t}^{1}-{\mu_{t}^{0}}^T\mathbf{\Lambda_{t}^{0}}\mu_{t}^{0}\right)-\frac{1}{2}\ln\left(\frac{\left|\mathbf{\Lambda_{t}^{0}}\right|}{\left|\mathbf{\Lambda_{t}^{1}}\right|}\right).
    \end{split}
\end{align}
The error estimation problem turns out to be an estimation of the Bayes error that coincides here with the true error of the designed QDA. Obviously, this classifier is independent from any observed sample as it is fixed assuming known true model parameters. Without loss of generality, we apply the TL-based BEE using labeled observations from a compound dataset compiled from target and source domains.

In the second set of experiments we investigate the behaviour of the TL-based BEE within the class of sub-optimal classifiers. To this end, we consider a linear classifier derived through linear discriminant analysis (LDA) and we define
$\Psi_{LDA}\left(\mathbf{x}\right)=\mathbf{a}^{T}\mathbf{x}+b$ where $\mathbf{a}=\mathbf{S}_{t}^{-1}\left(\mu_{t}^{1}-\mu_{t}^{0}\right)$, $b=-\frac{1}{2}~\mathbf{a}^{T}\left(\mu_{t}^{1}+\mu_{t}^{0}\right)$, and the average covariance $\mathbf{S}_{t}$ is given by 
\begin{equation}
    \mathbf{S}_{t}=\frac{\left(\mathbf{\Lambda}_{t}^{0}\right)^{-1} +\left(\mathbf{\Lambda}_{t}^{1}\right)^{-1} }{2}.
\end{equation}
Our goal is then to approximate the true error of this sub-optimal classifier using transfer learning.

Next, we evaluate the performance of the TL-based BEE for the OBTL classifier that can take advantage of both source and target domain data. The OBTL classifier is defined by 
\begin{equation}
    \Psi_{OBTL}\left(\mathbf{x}\right) = \operatorname*{arg~max}_{y \in \left\{0,1\right\}} \mathcal{O}_{OBTL}\left(\mathbf{x}|y\right),
\end{equation}
where the objective function $\mathcal{O}_{OBTL}\left(\mathbf{x}|y\right)$ denotes the effective class-conditional density $p\left(\mathbf{x}|y\right)$ given by the following theorem:

\emph{Theorem 2~\citep{Karbalayghareh2018}:}
The effective class-conditional density, denoted by $p\left(\mathbf{x}|y\right) = \mathcal{O}_{OBTL}\left(\mathbf{x}|y\right)$, in the target domain is given by
\begin{align}
    \begin{split}
        &\mathcal{O}_{OBTL}\left(\mathbf{x}|y\right)=\pi^{-\frac{d}{2}}\left(\frac{\kappa_{t,n}^{y}}{\kappa_{x}^{y}}\right)^{\frac{d}{2}}\Gamma_{d}\left(\frac{\nu^{y}+n_{t}^{y}+1}{2}\right)\\
        &\times \Gamma_{d}^{-1}\left(\frac{\nu^{y}+n_{t}^{y}}{2}\right)\left|\mathbf{T}_{x}^{y}\right|^{\frac{\nu^{y}+n_{t}^{y}+1}{2}}\left|\mathbf{T}_{t}^{y}\right|^{-\frac{\nu^{y}+n_{t}^{y}}{2}}\\
        &\times \pFq{2}{1}{\frac{\nu^{y}+n_{s}^{y}}{2}, \frac{\nu^{y}+n_{t}^{y}+1}{2}}{\frac{\nu^{y}}{2}}{\mathbf{T}_{s}^{y}\mathbf{F}^{y}\mathbf{T}_{x}^{y}{\mathbf{F}^{y}}^{T}}\\
        &\times \pFq{2}{1}{\frac{\nu^{y}+n_{s}^{y}}{2}, \frac{\nu^{y}+n_{t}^{y}}{2}}{\frac{\nu^{y}}{2}}{\mathbf{T}_{s}^{y}\mathbf{F}^{y}\mathbf{T}_{t}^{y}{\mathbf{F}^{y}}^{T}}^{-1},
    \end{split}
\end{align}
where 
\begin{align}
    \begin{split}
        \kappa_{x}^{y} &= \kappa_{t,n}^{y}+1 = \kappa_{t}^{y}+n_{t}^{y}+1,\\
        \left(\mathbf{T}_{x}^{y}\right)^{-1} &= \left(\mathbf{T}_{t}^{y}\right)^{-1} + \frac{\kappa_{t,n}^{y}}{\kappa_{t,n}^{y}+1}\left(\mathbf{m}_{t,n}^{y}-\mathbf{x}\right)\left(\mathbf{m}_{t,n}^{y}-\mathbf{x}\right)^{T}.
    \end{split}
\end{align}

\subsection{Simulation set-up}

Figure~\ref{flow_chart} provides a combined illustration of the simulation set-up for all three classifiers.
For rigorous evaluation of the performance of the proposed TL-based BEE, we primarily focus our experiments on assessing the impact of using different types and amounts of source data. This is enabled by the joint prior imposed over the model parameters and controlled by the relatedness coefficient $\left|\alpha\right|$ that dictates the extent of interaction between the features in the two domains. For this purpose, we repeatedly conduct experiments following the flow chart in Fig.~\ref{flow_chart} with different relatedness values ($\left|\alpha\right| = \left[0.1, 0.3, 0.5, 0.7, 0.9, 0.95\right]$) where $\left|\alpha\right|=0.1$ corresponds to the lowest relatedness bewteen the two domains and $\left|\alpha\right|=0.95$ reflects the highest relatedness within the range of studied values.

\input{figures/flow_chart}

In the first set of experiments we start by drawing a joint sample $\left(\mathbf{\Lambda}_{t}^{y},\mathbf{\Lambda}_{s}^{y}\right)$ for each class $y\in \left\{0,1\right\}$ as previously described. Next, we iterate over the values of the hyperparameter $\vartheta$ to control $\mathbf{m}_{t}\left(\vartheta\right)$ through a dichotomic search to get a desired value $\tau$ of the Bayes error. This is achieved by drawing a sample $\mu_{t}^{y}\sim \mathcal{N}\left(\mathbf{m}_{t}\left(\vartheta\right), \left(\kappa_{t}^{y}\mathbf{\Lambda}_{t}^{y}\right)^{-1}\right)$ and then generating a test set based on the joint sample $\left(\mu_{t}^{y}, \mathbf{\Lambda}_{t}^{y}\right)$. Using this test set, we determine the true error of the optimal QDA derived from $\left(\mu_{t}^{y}, \mathbf{\Lambda}_{t}^{y}\right)$. If the desired Bayes error (true error of the designed QDA) is attained then the iteration stops, otherwise we update $\vartheta$ and reiterate.
In our experiments, we set $\tau=0.2$ to mimic a moderate level of classification complexity. This step is indeed crucial as it maintains the same level of complexity across the experiments and guarantees a fair comparison across different levels of relatedness. We note that this procedure is valid for general covariances as it acts only on updating the value of the mean parameter without altering the structure of the covariances nor the random mean vectors. Obviously, this approach to specify the Bayes error maintains the Bayesian transfer learning framework intact. However, it is not guaranteed to find values of target parameters that correspond to the desired Bayes error especially for high dimensions and complex classification (large Bayes error) as we will discuss in the next section. Once the problem complexity is set and the classifier is fixed, we generate $N_d=10,000$ training datasets that we use to evaluate the MSE of the TL-based BEE as depicted in Fig.~\ref{flow_chart}.
To estimate the TL-based BEE, we employ the importance sampling set-up previously described and we draw 1,000 MC samples from the importance density using HMC sampler.

In the second set of experiments, we follow a similar set-up using an LDA classifier designed based on the true model parameters. As before, we employ QDA to determine the Bayes error to maintain the same complexity level across different experiments. As in the first set of experiments, we use the TL-based BEE to estimate the true error of the designed LDA classifier.

In the last set of experiments on synthetic datasets, we conduct a comparative analysis study using an OBTL classifier designed using training datasets generated from the model parameters specified by the Bayes error. The error estimation task, in this scenario, aims at approximating the true error of the designed OBTL classifier determined using a large test set generated from the true feature-label distributions. As illustrated in Fig.~\ref{flow_chart}, QDA and LDA classifiers are fixed and derived from the true model parameters while the OBTL classifier is designed based on training datasets collected from the underlying feature-label distributions that correspond to the specified Bayes error. In all simulations, the designed classifiers are fixed given the observed samples and the TL-based BEE estimator is safely applied.
Finally, regarding synthetic datasets, we note that the flow chart in Fig.~\ref{flow_chart} is valid for all classifiers (QDA, LDA, and OBTL) and the notation $\Psi$ designates the classifier of interest in the corresponding set of experiments. For instance, in the second set of experiments, $\Psi$ refers to $\Psi_{LDA}$.

In addition to this in-depth analysis of the performance, behavior, and characteristics of our proposed TL-based BEE based on synthetic datasets, we also performed additional validation based on real-world biological datasets. By using RNA-seq datasets {\tt syn2759792} and {\tt syn4590909}  taken from different brain regions for studying brain disorders, we train a QDA classifier using the  target data from the RNA-seq dataset {\tt syn2759792}, and we leverage the source data from {\tt syn4590909} to evaluate the performance of the proposed TL-based BEE.


In the following section, we present the experimental results with in-depth discussions.

\section{Results}
\label{sec:result}

\subsection{Performance on synthetic datasets}

We start by evaluating the performance of the proposed TL-based BEE in estimating the Bayes error, that corresponds to the true error of the QDA in the target domain, for different levels of $\left|\alpha\right|$ and different size combinations of the utilized source and target datasets.

\input{figures/QDA/QDA_n_s}
In Fig.~\ref{fig:Figure1} we investigate the behavior of the TL-based BEE when the target data is fixed while we vary the size of the source data. We show the results for $d=2$ in the top row, the results for $d=3$ in the second row, and the results for $d=5$ in the last row. The columns correspond to the results for target datasets with different sizes: $n_{t}=20$ on the left and $n_{t}=50$ on the right. The MSE curves show similar trends for all three values of $d$, where we can see that the deviation of the error estimate from the true error significantly decreases when highly related source data are employed. This behavior diminishes as the relatedness between the two domains decreases.
Notably, using large source datasets ($n_{s}\geq 200$) of moderate to small relatedness values ($\left|\alpha\right|\leq 0.7$) does not negatively impact the performance of the estimator for low dimensions ($d\in \left\{2,3\right\}$) as shown in rows 1 and 2 of Fig.~\ref{fig:Figure1}. As the dimensionality further increases ($d=5$), large source datasets of moderate to small relatedness slightly increase the deviation of the estimated error from the true error (i.e. $\left|\alpha\right|=0.7$ in the third row). This tiny asymptotic deviation is explained by potential undesirable effects of relying on large source datasets of modest relatedness. However, it is important to note that the proposed TL-based BEE in the context of the given Bayesian TL framework suppresses this behaviour,  as it does not directly depend on the source data but the information transfer occurs through the joint prior. The joint prior acts like a bridge through which the useful knowledge passes from the source to the target domain. Effects of using source data in different TL settings (especially, a non-Bayesian setting) may require further investigation.
Moreover, the simulation results in different columns show that the MSE deviation decreases as we rely on larger target datasets. However, the gain in performance as we use additional source data is reduced when target data are more abundant. This is illustrated by the slope of the MSE graphs that flattens as $n_{t}$ increases.
Finally, Fig.~\ref{fig:Figure1} shows that for higher dimensions, the MSE deviation tends to increase. This is expected as increasing the dimensionality generally leads to a more difficult error estimation problem.

\input{figures/QDA/QDA_n_t}
Next, Fig.~\ref{fig:Figure2} shows the MSE deviation with respect to the size of the target dataset for dimensions 2, 3, and 5.
The first column corresponds to the case of using source datasets of size $n_{s}=50$ and the second column shows the results for $n_{s}=200$. The  performance of the TL-based BEE estimator improves with the increasing availability of target data. We can also clearly see that the MSE deviation from the true error asymptotically converges to comparable values for all relatedness levels. When highly related source data are available, the TL-based estimator yields accurate estimation results even when the target dataset is small.
These results consolidate the findings in Fig.~\ref{fig:Figure1} about the redundancy of source data in the presence of abundant target data. Across all graphs in Fig.~\ref{fig:Figure2}, we can see that a relatedness coefficient $\left|\alpha\right|=0.95$ results in a nearly constant deviation from the true error as a function of target data size, which suggests that highly related source data $\left|\alpha\right|>0.95$ acts almost identically like the target data, regardless of the shift across the domains in terms of their means.
Similar to the trends shown in Fig.~\ref{fig:Figure1}, results across different rows of Fig.~\ref{fig:Figure2} demonstrate that the error estimation difficulty increases with the increase of dimensionality. This is clearly reflected in the MSE deviation from the true error in Fig.~\ref{fig:Figure2}, which shows that as the dimension increases from $d=2$ (top row) to $d=5$ (last row), the MSE increases by one order of magnitude.

Now, we aim at investigating the effect of classification complexity on the performance of the proposed TL-based BEE. To this end, we conduct simulations, in which we vary the Bayes error through a wide range of possible values and evaluate the TL-based BEE at each given Bayes error for different sizes of target data while using source datasets of a fixed size $n_{s}=200$. In binary classification, the Bayes error has an upper bound specified by the true error of random classification that is $0.5$, as every data point can be randomly assigned one of the class labels. Ideally, we would vary the Bayes error across the interval $\left[0,~0.5\right]$ as in~\citep{Dalton2011_B}. However, in our set-up, we do not impose any structure on the covariance matrices, nor do we assume they are scaled identities. This makes the control of the Bayes error much more difficult. Additionally, the joint sampling set-up within our Bayesian transfer learning framework inhibits any modification of the randomized parameters. Consequently, the only practical way to adjust the Bayes error is to tune the mean vector parameters $\mathbf{m}_{t}^{y}$ that specify the means for the class mean vectors $\mu_{t}^{y}$ with $y\in \left\{0,1\right\}$.
In our experiments, we were able to fully control the Bayes error for $d=2$ and we considered the following values $\left[0.05,~0.1,~0.15,~0.2,~0.25,~0.3,~0.35,~0.4,~0.45,~0.5\right]$. Achieving the same range of values for $d=3$ and $d=5$ was more challenging, and our implemented heuristic did not converge for high values of Bayes error as setting $\mathbf{m}_{t}^{0}=\mathbf{m}_{t}^{1}$ did not help in increasing the Bayes error. However, we were able to vary the Bayes error for $d=3$ within the range $\left[0.05,~0.1,~0.15,~0.2,~0.25,~0.3,~0.35,~0.4,~0.45\right]$, and for $d=5$, within $\left[0.05,~0.1,~0.15,~0.2,~0.25,~0.3,~0.35,~0.4\right]$, sufficient for observing the trends.

\input{figures/QDA/QDA_n_s_Bayes}
Figure~\ref{fig:qdansBayes} shows the MSE deviation with respect to the Bayes error for dimensions 2, 3, and 5. Results in the first column are obtained using target datasets of size 20 and those in the second column are obtained using target datasets of size 50. We can see that the Bayesian MMSE estimator performs best when using source data of high relatedness to the target domain as expected. For Bayes error in the range $\left [ 0.25, 0.35\right ]$, the MSE deviation from the true error is very high, which makes this range of Bayes error as the most challenging setting for error estimation. For a Bayes error of 0.2, the MSE deviation is average across all the experiments which confirms the validity of our previous assumption in selecting this value to investigate classification problems of moderate difficulty.
We note that the TL-based BEE shifts the performance in favor of low and high Bayes error levels. Indeed, the TL-based BEE performs well in this case because the estimated target parameters are sufficiently accurate, even with a small target sample.

\input{figures/QDA/QDA_d_5_n_s_flipped}
In addition to investigating the effect of different relatedness levels between source and target domains, we have examined in Fig.~\ref{fig:qdansFlipped} the performance of the TL-based BEE for the case  when the source class means are swapped between the two classes, such that they show opposite trends compared to the class means in the target domain. For this purpose, we reproduced the experiments in Fig.~\ref{fig:Figure1} after flipping the class means of source datasets with respect to the target classes (\textit{i.e.}, $\mathbf{m}_{s}^{y} = \mathbf{m}_{t}^{1-y}$, for $y\in \left\{0,1\right\}$). In the first column of Fig.~\ref{fig:qdansFlipped}, we use the generated source datasets as observed samples from the source domain. Interestingly, the obtained results match those observed in Fig.~\ref{fig:Figure1}. This postulates that the knowledge transfer across source and target domains in the context of the studied Bayesian TL framework does not depend on the arrangement of the class means in the source and target domains but only rests on the level of relatedness between the two domains. For verification, we have intentionally considered the same source datasets in the previous experiment as target datasets for estimating the TL-based BEE and we plotted the obtained results in the second column of Fig.~\ref{fig:qdansFlipped}.
Clearly, the TL-based BEE veers away from the true error as we consider additional source data points. This deviation is worse with poorly related source data ($\left|\alpha\right|=0.1$). These results confirm previous findings in~\citep{Karbalayghareh2018} that the joint prior model in the utilized Bayesian TL framework acts like a bridge that distills the useful knowledge from the source domain and effectively transfers it to the target domain.

\input{figures/LDA/LDA_n_s_Bayes}
Results from the second set of experiments that use an LDA classifier were similar as the ones obtained using the QDA classifier except for some differences in the performance of the TL-based BEE with respect to the Bayes error that we report in Fig.~\ref{fig:ldansBayes} (see Appendix E for additional results).
The TL-based BEE performance has similar trends with respect to small and moderate Bayes errors when compared to the presented results obtained using the QDA classifier.
A notable difference here is observed for large values of Bayes error where the TL-based BEE shows decreased performance in terms of MSE deviation from the true error, which is due to the fact that the employed LDA classifier is sub-optimal compared to the Bayes classifier. This is expected as linear decision boundaries tend to be more sensitive to deviations from true model parameters for highly overlapping class-conditional distributions.

\input{figures/OBTL/standard_compare}
In our final set of experiments using synthetic datasets, we compare the performance of the proposed TL-based BEE to standard error estimators for different dimensions and various source datasets of relatedness level $\left|\alpha\right|=0.9$ to the target domain for an OBTL classifier. We show in Fig.~\ref{fig:obtlnt} the MSE deviation with respect to different target dataset size. As clearly shown, our proposed TL-based BEE significantly outperforms all other standard error estimators by a substantial margin.
In agreement with previous findings in the literature, the standard error estimators perform comparably for low dimensions (\textit{i.e.}, $d=2$), where the bootstrap may show a slight advantage. As the dimensionality increases (\textit{i.e.}, $d=5$), the performance shift of the studied estimators becomes more apparent. For example, the resubstitution estimator performs poorly in the small sample regime while the bootstrap estimator outperforms LOO and CV. Furthermore, we could notice that increasing the size of the source dataset does not lead to any apparent performance improvement for the standard estimators. This is because these estimators do not directly depend on the source data for error estimation (as they are incapable of taking advantage of data from different yet relevant domains).
However, providing additional source data to the TL-based BEE considerably reduces the MSE deviation from the true error for all dimensions as shown in Fig.~\ref{fig:obtlnt}.
\subsection{Performance on real-world RNA-seq datasets}
To analyze the performance of the TL-based BEE on real-world data, we have trained a QDA classifier on a small target dataset that consists of five sample points per class extracted from {\tt syn2759792} in Table~\ref{Table1}. Using different source datasets collected from {\tt syn4590909}, we show in Fig.~\ref{fig:sczalpha} the MSE deviation of the TL-based BEE from the true error with respect to $\left|\alpha\right|$.
\input{figures/QDA/scz_alpha}
For all combinations and different sizes of source datasets, the FC brain region showed high relatedness to the DLPFC brain area where the optimal MSE deviation from the true error was obtained for $\left|\alpha\right|=0.99$.
Interestingly, findings in~\citep{Gandal2018} also confirm that {\tt syn4590909} and {\tt syn2759792} are highly related, as independent gene expression assays for both brain regions have consistently replicated the gradient of transcriptomic severity observed for three different types of psychiatric disorders, including bipolar disorder and Schizophrenia~\citep{Gandal2018}. We show in Fig.~\ref{fig:sczalphans} the increasing gain in accuracy of the TL-based BEE in estimating the classification error after using additional labeled observations from the source domain. These results again confirm the efficacy and advantages of our novel TL-based error estimation scheme, compared to other standard error estimation methods, when additional data are available from different source domains that are nevertheless relevant to the target domain.
From a practical perspective, our proposed TL-based BEE has the potential to facilitate the analysis of real-world datasets, in the context of small-sample classification. Challenges of designing and evaluating classifiers (\textit{e.g.}, for clinical diagnosis or prognosis) in a small-sample setting are prevalent in scientific studies in life sciences and physical sciences due to the formidable cost, time, and effort required for data acquisition. This is certainly the case for the example that we considered in this section, where invasive brain biopsies would be needed to get the data.  
\section{Conclusions}
\label{sec:conclusion}

In this study, we have introduced a novel  Bayesian MMSE estimator that draws from concepts and theories in transfer learning to enable accurate estimation of classification error in the (target) domain of interest by utilizing samples from other closely related (source) domains. We have developed an efficient and robust importance sampling set-up that can be used for accurate error estimation in small-sample scenarios that often arise in many real-world scientific problems. Extensive performance analysis based on both synthetic and real biological data demonstrates the outstanding performance of the proposed TL-based BEE clearly outperforming conventional estimators.

In our proposed framework, Laplace approximations were used to alleviate the complexity associated with the exact evaluation of generalized hypergeometric functions that appear in the posterior distribution of the target parameters.
Beyond the Gaussian model assumed in the validation experiments, we also provided a general mathematical definition for the TL-based BEE that can directly be extended to applications with non-Gaussian distributions where the model parameters can be inferred through MCMC methods.
In this study, target and source domains were related through the joint prior of the model parameters that transfers useful knowledge across domains. A key property of the proposed TL-based BEE is its elegant ability to handle the uncertainty about the model parameters by integrating this prior with data, deducing robust estimates by accounting for all possible parameter values.

Paramount practical challenges for the TL-based BEE include the identification of suitable source domains that share similar families of distributions as the target domain of interest. This is crucial as the relatedness across domains is mathematically modeled assuming the similarity of the feature-label distributions across domains. Furthermore, learning the joint prior for the distributions and modeling the relatedness between different domains may also present an engineering challenge. While techniques for knowledge-driven prior construction have been developed~\citep{Boluki2017a, Boluki2017b}, such techniques yet have to be developed for joint prior construction for relevant domains, which is an important future research direction.
\section*{Acknowledgment}
This work was supported in part by the Department of Energy (DOE) under Award DE-SC0019303.\\
Portions of this research were conducted with the advanced computing resources provided by Texas A\&M High Performance Research Computing.
\newpage
\begin{appendices}
\renewcommand{\theequation}{\thesection.\arabic{equation}}
\setcounter{equation}{0}
\section{Binary classification}
A standard binary classification problem involves a sample space $\mathcal{X}$ associated with the real-valued $d$-dimensional feature space $\mathbb{R}^{d}$ where the class labels are modeled by a random variable $y$ in the output space $\mathcal{Y} = \left \{ 0,1 \right \}$.
Classification aims at designing an optimal function $\psi$ that assigns each feature vector of the form $\mathbf{x}=\left ( x_{1}, x_{2},\cdots,x_{d}  \right ) \in \mathcal{X}$ to its corresponding true class-label $y \in \mathcal{Y}$. The joint relationship between $\mathbf{x}$ and $y$ is fully characterised by the feature-label distribution $f\left ( \mathbf{x},y \right )$.\\
Using the standard Bayesian terminology, the class-conditional density is denoted by $f\left ( \mathbf{x}~|~y \right )$ and the prior probability by $P\left ( y\right )$. In binary classification, the class prior probabilities are fully characterized by one parameter $c=P\left ( y=0\right )=1-P\left ( y=1\right )$.
The classification error $\varepsilon$ of a classifier $\psi(\cdot)$ is determined either using the feature-label distribution $f\left ( \mathbf{x},y \right )$ or the class-conditional distribution $f\left ( \mathbf{x}~|~y \right )$.\\
Using $f\left ( \mathbf{x},y \right )$, $\varepsilon$ represents the probability of misclassification given by
\begin{equation}
    P\left ( \psi\left ( \mathbf{x} \right )\neq y \right )=E\left [ \left | y- \psi\left ( \mathbf{x} \right )\right | \right ].
\end{equation}
Using the class-conditional densities $f\left ( \mathbf{x}~|~y \right )$, $\varepsilon$ is decomposed into two error components contributed by each class and denoted by $\varepsilon^{y}$.\\
Thus, $\varepsilon$ is expressed as
\begin{equation}
    \varepsilon = c~\varepsilon^{0}+\left ( 1-c\right )~\varepsilon^{1},
\end{equation}
where
\begin{align}
    \begin{split}
        \varepsilon^{y}&=P\left( \psi\left(\mathbf{x}\right)=1-y~|~y\right)\\
        &=\int_{\psi\left(\mathbf{x}\right)=1-y}f\left( \mathbf{x}~|~y\right)~d\mathbf{x}.
    \end{split}
\end{align}

When the feature-label distribution is unknown, a sample $\mathcal{S}_{n}=\left \{ \left ( \mathbf{x}_{1},y_{1}\right ), \left ( \mathbf{x}_{2},y_{2}\right ), \cdots, \left ( \mathbf{x}_{n},y_{n}\right )\right\}$ of size $n$ is drawn from $f\left ( \mathbf{x},y \right )$ through measurements collected from the data generating process.\\
Instead of finding $\left (\psi, \varepsilon \right )$, the classification problem turns out to be a search for another pair $\left (\psi_{n}, \varepsilon_{n} \right )$ that is a function of the observed sample $\mathcal{S}_{n}$. As $\varepsilon_{n}$ is also unknown for unknown feature-label distributions, an estimate $\hat{\varepsilon}$ that accurately approximates $\varepsilon_{n}$ is considered. Designing $\psi_{n}$ is generally possible through a classification rule ($\Psi\left( \mathcal{S}_{n}\right )=\psi_{n}$) defined by
\begin{equation}
    \Psi :\left [ \mathbb{R}^{d} \times \left \{ 0,1 \right \}\right ]^{n}\rightarrow \mathcal{C},
\end{equation}
where
\begin{equation}
    \mathcal{C}=\left \{ \psi\ |\ \psi : \mathbb{R}^{d}\rightarrow \left \{ 0,1 \right \} \right \}
\end{equation}
is the class containing all possible classifiers.\\
Similarly, deriving $\hat{\varepsilon}$ is possible through an estimation rule ($\Upsilon\left( \mathcal{S}_{n}\right )=\hat{\varepsilon}$) defined by
\begin{equation}
    \Upsilon :\left [ \mathbb{R}^{d} \times \left \{ 0,1 \right \}\right ]^{n}\rightarrow \left [ 0,1\right ].
\end{equation}
From an error estimation perspective, we lock our focus on evaluating the performance of $\hat{\varepsilon}$ regardless of the design of $\psi_{n}$.
\setcounter{equation}{0}
\section{Bayesian MMSE estimator}
The BEE has been introduced for the first time in~\citep{Dalton2011_A, Dalton2011_B} as a parametric estimator that assumes uncertainty in the parameters of a family of feature-label distributions $f\left ( \mathbf{x},y \right )$.
For a specific sample realization $S_{n} \in \mathcal{S}_{n}$, and a parameter vector $\mathbf{\theta}$, for which the notation of the feature-label distribution reduces to $f_{\mathbf{\theta}}\left ( \mathbf{x},y \right )$, the BEE is given by
\begin{equation}\label{BEE_long}
    \hat{\varepsilon}\left (S_{n} \right )=E_{\mathbf{\theta}}\left[\varepsilon_{n}\left( \mathbf{\theta},S_{n}~|~S_{n}\right) \right].
\end{equation}
After omitting the sample $S_{n}$ from the notation of the posterior density $\pi^{*}\left( \theta~|~S_{n} \right)$ for simplicity, the BEE is shortly expressed as
\begin{equation}
    \hat{\varepsilon}=E_{\pi^{*}}\left[\varepsilon_{n} \right].
\end{equation}
The BEE has proven to be optimal when averaged over a given family of feature-label distributions and unbiased when averaged over a given family and all samples~\citep{Dalton2011_A}.
In binary classification the parameter vector $\theta$ can be expressed using three components ($\theta=\left [\theta_{0}, \theta_{1}, c \right ]$), where $\theta_{0}$ and $\theta_{1}$ model the class conditional distributions and $c$ represents the prior probability of class 0.\\
Assuming that $c$, $\theta_{0}$, and $\theta_{1}$ are independent prior to observing the data, the BEE can be further decomposed as
\begin{equation}
    \hat{\varepsilon}=E_{\pi^{*}}\left[c \right]~E_{\pi^{*}}\left[\varepsilon_{n}^{0} \right]+\left(1-E_{\pi^{*}}\left[c \right]\right)~E_{\pi^{*}}\left[\varepsilon_{n}^{1} \right],
\end{equation}
where
\begin{equation}
    E_{\pi^{*}}\left[\varepsilon_{n}^{y}\right]=\int_{\Theta_{y}}\varepsilon_{n}^{y}\left(\theta_{y}\right)~\pi^{*}\left(\theta_{y}\right)~d\theta_{y}
\end{equation}
with $\Theta_{y}$ being the parameter space that contains all possible values for $\theta_{y}$.

\setcounter{equation}{0}
\section{Laplace approximations of confluent and Gauss hypergeometric
functions of matrix argument}
\emph{Definition 1~\citep{Nagar2017}:} The generalized hypergeometric function of one matrix argument is defined by
\begin{equation}\label{HG_series}
    \pFq{p}{q}{a_{1}, \cdots, a_{p}}{b_{1}, \cdots, b_{q}}{\mathbf{X}}=\sum_{k=0}^{\infty}\sum_{\kappa \vdash k}\frac{\left(a_{1}\right)_{\kappa}\cdots \left(a_{p}\right)_{\kappa}}{\left(b_{1}\right)_{\kappa}\cdots \left(b_{q}\right)_{\kappa}}\frac{C_{\kappa}\left(\mathbf{X}\right)}{k!},
\end{equation}
where $\left(a_{i}\right)_{i=1\cdots p}$ and $\left(b_{j}\right)_{j=1\cdots q}$ are arbitrary complex numbers, $C_{\kappa}\left(\mathbf{X}\right)$ is the zonal polynomial of the complex symmetric matrix $\mathbf{X}$ of order $d$ that corresponds to the ordered partition $\kappa$ of $k$ defined as $\kappa=\left(k_{1}, \cdots, k_{d}\right),~k_{1}\geq \cdots \geq k_{d} \geq 0,~\sum_{i=1}^{d}k_{i}=k$, and $\sum_{\kappa \vdash k}$ denotes the summation over all ordered partitions $\kappa$ of $k$. The generalized hypergeometric coefficient $\left(a\right)_{\kappa}$ is defined by
\begin{equation}
    \left(a\right)_{\kappa}=\prod_{i=1}^{d}\left(a-\frac{i-1}{2}\right)_{k_{i}},
\end{equation}
where $\left(a\right)_{r}=a\left(a+1\right)\cdots\left(a+r-1\right),~r=1,~2\cdots,$ with $\left(a\right)_{0}=1$.\\
Conditions for convergence of the series in~\eqref{HG_series} are covered in~\citep{Constantine1963}. Some special cases of matrix-variate generalized hypergeometric functions include
\begin{align}
    \begin{split}
        \pFq{0}{1}{.}{b}{\mathbf{X}}&=\sum_{k=0}^{\infty}\sum_{\kappa \vdash k}\frac{C_{\kappa}\left(\mathbf{X}\right)}{\left(b\right)_{\kappa}~k!}; \\
        \pFq{1}{1}{a}{b}{\mathbf{X}}&=\sum_{k=0}^{\infty}\sum_{\kappa \vdash k}\frac{\left(a\right)_{\kappa}}{\left(b\right)_{\kappa}}\frac{C_{\kappa}\left(\mathbf{X}\right)}{k!}; \\
        \pFq{2}{1}{a,b}{c}{\mathbf{X}}&=\sum_{k=0}^{\infty}\sum_{\kappa \vdash k}\frac{\left(a\right)_{\kappa}~\left(b\right)_{\kappa}}{\left(c\right)_{\kappa}}\frac{C_{\kappa}\left(\mathbf{X}\right)}{k!},~\left \| \mathbf{X} \right \|<1,
    \end{split}
\end{align}
where $\left \| \mathbf{X} \right \|$ denotes the maximum of the absolute values of the eigenvalues of $\mathbf{X}$. $\pFq{1}{1}{a}{b}{\mathbf{X}}$ and $\pFq{2}{1}{a,b}{c}{\mathbf{X}}$ are respectively called confluent and Gauss matrix-variate hypergeometric functions.

The confluent hypergeometric function has the following integral representation:
\begin{equation}
    \pFq{1}{1}{a}{b}{\mathbf{Z}}=B_{d}^{-1}\left(a,b-a\right)\int_{0_{d}<\mathbf{Y}<\mathbf{I}_{d}}\mbox{etr}\left(\mathbf{Z}\mathbf{Y}\right)\left|\mathbf{Y}\right|^{a-\frac{d+1}{2}}\left|\mathbf{I}_{d}-\mathbf{Y}\right|^{b-a-\frac{d+1}{2}}d\mathbf{Y}, 
\end{equation}
which is valid under the following conditions:
\begin{itemize}
    \item $\mathbf{Z}\in \mathbb{C}^{d\times d}$ is symmetric; 
    \item $\Re\left(a\right)>\frac{d-1}{2}$; 
    \item $\Re\left(b-a\right)>\frac{d-1}{2}$,
\end{itemize}
where $B_{d}\left(\alpha,\beta\right)$ is the multivariate beta function defined by:
\begin{equation}
    B_{d}\left(\alpha,\beta\right)=\frac{\Gamma_{d}\left(\alpha\right)\Gamma_{d}\left(\beta\right)}{\Gamma_{d}\left(\alpha+\beta\right)}=\int_{0_{d}<\mathbf{Y}<\mathbf{I}_{d}}\left|\mathbf{Y}\right|^{\alpha-\frac{d+1}{2}}\left|\mathbf{I}_{d}-\mathbf{Y}\right|^{\beta-\frac{d+1}{2}}d\mathbf{Y}
\end{equation}
with $\Gamma\left(a\right)$ denoting the multivariate gamma function defined by:
\begin{equation}
    \Gamma\left(a\right)=\int_{0_{d}<\mathbf{Y}<\mathbf{I}_{d}}\mbox{etr}\left(\mathbf{-Y}\right)\left|\mathbf{Y}\right|^{a-\frac{d+1}{2}}d\mathbf{Y}.
\end{equation}
Laplace approximations provide accurate evaluations of integrals of the form:
\begin{equation}
    I=\int_{y\in\mathbf{D}}h\left(y\right)\mbox{e}^{-\lambda g\left(y\right)}dy, 
\end{equation}
where $\mathbf{D}\subseteq\mathbb{R}^{d}$ is an open set and $\lambda$ is a real-valued parameter, when $g\left(y\right)$ has a unique minimum over the closure of $\mathbf{D}$, and this minimum occurs at stationary point $\hat{y}\in \mathbf{D}$ of $g\left(y\right)$.

This approximation is given by:
\begin{equation}\label{Laplace}
    \tilde{I}=\left(2\pi\right)^{\frac{d}{2}}\lambda^{\frac{-d}{2}}\left|{g}''\left(\hat{y}\right)\right|^{\frac{-1}{2}}h\left(\hat{y}\right)\mbox{e}^{-\lambda g\left(\hat{y}\right)}, 
\end{equation}
where ${g}''\left(y\right)=\frac{\partial^{2}g\left(y\right)}{\partial y\partial y^{T}}$ is the Hessian of $g\left(y\right)$.
In the case of matrix-variate hypergeometric functions, an important invariance property that facilitates the selection of $\left(g;h\right)$ is that $\pFq{1}{1}{a}{b}{\mathbf{Z}}$ and $\pFq{2}{1}{a,b}{c}{\mathbf{Z}}$ depend only on the eigenvalues of $\mathbf{Z}$ when $\mathbf{Z}$ is symmetric. As such, $\mathbf{Z}$ can be assumed diagonal. Furthermore, we focus on the case where $\mathbf{Z}$ is real and we write instead $\mathbf{X}$.

A favorable representation of $\left(g;h\right)$ as discussed in~\citep{Butler2002} is given by:
\begin{equation}
    g\left(\mathbf{Y}\right)=-\mbox{tr}\left(\mathbf{X}\mathbf{Y}\right)-a\log\left|\mathbf{Y}\right|-\left(b-a\right)\log\left|\mathbf{I}_{d}-\mathbf{Y}\right|
\end{equation}
and
\begin{equation}
    h\left(\mathbf{Y}\right)=B_{d}^{-1}\left(a,b-a\right)\left|\mathbf{Y}\right|^{-\frac{d+1}{2}}\left|\mathbf{I}_{d}-\mathbf{Y}\right|^{-\frac{d+1}{2}}. 
\end{equation}
Using~(\ref{Laplace}), the Laplace approximation to $\pFq{1}{1}{a}{b}{\mathbf{X}}$ is given by:
\begin{equation}
    \pFqtilde{1}{1}{a}{b}{\mathbf{X}}=2^{\frac{d}{2}}\pi^{\frac{d\left(d+1\right)}{4}}B_{d}^{-1}\left(a,b-a\right)J_{1,1}^{-\frac{1}{2}}\prod_{i=1}^{d}\left\{\hat{y}_{i}^{a}\left(1-\hat{y}_{i}\right)^{b-a}\mbox{e}^{x_{i}\hat{y}_{i}}\right\}, 
\end{equation}
where $\hat{y}_{i}$ is given by:
\begin{equation}
    \hat{y}_{i}=\frac{2a}{b-x_{i}+\sqrt{\left(x_{i}-b\right)^{2}+4ax_{i}}},
\end{equation}
and
\begin{equation}
    J_{1,1}=\prod_{i=1}^{d}\prod_{j=1}^{d}\left\{a\left(1-\hat{y}_{i}\right)\left(1-\hat{y}_{j}\right)+\left(b-a\right)\hat{y}_{i}\hat{y}_{j}\right\}. 
\end{equation}
A more accurate approximation emerges from the calibrated expression given by:
\begin{equation}
    \pFqhat{1}{1}{a}{b}{\mathbf{X}}=\frac{\pFqtilde{1}{1}{a}{b}{\mathbf{X}}}{\pFqtilde{1}{1}{a}{b}{\mathbf{0}}}
\end{equation}
since $\pFq{1}{1}{a}{b}{\mathbf{0}}=1$. Consequently, the calibrated Laplace approximation to $\pFq{1}{1}{a}{b}{\mathbf{X}}$ is given by:
\begin{equation}
    \pFqhat{1}{1}{a}{b}{\mathbf{X}}=b^{bd-\frac{d\left(d+1\right)}{4}}R_{1,1}^{-\frac{1}{2}}\prod_{i=1}^{d}\left\{\left(\frac{\hat{y}_{i}}{a}\right)^{a}\left(\frac{1-\hat{y}_{i}}{b-a}\right)^{b-a}\mbox{e}^{x_{i}\hat{y}_{i}}\right\}, 
\end{equation}
where
\begin{equation}
    R_{1,1}=\prod_{i=1}^{d}\prod_{j=1}^{d}\left\{\frac{\hat{y}_{i}\hat{y}_{j}}{a}+\frac{\left ( 1-\hat{y}_{i} \right )\left ( 1-\hat{y}_{j} \right )}{b-a}\right\}.
\end{equation}
Similarly, the Gauss hypergeometric function has the following representation:
\begin{equation}
    \pFq{2}{1}{a,b}{c}{\mathbf{Z}}=B_{d}^{-1}\left(a,c-a\right)\int_{0_{d}<\mathbf{Y}<\mathbf{I}_{d}}\left|\mathbf{Y}\right|^{a-\frac{d+1}{2}}\left|\mathbf{I}_{d}-\mathbf{Y}\right|^{c-a-\frac{d+1}{2}}\left|\mathbf{I}_{d}-\mathbf{Z}\mathbf{Y}\right|^{-b}d\mathbf{Y}, 
\end{equation}
which is valid under the following conditions:
\begin{itemize}
    \item $\mathbf{Z}\in \mathbb{C}^{d\times d}$ is symmetric; 
    \item $\Re\left(\mathbf{Z}\right)<\mathbf{I}_{d}$; 
    \item $\Re\left(a\right)>\frac{d-1}{2}$; 
    \item $\Re\left(c-a\right)>\frac{d-1}{2}$.
\end{itemize}
By following similar steps as in~\citep{Karbalayghareh2018}, the calibrated Laplace approximation to $\pFq{2}{1}{a,b}{c}{\mathbf{X}}$ is given by:
\begin{align}
    \pFqhat{2}{1}{a,b}{c}{\mathbf{X}}=c^{cd-\frac{d\left(d+1\right)}{4}}R_{2,1}^{-\frac{1}{2}}\prod_{i=1}^{d}\left\{\left(\frac{\hat{y}_{i}}{a}\right)^{a}\left(\frac{1-\hat{y}_{i}}{c-a}\right)^{c-a}\left(1-x_{i}\hat{y}_{i}\right)^{-b}\right\},
\end{align}
where
\begin{equation}
    R_{2,1}=\prod_{i=1}^{d}\prod_{j=1}^{d}\left\{\frac{\hat{y}_{i}\hat{y}_{j}}{a}+\frac{\left(1-\hat{y}_{i}\right)\left(1-\hat{y}_{j}\right)}{c-a}-\frac{bx_{i}x_{j}\hat{y}_{i}\hat{y}_{j}\left(1-\hat{y}_{i}\right)\left(1-\hat{y}_{j}\right)}{\left(1-x_{i}\hat{y}_{i}\right)\left(1-x_{j}\hat{y}_{j}\right)a\left(c-a\right)}\right\}.
\end{equation}
Detailed discussions in~\citep{Butler2002} show that the relative errors of the approximations to $\pFq{1}{1}{a}{b}{\mathbf{X}}$ and $\pFq{2}{1}{a,b}{c}{\mathbf{X}}$ are uniformly bounded. These bounds are as follows:
\begin{equation}
    \operatorname*{sup}_{b\geq b_{0},~a\in \mathbb{R},~\mathbf{X}\in \mathbb{R}^{d\times d}}\left|\log\left(\pFqhat{1}{1}{a}{b}{\mathbf{X}}\right)-\log\left(\pFq{1}{1}{a}{b}{\mathbf{X}}\right)\right|<\infty
\end{equation}
and
\begin{equation}
    \operatorname*{sup}_{c\geq c_{0},~a,b\in \mathbb{R},~\mathbf{0}_{d}\leq\mathbf{X}< \left(1-\varepsilon\right)\mathbf{I}_{d}}\left|\log\left(\pFqhat{2}{1}{a,b}{c}{\mathbf{X}}\right)-\log\left(\pFq{2}{1}{a,b}{c}{\mathbf{X}}\right)\right|<\infty
\end{equation}
given that $d$ is fixed, $b_{0}>\frac{d-1}{2}$, $c_{0}>\frac{d-1}{2}$, and $\forall \varepsilon \in \left(0,1\right)$.

Similar to the examples provided in~\citep{Butler2002} and the illustrations presented in~\citep{Karbalayghareh2018}, we provide simulations in Fig.~\ref{fig:hypergeoFvalues} to show the good numerical accuracy of the discussed Laplace approximations as compared to the exact evaluations. 
\input{figures/hg}
As the hypergeometric functions of matrix argument depend only on eigenvalues of their argument, we set $\mathbf{X}=\tau\mathbf{I}_{d}$ and we evaluate in Figs~\ref{fig:1F1tau} and \ref{fig:2F1tau} the accuracy of the approximations to $\pFq{1}{1}{a}{b}{\tau\mathbf{I}_{d}}$ and $\pFq{2}{1}{a,b}{c}{\tau\mathbf{I}_{d}}$, respectively, for $d=5,~a=3,~b=4,~c=6$ and w.r.t $\tau$ where $0<\tau<1$ guarantees the convergence as mentioned in~\citep{Butler2002}.

Figs.~\ref{fig:1F1b} and \ref{fig:2F1c} show the exact and approximate values of confluent and Gauss hypergeometric functions for variable $b$ and $c$ values, respectively, when $d=10, a=30, \left(b=50,~\mbox{for}~\pFq{2}{1}{a,b}{c}{\tau\mathbf{I}_{d}}\right)$, and $\tau=0.01$. As the integral representations of $\pFq{1}{1}{a}{b}{\tau\mathbf{I}_{d}}$ and $\pFq{2}{1}{a,b}{c}{\tau\mathbf{I}_{d}}$ are only valid when $\Re\left(b-a\right)>\frac{d-1}{2}$ and $\Re\left(c-a\right)>\frac{d-1}{2}$, respectively, the simulation results show that the Laplace approximations still give good accuracy even for values of $b$ and $c$ where the integral is not valid. This behaviour has been discussed thoroughly in~\citep{Butler2002} and was explained as an advantage of the calibrated approximations $\pFqhat{1}{1}{a}{b}{\tau\mathbf{I}_{d}}$ and $\pFqhat{2}{1}{a,b}{c}{\tau\mathbf{I}_{d}}$ where the singularities of the invalid integrals are removed.

\setcounter{equation}{0}
\section{Monte Carlo sampling}
We consider the problem of approximating:
\begin{equation}
    I=\int_{\mathcal{D}}f\left(x\right)p\left(x\right)dx=E_{p}\left[f\left(\mathbf{x}\right)\right]
\end{equation}
for an integrand $f\left(x\right)$ and a probability density $p\left(x\right)$ that are defined on $\mathcal{D}\subseteq \mathbb{R}^{d}$ where $E_{p}\left[\cdot\right]$ denotes the expectation for $\mathbf{x}\sim p$.
\subsection{Importance sampling}
If q is a positive probability density function on $\mathbb{R}^{d}$, we can write:
\begin{equation}
    I=\int_{\mathcal{D}}\frac{f\left(x\right)p\left(x\right)}{q\left(x\right)}q\left(x\right) dx=E_{q}\left[\frac{f\left(\mathbf{x}\right)p\left(\mathbf{x}\right)}{q\left(\mathbf{x}\right)}\right]. 
\end{equation}
By sampling $\left(\mathbf{x}_{i}\right)_{i=1\cdots n}\sim q$, the importance sampling estimate of $I$ is given by:
\begin{equation}\label{eq_Iq}
    \hat{I}_{q}= \frac{1}{n}\sum_{i=1}^{n}\frac{f\left(\mathbf{x}_{i}\right)p\left(\mathbf{x}_{i}\right)}{q\left(\mathbf{x}_{i}\right)}. 
\end{equation}
\emph{Theorem 1~\citep{Owen2013}:} Let $\hat{I}_{q}$ be given by~\eqref{eq_Iq} where $q\left(x\right)>0$ whenever $f\left(x\right)p\left(x\right)\neq0$. Then $E_{q}\left[\hat{I}_{q}\right] = I$, and $\mbox{Var}_{q}\left(\hat{I}_{q}\right)=\frac{\sigma_{q}^{2}}{n}$ where:
\begin{equation}
    \sigma_{q}^{2} = \int_{\mathcal{Q}}\frac{\left(f\left(x\right)p\left(x\right)\right)^{2}}{q\left(x\right)}dx-I^{2}=\int_{\mathcal{Q}}\frac{\left(f\left(x\right)p\left(x\right)-Iq\left(x\right)\right)^{2}}{q\left(x\right)}dx, 
\end{equation}
where $\mathcal{Q}=\left\{x~|~q\left(x\right)>0\right\}$.
\subsection{Self-normalized importance sampling}
When $p$ or $q$ has an unknown normalization constant, we resort to estimate the ratio $\frac{p\left(x\right)}{q\left(x\right)}$. Suppose we can compute an unnormalized ratio $\frac{p_{u}\left(x\right)}{q_{u}\left(x\right)}$ where $p_{u}\left(x\right) = a~p\left(x\right)$ and $q_{u}\left(x\right) = b~q\left(x\right)$ for $a,b>0$.

Then, we compute the ratio $\mathcal{L}_{u}\left(x\right)=\frac{p_{u}\left(x\right)}{q_{u}\left(x\right)}=\frac{a}{b}\frac{p\left(x\right)}{q\left(x\right)}$ and we consider the self-normalized importance sampling estimate given by:
\begin{equation}\label{eq_I_self_norm}
    \tilde{I}_q=\frac{\sum_{i=1}^{n}f\left(\mathbf{x}_{i}\right)\mathcal{L}_{u}\left(\mathbf{x}_{i}\right)}{\sum_{i=1}^{n}\mathcal{L}_{u}\left(\mathbf{x}_{i}\right)}, 
\end{equation}
where $\mathbf{x}_{i}\sim q$ are independent. The factor $\frac{a}{b}$ cancels from the numerator and denominator leading to:
\begin{equation}
    \tilde{I}_q=\frac{\sum_{i=1}^{n}f\left(\mathbf{x}_{i}\right)\mathcal{L}\left(\mathbf{x}_{i}\right)}{\sum_{i=1}^{n}\mathcal{L}\left(\mathbf{x}_{i}\right)}, 
\end{equation}
where $\mathcal{L}\left(x\right)=\frac{p\left(x\right)}{q\left(x\right)}$.

\emph{Theorem 2~\citep{Owen2013}:} Let p be a probability density function on $\mathbb{R}^{d}$ and let $f\left(x\right)$ be a function such that $I=\int f\left(x\right)p\left(x\right)dx$ exists. Suppose that $q\left(x\right)$ is a probability density function on $\mathbb{R}^{d}$ with $q\left(x\right)>0$ whenever $p\left(x\right)>0$. Let $\left(\mathbf{x}_{i}\right)_{i=1\cdots n}\sim q$ be independent and let $\tilde{I}_{q}$ be the self-normalized importance estimate in \eqref{eq_I_self_norm}. Then
\begin{equation}
    P\left(\operatorname{lim}_{n\to \infty}\tilde{I}_{q}=I\right)=1.
\end{equation}
\subsection{Control variates in importance sampling}
A control variate function $h\left(x\right)$ can be usefully combined with importance sampling if it satisfies $\int h\left(x\right)q\left(x\right)dx=\delta$, where $\delta$ is a known constant.

A more stable estimate to $I$ is given by:
\begin{equation}
    \tilde{I}_{q,\beta}=\frac{1}{n}\sum_{i=1}^{n}\left[\frac{f\left(\mathbf{x}_{i}\right)p\left(\mathbf{x}_{i}\right)}{q\left(\mathbf{x}_{i}\right)}- \beta h\left(\mathbf{x}_{i}\right)\right]+\beta \delta, 
\end{equation}
where $\left(\mathbf{x}_{i}\right)_{i=1\cdots n}\sim q$ and $\beta$ is a hyperparameter that controls the variance reduction.

\emph{Theorem 3~\citep{Owen2013}:} Let $q$ be a probability density function with $q\left(x\right)>0$ whenever $h\left(x\right)\neq0$ or $f\left(x\right)p\left(x\right)\neq0$. Then $E_{q}\left[\tilde{I}_{q,\beta}\right]=I$ for any $\beta \in \mathbb{R}$.

Let $\varsigma\left(x\right)=\frac{f\left(x\right)p\left(x\right)}{q\left(x\right)}$. The optimal value of $\beta$ is chosen in a way to minimize the variance of the random variable $Z = \varsigma\left(\mathbf{x}\right)-\beta\left(h\left(\mathbf{x}\right)-\delta\right)$.

We have:
\begin{equation}
    \mbox{Var}\left(Z\right) = \mbox{Var}\left(\varsigma\right) - 2\beta \mbox{Cov}\left[\varsigma, h\right]+\beta^{2}\mbox{Var}\left(h\right).
\end{equation}
Solving for $\beta$ we get:
\begin{equation}
    \beta^{\star}=\frac{\mbox{Cov}\left[\varsigma, h\right]}{\mbox{Var}\left(h\right)}. 
\end{equation}
The reduced variance is then given by:
\begin{equation}
    \mbox{Var}\left(Z\right)=\mbox{Var}\left(\varsigma\right)-\frac{\mbox{Cov}^{2}\left[\varsigma, h\right]}{\mbox{Var}\left(h\right)} = \mbox{Var}\left(\varsigma\right)\left(1-\rho^{2}\left[\varsigma, h\right]\right), 
\end{equation}
where $\rho\left[\varsigma, h\right]$ is a correlation coefficient given by:
\begin{equation}
    \rho\left[\varsigma, h\right]=\mbox{Corr}\left[\varsigma, h\right]=\frac{\mbox{Cov}\left[\varsigma, h\right]}{\mbox{Var}\left(\varsigma\right)\mbox{Var}\left(h\right)}. 
\end{equation}
In practice $\beta^{\star}$ is unknown. Instead, it can be estimated from the Monte Carlo samples. For $n$ independent samples $\left(\mathbf{x}_{i}\right)_{i=1\cdots n} \sim q$ we can compute:
\begin{equation}
\left\{\begin{matrix}
\widehat{\mbox{Var}}\left(h \right )=\frac{1}{n}\sum_{i=1}^{n}\left(h\left(\mathbf{x}_{i}\right )-\delta \right )^{2}\\ 
\widehat{I}=\frac{1}{n}\sum_{i=1}^{n}\varsigma\left(\mathbf{x}_{i}\right)\\ 
\widehat{\mbox{Cov}}\left[\varsigma,h \right ]=\frac{1}{n}\sum_{i=1}^{n}\left(\varsigma\left(\mathbf{x}_{i}\right)- \widehat{I}\right )\left(h\left(\mathbf{x}_{i}\right)- \delta\right )
\end{matrix}\right.
\end{equation}
and thus we determine:
\begin{equation}
    \widehat{\beta}^{\star}=\frac{\widehat{\mbox{Cov}}\left[\varsigma, h\right]}{\widehat{\mbox{Var}}\left(h\right)}. 
\end{equation}
\subsection{Integral evaluation of LDA error expectation as a control variate function}
From~\citep{Dalton2011_B}, we know that the posterior density of model parameters $\left(\mu^{y}, \Sigma^{y}\right)$ after observing the sample $S_{n}$ is a Gaussian inverse-Wishart density of covariance $\Sigma^{y}\sim \mathcal{W}^{-1}\left(\mathbf{S}^{y}, \nu^{y}\right)$ and mean $\mu^{y}\sim \mathcal{N}\left(\mathbf{m}^{y}, \frac{\Sigma^{y}}{\kappa^{y}}\right)$.

For an LDA classifier defined by:
\begin{equation}
\psi_{n}\left(x \right )=\left\{\begin{matrix}
0, &\mbox{if }g\left(x \right )\leq0 \\ 
1, &otherwise 
\end{matrix}\right.
\end{equation}
where $g\left(x\right)=\mathbf{a}^{T}x+b$ with $\mathbf{a}$ and $b$ are a constant vector and a constant scalar, respectively, the standard BEE has been derived in closed-form as:
\begin{equation}
E_{\pi^{*}}\left[\varepsilon _{n}^{y}\right]=\frac{1}{2}+\frac{\mbox{sgn}\left(A\right)}{2}\mathcal{I}\left(\frac{A^{2}}{A^{2}+{\mathbf{a}}^{T}\mathbf{S}^{y}\mathbf{a}};\frac{1}{2},\frac{\nu^{y}-d+1}{2}\right),
\end{equation}
where $\mbox{sgn}\left(\cdot\right)$ is the sign function,
\begin{equation}
    A=\left(-1\right)^{y}g\left(\mathbf{m}^{y}\right)\sqrt{\frac{\kappa^{y}}{1+\kappa^{y}}},
\end{equation}
and $\mathcal{I}\left(\cdot;\cdot,\cdot\right)$ denotes the regularized incomplete beta function given by
\begin{equation}
    \mathcal{I}\left(x;a,b\right)=\frac{\Gamma\left(a+b\right)}{\Gamma\left(a\right)\Gamma\left(b\right)}\int_{0}^{x}t^{a-1}\left(1-t\right)^{b-1}dt,
\end{equation}
with $\Gamma\left(\cdot\right)$ being the regular univariate gamma function.

In our TL Bayesian framework we have: $\Phi^{*}= \pi^{*}\left(\mu_{t}^{y}, \Lambda_{t}^{y}\right)=\pi^{*}\left(\mu_{t}^{y}| \Lambda_{t}^{y}\right)\pi^{*}\left(\Lambda_{t}^{y}\right)$.

Clearly, $\Phi^{*}$ is a Gaussian Wishart posterior density of model parameters $\left(\mu_{t}^{y}, \Lambda_{t}^{y}\right)$ with a precision matrix $\Lambda_{t}^{y} \sim \mathcal{W}\left(\mathbf{M}_{t,n}^{y}, \nu^{y}+n_{t}^{y}\right)$ and a mean $\mu_{t}^{y}\sim \mathcal{N}\left(\mathbf{m}_{t,n}^{y}, \left(\kappa_{t,n}^{y}\Lambda_{t}^{y}\right)^{-1}\right)$.

After making the change of variable $\Sigma_{t}^{y}=\left(\Lambda_{t}^{y}\right)^{-1}$ we get:

$\Sigma_{t}^{y} \sim \mathcal{W}\left(\left[\mathbf{M}_{t,n}^{y}\right]^{-1}, \nu^{y}+n_{t}^{y}\right)$ and a mean $\mu_{t}^{y}\sim \mathcal{N}\left(\mathbf{m}_{t,n}^{y}, \frac{\Sigma_{t}^{y}}{\kappa_{t,n}^{y}}\right)$.
By analogy, the rest of the derivation follows from~\citep{Dalton2011_B}.

\setcounter{equation}{0}
\section{Additional results for linear classifiers}
In Fig.~\ref{fig:ldans} we consider an LDA classifier and we investigate the behavior of the TL-based BEE when the target data are fixed while we vary the size of the source data. We show in the left column results for $d=2$, in the middle column results for $d=3$, and in the right column results for $d=5$. The rows correspond to results for different sizes of target datasets. Obviously, The MSE has similar trends across the three dimensions as compared to the QDA results in the main text. The deviation from the true error significantly decreases when highly related source data are employed.
\input{figures/LDA/LDA_n_s}

We show in Fig.~\ref{fig:ldant} the MSE deviation with respect to the size of target samples for dimensions 2, 3, and 5.
First column corresponds to results when using source datasets of size $n_{s}=50$ and second row shows results for $n_{s}=200$. In addition to the enhanced performance of the TL-based BEE estimator with the increasing availability of target data, we can clearly see that the MSE deviation from the true error asymptotically converges to comparable values. This convergence is observed for all relatedness levels. These results match the observed behaviour with QDA.
\input{figures/LDA/LDA_n_t}
\end{appendices}


\end{document}

%% file: figures/flow_chart.tex
\begin{figure}[h!]
\tikzstyle{startstop} = [rectangle, rounded corners, minimum width=3cm, minimum height=1cm, align=center, draw=black, fill=green!10]
\tikzstyle{figure} = [rectangle, rounded corners, minimum width=1cm, minimum height=1cm, align=center, draw=black, fill=blue!10]
\tikzstyle{io} = [trapezium, trapezium left angle=70, trapezium right angle=110, minimum width=3cm, minimum height=1cm, text centered, draw=black, fill=blue!30]
\tikzstyle{process} = [rectangle, rounded corners, minimum width=3cm, minimum height=1cm, text centered, draw=black, fill=orange!30]
\tikzstyle{decision} = [diamond, rounded corners, minimum width=0cm, minimum height=0cm, text centered, draw=black, fill=yellow!10]
\tikzstyle{parameter} = [ellipse, minimum width=1cm, minimum height=1cm, text centered, draw=black, fill=red!10]
\tikzstyle{data} = [cloud, minimum width=1cm, minimum height=1cm, text centered, draw=black, fill=blue!10]
\tikzstyle{arrow} = [thick,->,>=stealth]
\tikzstyle{line} = [draw, -latex']
\tikzstyle{empty} = [ellipse, minimum width=0cm, minimum height=0cm, text centered, draw=white, fill=white!0]
\begin{center}
\begin{adjustbox}{width=.55\linewidth}
\begin{tikzpicture}[node distance=1.5 cm, scale=0.8]
\node (L_0_N_0) [startstop] {
\textbf{Hyperparameters}\\
$d$, $Y$, $\mathbf{m}_{z}$, $\kappa_{z}$, $\nu$, $\mathbf{M}$, $\left|\alpha\right|$\\
$N_{p}$, $N_{d}$, $N_{t}$, $N_{s}$, $\vartheta$, $\tau$

};

\node (L_1_N_0) [parameter, below of=L_0_N_0] {$\Lambda^{y}$};

\node (L_2_N_0) [parameter, below of=L_1_N_0, xshift=3.5cm]{$\mu_{s}^{y}$};
\node (L_2_N_1) [parameter, left of=L_2_N_0, xshift=-0.5cm]{$\Lambda_{s}^{y}$};
\node (L_2_N_2) [parameter, left of=L_2_N_1, xshift=-1.5cm]{$\Lambda_{t}^{y}$};
\node (L_2_N_3) [parameter, left of=L_2_N_2, xshift=-0.5cm]{$\mu_{t}^{y}$};

\begin{scope}[on background layer]
\node (B_L_2_N_0)[fit=(L_2_N_0)(L_2_N_1), draw, dotted, thick, blue, fill=black!3]{};
\end{scope}
\begin{scope}[on background layer]
\node (B_L_2_N_1) [fit=(L_2_N_2)(L_2_N_3), draw, dotted, thick, blue, fill=black!3]{};
\end{scope}

\node (L_3_N_0)[data, below of=L_2_N_0]{$\mathcal{D}_{s}^{y}$};
\node (L_3_N_1)[data, below of=L_2_N_1]{$\mathcal{D}_{t}^{y}$};
\node (L_3_N_2)[data, below of=B_L_2_N_1]{$\mathcal{D}_{t}^{y}$};

\begin{scope}[on background layer]
\node (B_L_3_N_0) [fit=(L_3_N_0)(L_3_N_1), draw, dotted, thick, blue, fill=black!3]{};
\end{scope}

\node (L_4_N_0)[process, align=center, below of=B_L_3_N_0]{Derive\\$\Psi_{OBTL}\left ( \mu_{t}, \Lambda_{t}\right )$};
\node (L_4_N_1)[process, align=center, below of=L_3_N_2, xshift=0.5cm]{Derive\\$\Psi_{LDA}\left ( \mu_{t}, \Lambda_{t}\right )$};
\node (L_4_N_2)[process, align=center, left of=L_4_N_1, xshift=-2.5cm]{Derive\\$\Psi_{QDA}\left ( \mu_{t}, \Lambda_{t}\right )$};

\node (L_5_N_0)[process, align=center, below of=L_4_N_0]{Estimate\\$\varepsilon_{OBTL}$};
\node (L_5_N_1)[process, align=center, below of=L_4_N_1]{Estimate\\$\varepsilon_{LDA}$};
\node (L_5_N_2)[process, align=center, below of=L_4_N_2]{Estimate\\$\varepsilon_{QDA}=\varepsilon_{Bayes}$};

\begin{scope}[on background layer]
\node (B_L_4_N_0) [fit=(L_4_N_0)(L_5_N_0), draw, dotted, thick, blue, fill=black!3]{};
\end{scope}

\begin{scope}[on background layer]
\node (B_L_4_N_1) [fit=(L_4_N_1)(L_5_N_1), draw, dotted, thick, blue, fill=black!3]{};
\end{scope}

\begin{scope}[on background layer]
\node (B_L_4_N_2) [fit=(L_4_N_2)(L_5_N_2), draw, dotted, thick, blue, fill=black!3]{};
\end{scope}

\node (L_6_N_0)[process, align=center, below of=L_5_N_0]{Draw\\$\left( \mu_{t}^{y}, \Lambda_{t}^{y}\right )_{i} \sim \pi^{*} \left (\mu_{t}^{y}, \Lambda_{t}^{y} | \mathcal{D}_{t}^{y},\mathcal{D}_{s}^{y}\right )$};

\node (L_7_N_0)[process, below of=L_6_N_0]{ 
$BEE\left ( \varepsilon_{\Psi}^{y} \right ) =E_{\pi^{*}}\left [ \varepsilon_{\Psi}^{y}  \right ]$
};

\node (L_8_N_0)[process, below of=L_7_N_0]{$SE\left ( BEE\right )=\left ( BEE\left ( \varepsilon_{\Psi}\right )-\varepsilon_{\Psi}\right )^{2}$};

\node (L_9_N_0)[decision, below of=L_5_N_2, yshift=-1cm]{$\varepsilon_{Bayes}=\tau$?};

\node (L_10_N_0)[decision, below of=L_8_N_0, yshift=-1cm]{$N_{d}$ repetitions?};

\node (L_11_N_0)[process, below of=L_10_N_0, yshift=-1cm]{MSE = $\frac{1}{N_{d}}\sum_{1}^{N_{d}}\left ( BEE\left ( \varepsilon_{\Psi}\right )-\varepsilon_{\Psi}\right )_{1..N_{d}}^{2}$};

\node (L_12_N_0)[process, below of=L_11_N_0, align=center]{Repeat for $N_{s}$\\$\left [ 10, 50:50:500\right ]$};

\node (L_13_N_0)[process, below of=L_12_N_0, align=center]{Repeat for $N_{t}$\\$\left [ 5, 10:10:50\right ]$};

\node (L_14_N_0)[decision, below of=L_13_N_0, yshift=-1cm]{$N_{p}$ repetitions?};

\node (L_14_N_1)[process, left of=L_14_N_0, xshift=-6cm]{$MSE_{Avg}$ = $\frac{1}{N_{p}}\sum_{1}^{N_{p}}\left ( MSE\left ( \varepsilon_{\Psi}\right )\right )_{1..N_{p}}$};

\node (L_15_N_0)[startstop, left of=L_11_N_0, align=center, xshift=-6cm]{\textbf{Evaluate MSE}\\For each $n_{t}\in N_{t}$\\
Evaluate $MSE_{Avg}\left ( BEE \left( \varepsilon_{\Psi}\right )\right )=f\left (n_{s}\in N_{s} \right )$\\
(Display $n_{t}~\&~\varepsilon_{Bayes}$)\\
\\
For each $n_{s}\in N_{s}$\\
Evaluate $MSE_{Avg}\left ( BEE \left( \varepsilon_{\Psi}\right )\right )=f\left (n_{t}\in N_{t} \right )$\\
(Display $n_{s}~\&~\varepsilon_{Bayes}$)};

\node (L_16_N_0)[figure, above of= L_15_N_0, xshift=2cm, yshift=2.5cm]{
\begin{tikzpicture}[scale=0.6, transform shape]
  \draw[->] (0.7, 0) -- (3, 0) node[below] {$n_s$};
  \draw[->] (0.7, 0) -- (0.7, 2) node[above] {$MSE$};
  \draw[scale=1, domain=0.8:2.5, smooth, variable=\x, red, very thick] plot ({\x}, {1/(\x*\x)});
  \draw [step=0.3,gray, very thin] (0.8,0) grid (2.7,1.7);
\end{tikzpicture}
};

\node (L_16_N_1)[figure, above of= L_15_N_0, xshift=-2cm, yshift=2.5cm]{
\begin{tikzpicture}[scale=0.6, transform shape]
  \draw[->] (0.7, 0) -- (3, 0) node[below] {$n_s$};
  \draw[->] (0.7, 0) -- (0.7, 2) node[above] {$MSE$};
  \draw[scale=1, domain=0.8:2.5, smooth, variable=\x, blue, very thick] plot ({\x}, {1/(\x*\x)});
  \draw [step=0.3,gray, very thin] (0.8,0) grid (2.7,1.7);
\end{tikzpicture}
};
\draw [arrow] (L_1_N_0) -- (L_2_N_1);
\draw [arrow] (L_1_N_0) -- (L_2_N_2);

\draw [arrow] (L_2_N_1) -- (L_2_N_0);
\draw [arrow] (L_2_N_2) -- (L_2_N_3);
\draw [arrow] (L_2_N_2) -- (L_3_N_2);
\draw [arrow] (L_2_N_3) -- (L_3_N_2);
\draw [arrow] (L_2_N_0) -- (B_L_3_N_0);
\draw [arrow] (L_2_N_1) -- (B_L_3_N_0);

\draw [arrow] (B_L_3_N_0) -- (L_4_N_0);
\draw [arrow] ($(L_3_N_2.west)$) -| ($(L_5_N_2.east)+(0.65,0)$) -- ($(L_5_N_1.west)$);
\draw [arrow] ($(L_5_N_2.east)+(0.65,0)$) -- ($(L_5_N_2.east)$);

\draw [arrow] (L_4_N_0) -- (L_5_N_0);

\draw [arrow] ($(B_L_3_N_0.east)+(0,-0.5)$) -| ($(L_6_N_0.east)+(0.5,0)$) -- (L_6_N_0);

\draw [arrow] (L_5_N_2) -- (L_9_N_0);
\draw [arrow] (L_6_N_0) -- (L_7_N_0);
\draw [arrow] (L_7_N_0) -- (L_8_N_0);
\draw [arrow] (L_8_N_0) -- (L_10_N_0);
\draw [arrow] (L_11_N_0) -- (L_12_N_0);
\draw [arrow] (L_12_N_0) -- (L_13_N_0);
\draw [arrow] (L_13_N_0) -- (L_14_N_0);

\draw [arrow] (L_4_N_1) -- (L_5_N_1);
\draw [arrow] (L_4_N_2) -- (L_5_N_2);

\draw [arrow] ($(L_9_N_0.east)$) node[anchor=south west]{Yes $\rightarrow$ save $\left \{\left ( \mu_{t}^{y}, \Lambda_{t}^{y}\right )~\&~\Psi \right \}$} -| ($(B_L_3_N_0.west)+(-1.3,0)$) -- (B_L_3_N_0);

\draw [arrow] ($(L_9_N_0.west)$) node[anchor=south east]{No}-| ($(B_L_2_N_1.west)+(-4.8,0)$) --node[anchor=south]{Update $\mathbf{m}_{t}^{y}\left ( \vartheta\right )$} (B_L_2_N_1);

\draw [arrow] ($(L_10_N_0.east)$)  node[anchor=south west]{No} -| ($(B_L_3_N_0.east)+(2,0)$) -- (B_L_3_N_0);

\draw [arrow] ($(L_10_N_0.south)$) -- node[anchor=west]{Yes} (L_11_N_0);

\draw [arrow] ($(L_12_N_0.west)$) -| ($(L_13_N_0.west)+(-0.5,0)$) node[anchor=north east, align=center]{$MSE_{n_{t}}\left ( \varepsilon_{\Psi}\right )=f\left ( n_{s}\right )$:\\
$N_{t}$ loop outer\\
$N_{s}$ loop inner}-- ($(L_13_N_0.west)$);

\draw [arrow] ($(L_13_N_0.east)$) node[anchor=north west, align=center]{Otherwise:\\
$N_{t}$ loop inner\\
$N_{s}$ loop outer}-|($(L_12_N_0.east)+(0.5,0)$) -- ($(L_12_N_0.east)$);

\draw [arrow] (L_14_N_1) -- (L_15_N_0);
\draw [arrow] ($(L_14_N_0.west)$) --node[anchor=south west]{Yes} (L_14_N_1);
\draw [arrow] ($(L_14_N_0.east)$) node[anchor=south west]{No}-| ($(L_1_N_0.east)+(7.2,0)$) -- ($(L_1_N_0.east)$);

\draw [arrow] ($(B_L_2_N_1.west)+(0,-0.5)$) -| (L_4_N_2);
\draw [arrow] ($(B_L_2_N_1.south)+(1.5,0)$) |- ($(L_4_N_1.north)+(0,0.5)$) -| ($(L_4_N_1.north)$);

\draw [arrow] ($(L_15_N_0.north)+(2.5,0)$) -| (L_16_N_0);
\draw [arrow] ($(L_15_N_0.north)+(-2.5,0)$) -| (L_16_N_1);

\end{tikzpicture}
\end{adjustbox}
\end{center}\caption{Flow chart illustrating the simulation set-up based on synthetic datasets.}
\label{flow_chart}
\end{figure}
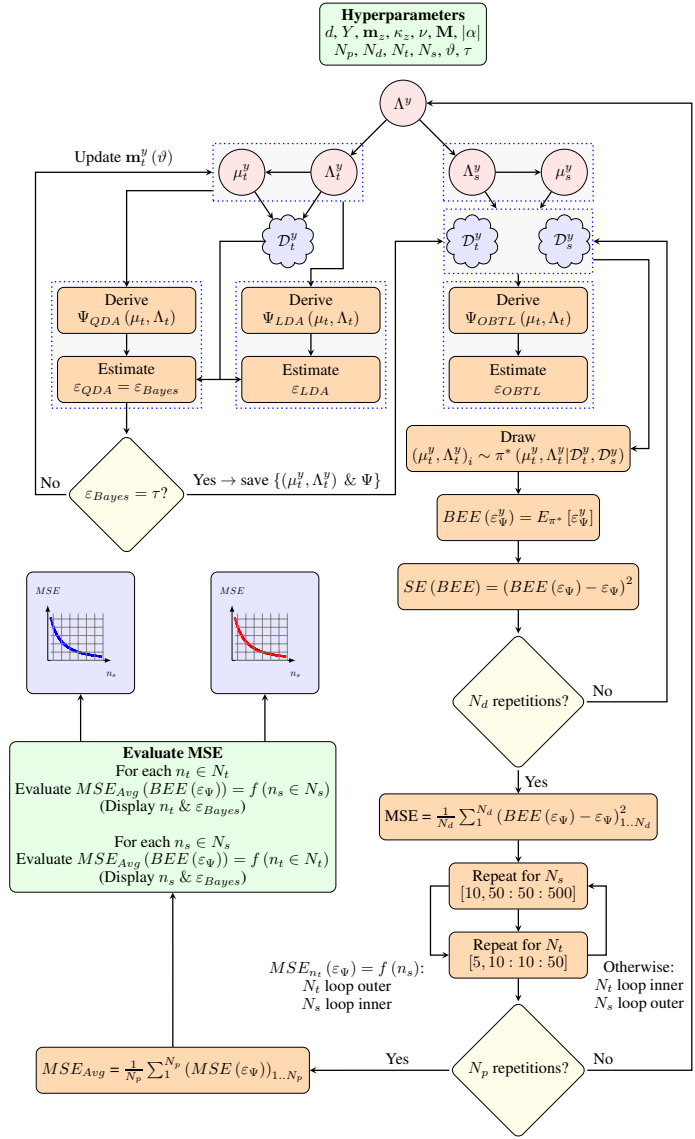

%% file: figures/QDA/QDA_n_s.tex
\begin{figure}[h!]
\begin{center}
\ref{named1}
\par
\bigskip
\begin{subfigure}[b]{0.3\textwidth}
\begin{adjustbox}{width=0.9\linewidth}
\begin{tikzpicture}
    \begin{axis}[width=1\linewidth,font=\footnotesize,cycle list name=colors_qda,
    xlabel= $n_s$, ylabel= MSE,ylabel near ticks,
    xlabel near ticks,
  xmin=10,xmax=500,xtick={10,100,200,300,400,500},scaled y ticks=base 10:4,
legend columns=3,
legend entries={$\left|\alpha\right|=0.1$,$\left|\alpha\right|=0.3$,$\left|\alpha\right|=0.5$,$\left|\alpha\right|=0.7$,$\left|\alpha\right|=0.9$,$\left|\alpha\right|=0.95$},
legend to name=named1,
    legend style={nodes=right},legend style={nodes={scale=0.8, transform shape}},
      xmajorgrids,
    grid style={dotted},
    ymajorgrids,
   ]

  \addplot plot coordinates {
(10,0.001174)(50,0.001169)(100,0.001167)(150,0.001167)(200,0.001170)(250,0.001172)(300,0.001172)(350,0.001171)(400,0.001168)(450,0.001164)(500,0.001157)
};
\addplot plot coordinates {
(10,0.001154)(50,0.001138)(100,0.001126)(150,0.001120)(200,0.001118)(250,0.001116)(300,0.001114)(350,0.001112)(400,0.001111)(450,0.001111)(500,0.001112)
};
     \addplot plot coordinates {
(10,0.001131)(50,0.001081)(100,0.001046)(150,0.001030)(200,0.001025)(250,0.001024)(300,0.001020)(350,0.001014)(400,0.001008)(450,0.001006)(500,0.001006)
};
    \addplot plot coordinates {
(10,0.001109)(50,0.001009)(100,0.000939)(150,0.000902)(200,0.000890)(250,0.000887)(300,0.000884)(350,0.000878)(400,0.000873)(450,0.000869)(500,0.000866)
};
     \addplot plot coordinates {
(10,0.001081)(50,0.000871)(100,0.000729)(150,0.000651)(200,0.000621)(250,0.000611)(300,0.000601)(350,0.000587)(400,0.000571)(450,0.000558)(500,0.000548)
};
      \addplot  plot coordinates {
(10,0.001073)(50,0.000821)(100,0.000639)(150,0.000541)(200,0.000503)(250,0.000491)(300,0.000477)(350,0.000455)(400,0.000433)(450,0.000417)(500,0.000405)
};
\end{axis}
\end{tikzpicture}
\end{adjustbox}
\caption{$d=2$, $n_t=20$}
\label{fig:qdansa}
\end{subfigure}
\begin{subfigure}[b]{0.3\textwidth}
\begin{adjustbox}{width=0.9\linewidth}
\begin{tikzpicture}
\begin{axis}[width=1\linewidth,font=\footnotesize,cycle list name=colors_qda,
    xlabel= $n_s$, ylabel= MSE,
  xmin=10,xmax=500,xtick={10,100,200,300,400,500},ylabel near ticks,ymax=6.5e-4,
  yticklabel style={
/pgf/number format/fixed,
/pgf/number format/fixed zerofill,
/pgf/number format/precision=1
},
    xlabel near ticks,
    legend style={nodes=right},legend style={nodes={scale=1, transform shape}},
      xmajorgrids,
    grid style={dotted},
    ymajorgrids,
   ]
  \addplot plot coordinates{
(10,0.000629)(50,0.000630)(100,0.000630)(150,0.000630)(200,0.000630)(250,0.000630)(300,0.000630)(350,0.000630)(400,0.000631)(450,0.000631)(500,0.000632)
};
 \addplot plot coordinates{
(10,0.000623)(50,0.000618)(100,0.000615)(150,0.000613)(200,0.000613)(250,0.000614)(300,0.000615)(350,0.000615)(400,0.000613)(450,0.000610)(500,0.000606)
};
 \addplot plot coordinates{
(10,0.000619)(50,0.000607)(100,0.000597)(150,0.000590)(200,0.000586)(250,0.000584)(300,0.000583)(350,0.000584)(400,0.000585)(450,0.000584)(500,0.000583)
};
 \addplot plot coordinates{
(10,0.000600)(50,0.000585)(100,0.000573)(150,0.000563)(200,0.000555)(250,0.000550)(300,0.000546)(350,0.000544)(400,0.000542)(450,0.000541)(500,0.000540)
};
 \addplot plot coordinates{
(10,0.000598)(50,0.000570)(100,0.000546)(150,0.000526)(200,0.000509)(250,0.000496)(300,0.000485)(350,0.000477)(400,0.000471)(450,0.000468)(500,0.000468)
};
 \addplot plot coordinates{
(10,0.000596)(50,0.000555)(100,0.000519)(150,0.000488)(200,0.000463)(250,0.000441)(300,0.000423)(350,0.000410)(400,0.000400)(450,0.000396)(500,0.000397)
};
\end{axis}
\end{tikzpicture}
\end{adjustbox}
\caption{$d=2$, $n_t=50$}
\label{fig:qdansd}
\end{subfigure}
\par\medskip 
\begin{subfigure}[b]{0.3\textwidth}
\begin{adjustbox}{width=0.9\linewidth}
\begin{tikzpicture}
    \begin{axis}[ width=1\linewidth,font=\footnotesize,cycle list name=colors_qda,
    xlabel= $n_s$, ylabel= MSE,ylabel near ticks,scaled y ticks=base 10:4,
    xlabel near ticks,
  xmin=10,xmax=500,xtick={10,100,200,300,400,500},
      xmajorgrids,
    grid style={dotted},
    ymajorgrids,
   ]

  \addplot plot coordinates {
(10,0.002067)(50,0.002064)(100,0.002063)(150,0.002064)(200,0.002068)(250,0.002073)(300,0.002076)(350,0.002075)(400,0.002073)(450,0.002068)(500,0.002063)
};
\addplot plot coordinates {
(10,0.001980)(50,0.001979)(100,0.001980)(150,0.001983)(200,0.001987)(250,0.001994)(300,0.002001)(350,0.002005)(400,0.002006)(450,0.002002)(500,0.001993)
};
     \addplot plot coordinates {
(10,0.001884)(50,0.001886)(100,0.001891)(150,0.001900)(200,0.001911)(250,0.001918)(300,0.001920)(350,0.001919)(400,0.001916)(450,0.001912)(500,0.001909)
};
    \addplot plot coordinates {
(10,0.001736)(50,0.001720)(100,0.001713)(150,0.001717)(200,0.001725)(250,0.001731)(300,0.001731)(350,0.001727)(400,0.001721)(450,0.001715)(500,0.001707)
};
     \addplot plot coordinates {
(10,0.001506)(50,0.001337)(100,0.001228)(150,0.001183)(200,0.001182)(250,0.001190)(300,0.001185)(350,0.001168)(400,0.001148)(450,0.001134)(500,0.001124)
};
      \addplot  plot coordinates {
(10,0.001365)(50,0.001118)(100,0.000952)(150,0.000874)(200,0.000858)(250,0.000861)(300,0.000856)(350,0.000841)(400,0.000822)(450,0.000806)(500,0.000792)
};
\end{axis}
\end{tikzpicture}
\end{adjustbox}
\caption{$d=3$, $n_t=20$}
\label{fig:qdansb}
\end{subfigure}
\begin{subfigure}[b]{0.3\textwidth}
\begin{adjustbox}{width=0.9\linewidth}
\begin{tikzpicture}
\begin{axis}[width=1\linewidth,font=\footnotesize,cycle list name=colors_qda,
    xlabel= $n_s$, ylabel= MSE,
  xmin=10,xmax=500,xtick={10,100,200,300,400,500},ylabel near ticks,
    xlabel near ticks,
    yticklabel style={
/pgf/number format/fixed,
/pgf/number format/fixed zerofill,
/pgf/number format/precision=1
},
legend style={nodes=right},legend style={nodes={scale=1, transform shape}},
      xmajorgrids,
    grid style={dotted},
    ymajorgrids,
   ]
  \addplot plot coordinates{
(10,0.000840)(50,0.000837)(100,0.000835)(150,0.000834)(200,0.000834)(250,0.000833)(300,0.000832)(350,0.000831)(400,0.000829)(450,0.000826)(500,0.000823)
};
 \addplot plot coordinates{
(10,0.000816)(50,0.000811)(100,0.000807)(150,0.000804)(200,0.000803)(250,0.000803)(300,0.000803)(350,0.000802)(400,0.000802)(450,0.000799)(500,0.000796)
};
 \addplot plot coordinates{
(10,0.000801)(50,0.000800)(100,0.000798)(150,0.000795)(200,0.000793)(250,0.000792)(300,0.000792)(350,0.000792)(400,0.000789)(450,0.000782)(500,0.000771)
};
 \addplot plot coordinates{
(10,0.000777)(50,0.000769)(100,0.000763)(150,0.000760)(200,0.000758)(250,0.000757)(300,0.000755)(350,0.000753)(400,0.000749)(450,0.000739)(500,0.000726)
};
 \addplot plot coordinates{
(10,0.000682)(50,0.000674)(100,0.000667)(150,0.000663)(200,0.000660)(250,0.000654)(300,0.000646)(350,0.000640)(400,0.000634)(450,0.000630)(500,0.000628)
};
 \addplot plot coordinates{
(10,0.000639)(50,0.000611)(100,0.000590)(150,0.000576)(200,0.000566)(250,0.000557)(300,0.000545)(350,0.000533)(400,0.000523)(450,0.000517)(500,0.000514)
};
\end{axis}
\end{tikzpicture}
\end{adjustbox}
\caption{$d=3$, $n_t=50$}
\label{fig:qdanse}
\end{subfigure}
\par\medskip 
\begin{subfigure}[b]{0.3\textwidth}
\begin{adjustbox}{width=0.9\linewidth}
\begin{tikzpicture}
\begin{axis}[ width=1\linewidth,font=\footnotesize,cycle list name=colors_qda,
    xlabel= $n_s$, ylabel= MSE,
  xmin=10,xmax=500,xtick={10,100,200,300,400,500},ylabel near ticks,scaled y ticks=base 10:3,
    xlabel near ticks,
    legend style={nodes=right},legend style={nodes={scale=1, transform shape}},
      xmajorgrids,
    grid style={dotted},
    ymajorgrids,
   ]
  \addplot plot coordinates {
(10,0.027273)(50,0.027310)(100,0.027315)(150,0.027279)(200,0.027239)(250,0.027246)(300,0.027309)(350,0.027389)(400,0.027429)(450,0.027398)(500,0.027305)
};
  \addplot plot coordinates {
(10,0.026882)(50,0.026983)(100,0.027047)(150,0.027068)(200,0.027064)(250,0.027064)(300,0.027084)(350,0.027120)(400,0.027158)(450,0.027184)(500,0.027202)
};
  \addplot plot coordinates {
(10,0.025743)(50,0.025047)(100,0.024601)(150,0.024424)(200,0.024447)(250,0.024561)(300,0.024686)(350,0.024789)(400,0.024855)(450,0.024881)(500,0.024867)
};
  \addplot plot coordinates {
(10,0.023823)(50,0.021118)(100,0.019244)(150,0.018262)(200,0.017962)(250,0.018013)(300,0.018164)(350,0.018301)(400,0.018392)(450,0.018427)(500,0.018412)
};
  \addplot plot coordinates {
(10,0.021627)(50,0.015329)(100,0.010853)(150,0.008327)(200,0.007303)(250,0.007064)(300,0.007062)(350,0.007070)(400,0.007062)(450,0.007056)(500,0.007052)
};
  \addplot plot coordinates {
(10,0.021430)(50,0.013943)(100,0.008608)(150,0.005580)(200,0.004324)(250,0.003986)(300,0.003910)(350,0.003834)(400,0.003747)(450,0.003689)(500,0.003652)
};
\end{axis}
\end{tikzpicture}
\end{adjustbox}
\caption{$d=5$, $n_t=20$}
\label{fig:qdansc}
\end{subfigure}
\begin{subfigure}[b]{0.3\textwidth}
\begin{adjustbox}{width=0.9\linewidth}
\begin{tikzpicture}
\begin{axis}[ width=1\linewidth,font=\footnotesize,cycle list name=colors_qda,
    xlabel= $n_s$, ylabel= MSE,
  xmin=10,xmax=500,xtick={10,100,200,300,400,500},ylabel near ticks,scaled y ticks=base 10:3,
    xlabel near ticks,
    legend style={nodes=right},legend style={nodes={scale=1, transform shape}},
      xmajorgrids,
    grid style={dotted},
    ymajorgrids,
   ]
 \addplot plot coordinates{
(10,0.011504)(50,0.011496)(100,0.011496)(150,0.011505)(200,0.011514)(250,0.011511)(300,0.011498)(350,0.011489)(400,0.011499)(450,0.011533)(500,0.011589)
};
 \addplot plot coordinates{
(10,0.011320)(50,0.011371)(100,0.011422)(150,0.011475)(200,0.011522)(250,0.011551)(300,0.011560)(350,0.011562)(400,0.011577)(450,0.011619)(500,0.011684)
};
 \addplot plot coordinates{
(10,0.010834)(50,0.010780)(100,0.010763)(150,0.010788)(200,0.010841)(250,0.010904)(300,0.010966)(350,0.011019)(400,0.011057)(450,0.011073)(500,0.011069)
};
 \addplot plot coordinates{
(10,0.010313)(50,0.009662)(100,0.009214)(150,0.008982)(200,0.008919)(250,0.008953)(300,0.009030)(350,0.009119)(400,0.009195)(450,0.009243)(500,0.009266)
};
 \addplot plot coordinates{
(10,0.009997)(50,0.007805)(100,0.006285)(150,0.005397)(200,0.005014)(250,0.004926)(300,0.004957)(350,0.005009)(400,0.005043)(450,0.005049)(500,0.005032)
};
 \addplot plot coordinates{
(10,0.009920)(50,0.007255)(100,0.005253)(150,0.004034)(200,0.003447)(250,0.003235)(300,0.003184)(350,0.003177)(400,0.003171)(450,0.003157)(500,0.003140)
};
\end{axis}
\end{tikzpicture}
\end{adjustbox}
\caption{$d=5$, $n_t=50$}
\label{fig:qdansf}
\end{subfigure}

\caption{MSE deviation from true error for Gaussian distributions with respect to source sample size. The Bayes error is fixed at 0.2 in all figures.}
\label{fig:Figure1}
\end{center}
\end{figure}
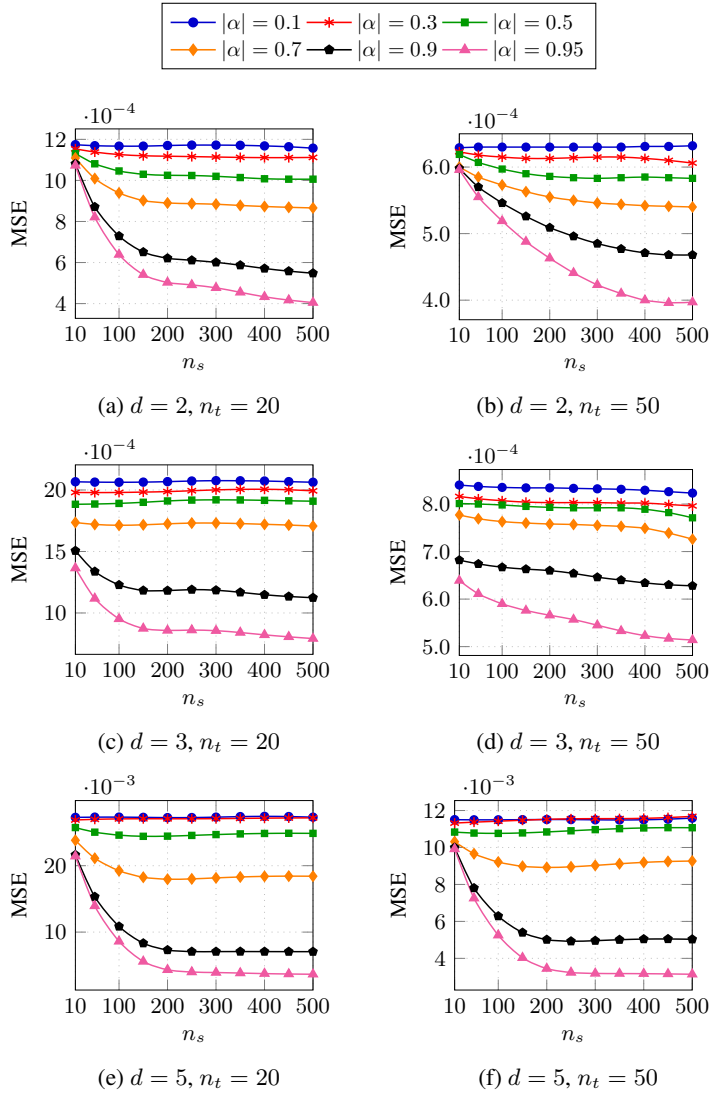

%% file: figures/QDA/QDA_n_t.tex
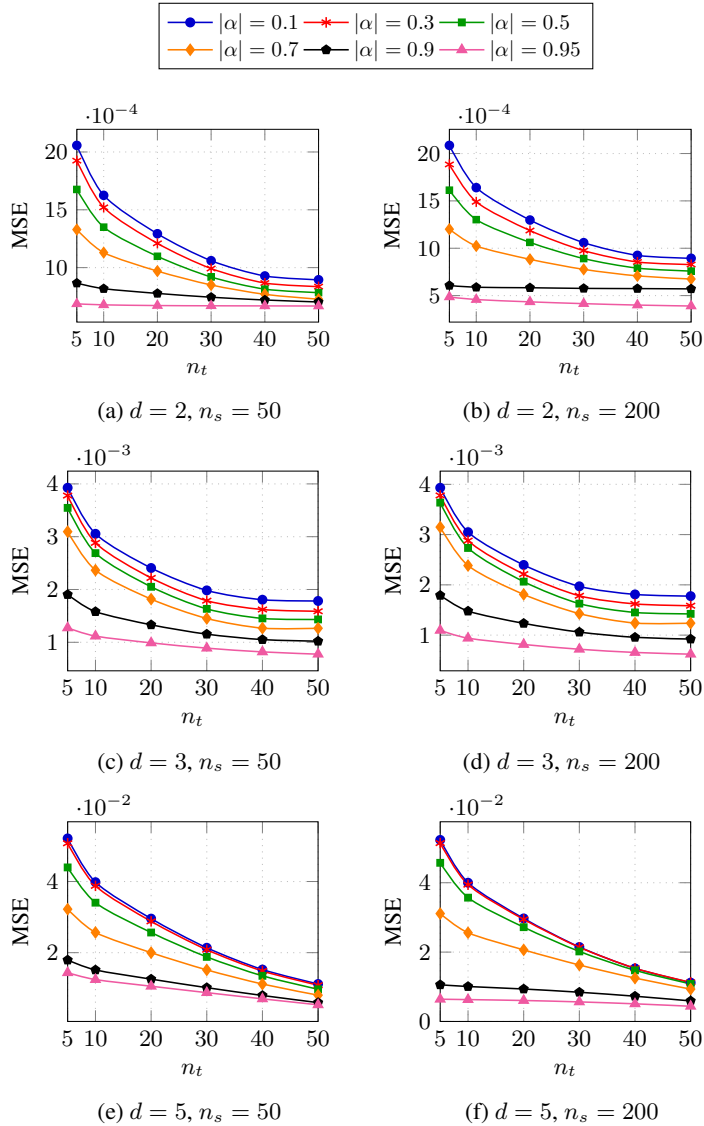
\begin{figure}[h!]
\begin{center}
\ref{named2}
\par
\bigskip
\begin{subfigure}[b]{0.3\textwidth}
\begin{adjustbox}{width=0.9\linewidth}
\begin{tikzpicture}
    \begin{axis}[ width=1\linewidth,font=\footnotesize,cycle list name=colors_qda,
    xlabel= $n_t$, ylabel= MSE, ylabel near ticks,
    xlabel near ticks,
  xmin=5,xmax=50,xtick={5,10,20,30,40,50}, scaled y ticks=base 10:4,
legend columns=3,
legend entries={$\left|\alpha\right|=0.1$,$\left|\alpha\right|=0.3$,$\left|\alpha\right|=0.5$,$\left|\alpha\right|=0.7$,$\left|\alpha\right|=0.9$,$\left|\alpha\right|=0.95$},
legend to name=named2,
    legend style={nodes=right},legend style={nodes={scale=0.8, transform shape}},
      xmajorgrids,
    grid style={dotted},
    ymajorgrids,
   ]

  \addplot plot coordinates {
(5,0.002057)(10,0.001625)(20,0.001293)(30,0.001060)(40,0.000928)(50,0.000895)
};
\addplot plot coordinates {
(5,0.001925)(10,0.001521)(20,0.001210)(30,0.000992)(40,0.000867)(50,0.000836)
};
     \addplot plot coordinates {
(5,0.001676)(10,0.001350)(20,0.001098)(30,0.000919)(40,0.000814)(50,0.000784)
};
    \addplot plot coordinates {
(5,0.001329)(10,0.001130)(20,0.000970)(30,0.000850)(40,0.000769)(50,0.000727)
};
     \addplot plot coordinates {
(5,0.000866)(10,0.000818)(20,0.000777)(30,0.000744)(40,0.000720)(50,0.000703)
};
      \addplot  plot coordinates {
(5,0.000687)(10,0.000678)(20,0.000672)(30,0.000670)(40,0.000669)(50,0.000668)
};
\end{axis}
\end{tikzpicture}
\end{adjustbox}
\caption{$d=2$, $n_s=50$}
\label{fig:qdanta}
\end{subfigure}
\begin{subfigure}[b]{0.3\textwidth}
\begin{adjustbox}{width=0.9\linewidth}
\begin{tikzpicture}
\begin{axis}[width=1\linewidth,font=\footnotesize,cycle list name=colors_qda,
    xlabel= $n_t$, ylabel= MSE, xmin=5,xmax=50,xtick={5,10,20,30,40,50},ylabel near ticks,  scaled y ticks=base 10:4,
    xlabel near ticks,
    legend style={nodes=right},legend style={nodes={scale=1, transform shape}},
      xmajorgrids,
    grid style={dotted},
    ymajorgrids,
   ]
  \addplot plot coordinates{
(5,0.002086)(10,0.001640)(20,0.001298)(30,0.001059)(40,0.000925)(50,0.000894)
};
 \addplot plot coordinates{
(5,0.001885)(10,0.001490)(20,0.001187)(30,0.000975)(40,0.000855)(50,0.000828)
};
 \addplot plot coordinates{
(5,0.001613)(10,0.001302)(20,0.001062)(30,0.000891)(40,0.000790)(50,0.000759)
};
 \addplot plot coordinates{
(5,0.001201)(10,0.001024)(20,0.000883)(30,0.000777)(40,0.000708)(50,0.000675)
};
 \addplot plot coordinates{
(5,0.000607)(10,0.000588)(20,0.000583)(30,0.000576)(40,0.000574)(50,0.000571)
};
 \addplot plot coordinates{
(5,0.000486)(10,0.000458)(20,0.000434)(30,0.000415)(40,0.000400)(50,0.000390)
};
\end{axis}
\end{tikzpicture}
\end{adjustbox}
\caption{$d=2$, $n_s=200$}
\label{fig:qdantd}
\end{subfigure}
\par\medskip 
\begin{subfigure}[b]{0.3\textwidth}
\begin{adjustbox}{width=0.9\linewidth}
\begin{tikzpicture}
    \begin{axis}[width=1\linewidth,font=\footnotesize,cycle list name=colors_qda,
    xlabel= $n_t$, ylabel= MSE, ylabel near ticks,
    xlabel near ticks,
  xmin=5,xmax=50,xtick={5,10,20,30,40,50},
      xmajorgrids,
    grid style={dotted},
    ymajorgrids,
   ]

  \addplot plot coordinates {
(5,0.003927)(10,0.003055)(20,0.002407)(30,0.001983)(40,0.001806)(50,0.001783)
};
\addplot plot coordinates {
(5,0.003779)(10,0.002883)(20,0.002219)(30,0.001788)(40,0.001621)(50,0.001588)
};
     \addplot plot coordinates {
(5,0.003546)(10,0.002689)(20,0.002051)(30,0.001633)(40,0.001452)(50,0.001433)
};
    \addplot plot coordinates {
(5,0.003095)(10,0.002365)(20,0.001818)(30,0.001452)(40,0.001268)(50,0.001266)
};
     \addplot plot coordinates {
(5,0.001904)(10,0.001581)(20,0.001331)(30,0.001154)(40,0.001051)(50,0.001021)
};
      \addplot  plot coordinates {
(5,0.001273)(10,0.001117)(20,0.000989)(30,0.000889)(40,0.000818)(50,0.000775)
};
\end{axis}
\end{tikzpicture}
\end{adjustbox}
\caption{$d=3$, $n_s=50$}
\label{fig:qdantb}
\end{subfigure}
\begin{subfigure}[b]{0.3\textwidth}
\begin{adjustbox}{width=0.9\linewidth}
\begin{tikzpicture}
\begin{axis}[width=1\linewidth,font=\footnotesize,cycle list name=colors_qda,
    xlabel= $n_t$, ylabel= MSE, xmin=5,xmax=50,xtick={5,10,20,30,40,50},ylabel near ticks,
    xlabel near ticks,
legend style={nodes=right},legend style={nodes={scale=1, transform shape}},
      xmajorgrids,
    grid style={dotted},
    ymajorgrids,
   ]
  \addplot plot coordinates{
(5,0.003932)(10,0.003049)(20,0.002396)(30,0.001971)(40,0.001810)(50,0.001776)
};
 \addplot plot coordinates{
(5,0.003779)(10,0.002880)(20,0.002214)(30,0.001783)(40,0.001621)(50,0.001585)
};
 \addplot plot coordinates{
(5,0.003636)(10,0.002734)(20,0.002064)(30,0.001627)(40,0.001451)(50,0.001423)
};
 \addplot plot coordinates{
(5,0.003149)(10,0.002384)(20,0.001810)(30,0.001428)(40,0.001240)(50,0.001238)
};
 \addplot plot coordinates{
(5,0.001791)(10,0.001478)(20,0.001234)(30,0.001060)(40,0.000956)(50,0.000922)
};
 \addplot plot coordinates{
(5,0.001095)(10,0.000939)(20,0.000814)(30,0.000719)(40,0.000655)(50,0.000621)
};
\end{axis}
\end{tikzpicture}
\end{adjustbox}
\caption{$d=3$, $n_s=200$}
\label{fig:qdante}
\end{subfigure}
\par\medskip 
\begin{subfigure}[b]{0.3\textwidth}
\begin{adjustbox}{width=0.9\linewidth}
\begin{tikzpicture}
\begin{axis}[ width=1\linewidth,font=\footnotesize,cycle list name=colors_qda,
    xlabel= $n_t$, ylabel= MSE, xmin=5,xmax=50,xtick={5,10,20,30,40,50},ylabel near ticks,
    xlabel near ticks,
    legend style={nodes=right},legend style={nodes={scale=1, transform shape}},
      xmajorgrids,
    grid style={dotted},
    ymajorgrids,
   ]
  \addplot plot coordinates {
(5,0.052118)(10,0.039817)(20,0.029572)(30,0.021380)(40,0.015256)(50,0.011199)
};
  \addplot plot coordinates {
(5,0.050743)(10,0.038780)(20,0.028808)(30,0.020825)(40,0.014845)(50,0.010866)
};
  \addplot plot coordinates {
(5,0.043964)(10,0.034051)(20,0.025668)(30,0.018810)(40,0.013492)(50,0.009711)
};
  \addplot plot coordinates {
(5,0.032224)(10,0.025686)(20,0.019999)(30,0.015160)(40,0.011181)(50,0.008058)
};
  \addplot plot coordinates {
(5,0.017928)(10,0.015149)(20,0.012562)(30,0.010166)(40,0.007966)(50,0.005962)
};
  \addplot plot coordinates {
(5,0.014369)(10,0.012438)(20,0.010570)(30,0.008763)(40,0.007022)(50,0.005346)
};
\end{axis}
\end{tikzpicture}
\end{adjustbox}
\caption{$d=5$, $n_s=50$}
\label{fig:qdantc}
\end{subfigure}
\begin{subfigure}[b]{0.3\textwidth}
\begin{adjustbox}{width=0.9\linewidth}
\begin{tikzpicture}
\begin{axis}[width=1\linewidth,font=\footnotesize,cycle list name=colors_qda,
    xlabel= $n_t$, ylabel= MSE, xmin=5,xmax=50,xtick={5,10,20,30,40,50},ymin=0,ylabel near ticks,
    xlabel near ticks,
    legend style={nodes=right},legend style={nodes={scale=1, transform shape}},
      xmajorgrids,
    grid style={dotted},
    ymajorgrids,
   ]
 \addplot plot coordinates{
(5,0.052346)(10,0.040001)(20,0.029712)(30,0.021474)(40,0.015304)(50,0.011198)
};
 \addplot plot coordinates{
(5,0.051435)(10,0.039448)(20,0.029436)(30,0.021394)(40,0.015338)(50,0.011265)
};
 \addplot plot coordinates{
(5,0.045741)(10,0.035698)(20,0.027184)(30,0.020194)(40,0.014744)(50,0.010830)
};
 \addplot plot coordinates{
(5,0.031099)(10,0.025579)(20,0.020640)(30,0.016281)(40,0.012511)(50,0.009329)
};
 \addplot plot coordinates{
(5,0.010578)(10,0.010074)(20,0.009357)(30,0.008424)(40,0.007279)(50,0.005922)
};
 \addplot plot coordinates{
(5,0.006410)(10,0.006302)(20,0.006045)(30,0.005639)(40,0.005085)(50,0.004382)
};
\end{axis}
\end{tikzpicture}
\end{adjustbox}
\caption{$d=5$, $n_s=200$}
\label{fig:qdantf}
\end{subfigure}

\caption{MSE deviation from true error for Gaussian distributions with respect to target sample size. The Bayes error is fixed at 0.2 in all figures.}
\label{fig:Figure2}
\end{center}
\end{figure}

%% file: figures/QDA/QDA_n_s_Bayes.tex
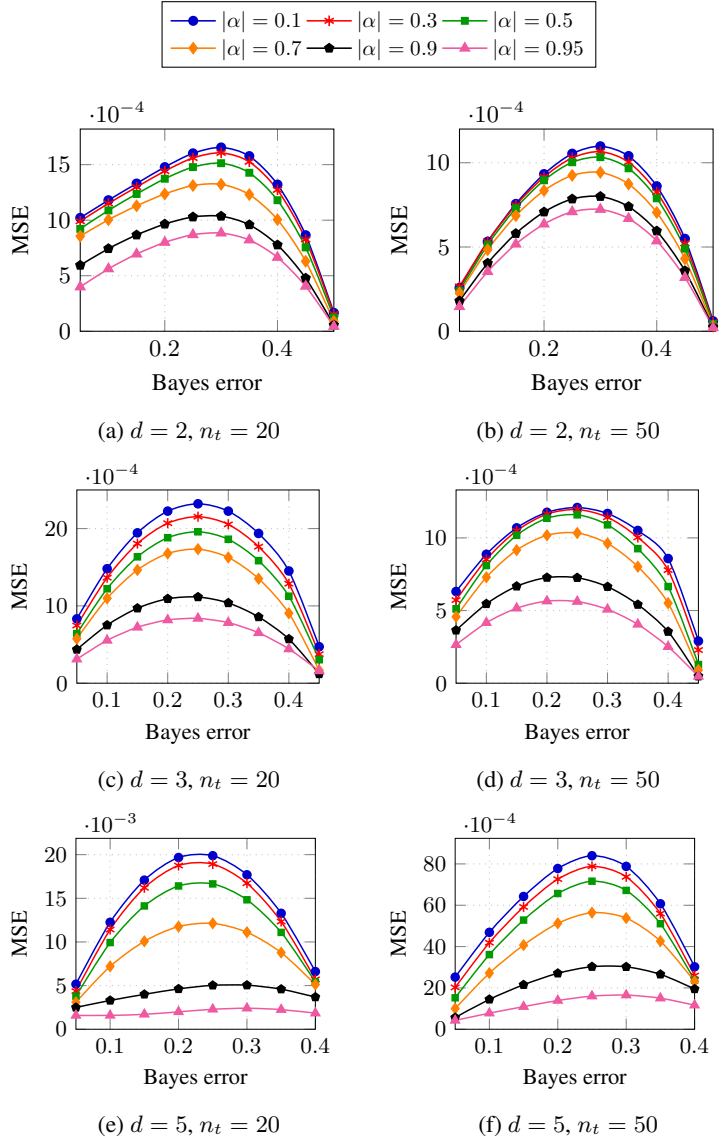
\begin{figure}[h!]
\begin{center}
\ref{named3}
\par
\bigskip
\begin{subfigure}[b]{0.3\textwidth}
\begin{adjustbox}{width=0.9\linewidth}
\begin{tikzpicture}
    \begin{axis}[ width=1\linewidth,font=\footnotesize,cycle list name=colors_qda,
    xlabel= Bayes error, ylabel= MSE,ylabel near ticks,
    xlabel near ticks,
  xmin=5.000000e-02,xmax=5.000000e-01,ymin=0, 
  scaled y ticks=base 10:4,
legend columns=3,
legend entries={$\left|\alpha\right|=0.1$,$\left|\alpha\right|=0.3$,$\left|\alpha\right|=0.5$,$\left|\alpha\right|=0.7$,$\left|\alpha\right|=0.9$,$\left|\alpha\right|=0.95$},
legend to name=named3,
    legend style={nodes=right},legend style={nodes={scale=0.8, transform shape}},
      xmajorgrids,
    grid style={dotted},
    ymajorgrids,
   ]

  \addplot plot coordinates {
(5.000000e-02,0.001021)(1.000000e-01,0.001180)(1.500000e-01,0.001330)(2.000000e-01,0.001478)(2.500000e-01,0.001602)(3.000000e-01,0.001655)(3.500000e-01,0.001577)(4.000000e-01,0.001321)(4.500000e-01,0.000865)(5.000000e-01,0.000166)
};
\addplot plot coordinates {
(5.000000e-02,0.000991)(1.000000e-01,0.001151)(1.500000e-01,0.001298)(2.000000e-01,0.001442)(2.500000e-01,0.001560)(3.000000e-01,0.001606)(3.500000e-01,0.001526)(4.000000e-01,0.001273)(4.500000e-01,0.000827)(5.000000e-01,0.000155)
};
     \addplot plot coordinates {
(5.000000e-02,0.000922)(1.000000e-01,0.001088)(1.500000e-01,0.001236)(2.000000e-01,0.001373)(2.500000e-01,0.001478)(3.000000e-01,0.001513)(3.500000e-01,0.001427)(4.000000e-01,0.001180)(4.500000e-01,0.000755)(5.000000e-01,0.000125)
};
    \addplot plot coordinates {
(5.000000e-02,0.000856)(1.000000e-01,0.001004)(1.500000e-01,0.001129)(2.000000e-01,0.001236)(2.500000e-01,0.001312)(3.000000e-01,0.001323)(3.500000e-01,0.001231)(4.000000e-01,0.001005)(4.500000e-01,0.000630)(5.000000e-01,0.000089)
};
     \addplot plot coordinates {
(5.000000e-02,0.000594)(1.000000e-01,0.000744)(1.500000e-01,0.000867)(2.000000e-01,0.000965)(2.500000e-01,0.001029)(3.000000e-01,0.001036)(3.500000e-01,0.000959)(4.000000e-01,0.000777)(4.500000e-01,0.000478)(5.000000e-01,0.000056)
};
      \addplot  plot coordinates {
(5.000000e-02,0.000399)(1.000000e-01,0.000562)(1.500000e-01,0.000695)(2.000000e-01,0.000800)(2.500000e-01,0.000869)(3.000000e-01,0.000884)(3.500000e-01,0.000822)(4.000000e-01,0.000665)(4.500000e-01,0.000405)(5.000000e-01,0.000044)
};
\end{axis}
\end{tikzpicture}
\end{adjustbox}
\caption{$d=2$, $n_t=20$}
\label{fig:qdansBa}
\end{subfigure}
\begin{subfigure}[b]{0.3\textwidth}
\begin{adjustbox}{width=0.9\linewidth}
\begin{tikzpicture}
\begin{axis}[width=1\linewidth,font=\footnotesize,cycle list name=colors_qda,
     xlabel= Bayes error, ylabel= MSE,
  xmin=5.000000e-02,xmax=5.000000e-01,ymin=0,ymax=1.2e-03, scaled y ticks=base 10:4,
     ylabel near ticks,
    xlabel near ticks,
    legend style={nodes=right},legend style={nodes={scale=1, transform shape}},
      xmajorgrids,
    grid style={dotted},
    ymajorgrids,
   ]
  \addplot plot coordinates{
(5.000000e-02,0.000258)(1.000000e-01,0.000532)(1.500000e-01,0.000756)(2.000000e-01,0.000933)(2.500000e-01,0.001054)(3.000000e-01,0.001098)(3.500000e-01,0.001040)(4.000000e-01,0.000861)(4.500000e-01,0.000549)(5.000000e-01,0.000060)
};
 \addplot plot coordinates{
 (5.000000e-02,0.000265)(1.000000e-01,0.000532)(1.500000e-01,0.000748)(2.000000e-01,0.000916)(2.500000e-01,0.001029)(3.000000e-01,0.001066)(3.500000e-01,0.001004)(4.000000e-01,0.000826)(4.500000e-01,0.000522)(5.000000e-01,0.000054)
};
 \addplot plot coordinates{
(5.000000e-02,0.000241)(1.000000e-01,0.000513)(1.500000e-01,0.000730)(2.000000e-01,0.000896)(2.500000e-01,0.001003)(3.000000e-01,0.001033)(3.500000e-01,0.000967)(4.000000e-01,0.000789)(4.500000e-01,0.000490)(5.000000e-01,0.000044)
};
 \addplot plot coordinates{
(5.000000e-02,0.000227)(1.000000e-01,0.000484)(1.500000e-01,0.000686)(2.000000e-01,0.000835)(2.500000e-01,0.000925)(3.000000e-01,0.000943)(3.500000e-01,0.000874)(4.000000e-01,0.000705)(4.500000e-01,0.000430)(5.000000e-01,0.000030)
};
 \addplot plot coordinates{
(5.000000e-02,0.000180)(1.000000e-01,0.000404)(1.500000e-01,0.000580)(2.000000e-01,0.000708)(2.500000e-01,0.000785)(3.000000e-01,0.000800)(3.500000e-01,0.000740)(4.000000e-01,0.000595)(4.500000e-01,0.000359)(5.000000e-01,0.000024)
};
 \addplot plot coordinates{
(5.000000e-02,0.000146)(1.000000e-01,0.000354)(1.500000e-01,0.000517)(2.000000e-01,0.000636)(2.500000e-01,0.000709)(3.000000e-01,0.000723)(3.500000e-01,0.000669)(4.000000e-01,0.000536)(4.500000e-01,0.000319)(5.000000e-01,0.000021)
};
\end{axis}
\end{tikzpicture}
\end{adjustbox}
\caption{$d=2$, $n_t=50$}
\label{fig:qdansBd}
\end{subfigure}
\par\medskip 
\begin{subfigure}[b]{0.3\textwidth}
\begin{adjustbox}{width=0.86\linewidth}
\begin{tikzpicture}
    \begin{axis}[ width=1\linewidth,font=\footnotesize,cycle list name=colors_qda,
    xlabel= Bayes error,xtick={0.1,0.2,0.3,0.4}, ylabel= MSE,ylabel near ticks, scaled y ticks=base 10:4,
    xlabel near ticks,
 xmin=5.000000e-02,xmax=4.500000e-01,ymin=0,ymax=2.5e-03,
        xmajorgrids,
    grid style={dotted},
    ymajorgrids,
   ]

  \addplot plot coordinates {
(5.000000e-02,0.000832)(1.000000e-01,0.001481)(1.500000e-01,0.001945)(2.000000e-01,0.002226)(2.500000e-01,0.002321)(3.000000e-01,0.002226)(3.500000e-01,0.001937)(4.000000e-01,0.001452)(4.500000e-01,0.000472)
};
\addplot plot coordinates {
(5.000000e-02,0.000746)(1.000000e-01,0.001365)(1.500000e-01,0.001806)(2.000000e-01,0.002070)(2.500000e-01,0.002154)(3.000000e-01,0.002055)(3.500000e-01,0.001766)(4.000000e-01,0.001287)(4.500000e-01,0.000376)
};
     \addplot plot coordinates {
(5.000000e-02,0.000641)(1.000000e-01,0.001222)(1.500000e-01,0.001635)(2.000000e-01,0.001882)(2.500000e-01,0.001959)(3.000000e-01,0.001862)(3.500000e-01,0.001585)(4.000000e-01,0.001125)(4.500000e-01,0.000304)
};
    \addplot plot coordinates {
(5.000000e-02,0.000575)(1.000000e-01,0.001098)(1.500000e-01,0.001465)(2.000000e-01,0.001678)(2.500000e-01,0.001734)(3.000000e-01,0.001627)(3.500000e-01,0.001352)(4.000000e-01,0.000905)(4.500000e-01,0.000167)
};
     \addplot plot coordinates {
(5.000000e-02,0.000436)(1.000000e-01,0.000751)(1.500000e-01,0.000970)(2.000000e-01,0.001092)(2.500000e-01,0.001115)(3.000000e-01,0.001038)(3.500000e-01,0.000857)(4.000000e-01,0.000571)(4.500000e-01,0.000120)
};
      \addplot  plot coordinates {
(5.000000e-02,0.000311)(1.000000e-01,0.000554)(1.500000e-01,0.000723)(2.000000e-01,0.000818)(2.500000e-01,0.000838)(3.000000e-01,0.000783)(3.500000e-01,0.000652)(4.000000e-01,0.000443)(4.500000e-01,0.000158)
};
\end{axis}
\end{tikzpicture}
\end{adjustbox}
\caption{$d=3$, $n_t=20$}
\label{fig:qdansBb}
\end{subfigure}
\begin{subfigure}[b]{0.3\textwidth}
\begin{adjustbox}{width=0.86\linewidth}
\begin{tikzpicture}
\begin{axis}[width=1\linewidth,font=\footnotesize,cycle list name=colors_qda,
    xlabel= Bayes error, ylabel= MSE,
  xmin=5.000000e-02,xmax=4.500000e-01,ylabel near ticks,ymin=0,xtick={0.1,0.2,0.3,0.4}, scaled y ticks=base 10:4,
    xlabel near ticks,
legend style={nodes=right},legend style={nodes={scale=1, transform shape}},
      xmajorgrids,
    grid style={dotted},
    ymajorgrids,
   ]
  \addplot plot coordinates{
(5.000000e-02,0.000631)(1.000000e-01,0.000887)(1.500000e-01,0.001069)(2.000000e-01,0.001176)(2.500000e-01,0.001209)(3.000000e-01,0.001168)(3.500000e-01,0.001051)(4.000000e-01,0.000858)(4.500000e-01,0.000290)
};
 \addplot plot coordinates{
(5.000000e-02,0.000572)(1.000000e-01,0.000852)(1.500000e-01,0.001049)(2.000000e-01,0.001164)(2.500000e-01,0.001195)(3.000000e-01,0.001142)(3.500000e-01,0.001003)(4.000000e-01,0.000778)(4.500000e-01,0.000228)
};
 \addplot plot coordinates{
(5.000000e-02,0.000511)(1.000000e-01,0.000810)(1.500000e-01,0.001018)(2.000000e-01,0.001135)(2.500000e-01,0.001159)(3.000000e-01,0.001090)(3.500000e-01,0.000926)(4.000000e-01,0.000665)(4.500000e-01,0.000128)
};
 \addplot plot coordinates{
(5.000000e-02,0.000457)(1.000000e-01,0.000729)(1.500000e-01,0.000916)(2.000000e-01,0.001018)(2.500000e-01,0.001034)(3.000000e-01,0.000962)(3.500000e-01,0.000802)(4.000000e-01,0.000551)(4.500000e-01,0.000091)
};
 \addplot plot coordinates{
(5.000000e-02,0.000364)(1.000000e-01,0.000546)(1.500000e-01,0.000667)(2.000000e-01,0.000727)(2.500000e-01,0.000726)(3.000000e-01,0.000664)(3.500000e-01,0.000541)(4.000000e-01,0.000355)(4.500000e-01,0.000048)
};
 \addplot plot coordinates{
(5.000000e-02,0.000265)(1.000000e-01,0.000416)(1.500000e-01,0.000516)(2.000000e-01,0.000565)(2.500000e-01,0.000562)(3.000000e-01,0.000508)(3.500000e-01,0.000405)(4.000000e-01,0.000251)(4.500000e-01,0.000047)
};
\end{axis}
\end{tikzpicture}
\end{adjustbox}
\caption{$d=3$, $n_t=50$}
\label{fig:qdansBe}
\end{subfigure}
\par\medskip 
\begin{subfigure}[b]{0.3\textwidth}
\begin{adjustbox}{width=0.9\linewidth}
\begin{tikzpicture}
\begin{axis}[ width=1\linewidth,font=\footnotesize,cycle list name=colors_qda,
    xlabel= Bayes error, ylabel= MSE,
  xmin=5.000000e-02,xmax=4.000000e-01,ymin=0, scaled y ticks=base 10:3,
 ylabel near ticks,xtick={0.1,0.2,0.3,0.4},
    xlabel near ticks,
    legend style={nodes=right},legend style={nodes={scale=1, transform shape}},
      xmajorgrids,
    grid style={dotted},
    ymajorgrids,
   ]
  \addplot plot coordinates {
(5.000000e-02,0.005161)(1.000000e-01,0.012255)(1.500000e-01,0.017081)(2.000000e-01,0.019675)(2.500000e-01,0.019880)(3.000000e-01,0.017696)(3.500000e-01,0.013284)(4.000000e-01,0.006607)
};
  \addplot plot coordinates {
(5.000000e-02,0.004366)(1.000000e-01,0.011419)(1.500000e-01,0.016203)(2.000000e-01,0.018752)(2.500000e-01,0.018920)(3.000000e-01,0.016724)(3.500000e-01,0.012335)(4.000000e-01,0.005713)
};
  \addplot plot coordinates {
(5.000000e-02,0.003789)(1.000000e-01,0.009927)(1.500000e-01,0.014124)(2.000000e-01,0.016413)(2.500000e-01,0.016649)(3.000000e-01,0.014830)(3.500000e-01,0.011094)(4.000000e-01,0.005409)
};
  \addplot plot coordinates {
(5.000000e-02,0.003106)(1.000000e-01,0.007201)(1.500000e-01,0.010076)(2.000000e-01,0.011763)(2.500000e-01,0.012127)(3.000000e-01,0.011113)(3.500000e-01,0.008784)(4.000000e-01,0.005128)
};
  \addplot plot coordinates {
(5.000000e-02,0.002497)(1.000000e-01,0.003301)(1.500000e-01,0.003996)(2.000000e-01,0.004622)(2.500000e-01,0.005032)(3.000000e-01,0.005048)(3.500000e-01,0.004591)(4.000000e-01,0.003685)
};
  \addplot plot coordinates {
(5.000000e-02,0.001580)(1.000000e-01,0.001587)(1.500000e-01,0.001711)(2.000000e-01,0.001991)(2.500000e-01,0.002287)(3.000000e-01,0.002408)(3.500000e-01,0.002249)(4.000000e-01,0.001842)
};
\end{axis}
\end{tikzpicture}
\end{adjustbox}
\caption{$d=5$, $n_t=20$}
\label{fig:qdansBc}
\end{subfigure}
\begin{subfigure}[b]{0.3\textwidth}
\begin{adjustbox}{width=0.9\linewidth}
\begin{tikzpicture}
\begin{axis}[ width=1\linewidth,font=\footnotesize,cycle list name=colors_qda,
    xlabel= Bayes error, ylabel= MSE,
  xmin=5.000000e-02,xmax=4.000000e-01,ymin=0, xtick={0.1,0.2,0.3,0.4},scaled y ticks=base 10:4,
   ylabel near ticks,
    xlabel near ticks,
    legend style={nodes=right},legend style={nodes={scale=1, transform shape}},
      xmajorgrids,
    grid style={dotted},
    ymajorgrids,
   ]
 \addplot plot coordinates{
(5.000000e-02,0.002523)(1.000000e-01,0.004690)(1.500000e-01,0.006425)(2.000000e-01,0.007778)(2.500000e-01,0.008397)(3.000000e-01,0.007880)(3.500000e-01,0.006072)(4.000000e-01,0.003021)
};
 \addplot plot coordinates{
(5.000000e-02,0.002023)(1.000000e-01,0.004187)(1.500000e-01,0.005919)(2.000000e-01,0.007264)(2.500000e-01,0.007881)(3.000000e-01,0.007375)(3.500000e-01,0.005597)(4.000000e-01,0.002596)
};
 \addplot plot coordinates{
(5.000000e-02,0.001522)(1.000000e-01,0.003612)(1.500000e-01,0.005282)(2.000000e-01,0.006569)(2.500000e-01,0.007164)(3.000000e-01,0.006718)(3.500000e-01,0.005107)(4.000000e-01,0.002371)
};
 \addplot plot coordinates{
(5.000000e-02,0.001002)(1.000000e-01,0.002717)(1.500000e-01,0.004072)(2.000000e-01,0.005126)(2.500000e-01,0.005642)(3.000000e-01,0.005380)(3.500000e-01,0.004263)(4.000000e-01,0.002314)
};
 \addplot plot coordinates{
(5.000000e-02,0.000565)(1.000000e-01,0.001446)(1.500000e-01,0.002152)(2.000000e-01,0.002707)(2.500000e-01,0.003027)(3.000000e-01,0.003021)(3.500000e-01,0.002658)(4.000000e-01,0.001948)
};
 \addplot plot coordinates{
(5.000000e-02,0.000428)(1.000000e-01,0.000778)(1.500000e-01,0.001093)(2.000000e-01,0.001390)(2.500000e-01,0.001603)(3.000000e-01,0.001653)(3.500000e-01,0.001503)(4.000000e-01,0.001164)};
\end{axis}
\end{tikzpicture}
\end{adjustbox}
\caption{$d=5$, $n_t=50$}
\label{fig:qdansBf}
\end{subfigure}

\caption{MSE deviation from QDA true error with respect to Bayes error. Source sample size was set to $n_{s}=200$ in all figures.}
\label{fig:qdansBayes}
\end{center}
\end{figure}

%% file: figures/QDA/QDA_d_5_n_s_flipped.tex
\begin{figure}[h!]
\begin{center}
\ref{named4}
\par
\bigskip
\begin{subfigure}[b]{0.3\textwidth}
\begin{adjustbox}{width=0.9\linewidth}
\begin{tikzpicture}
    \begin{axis}[ width=1\linewidth,font=\footnotesize,cycle list name=colors_qda,
    xlabel= $n_s$, ylabel= MSE,ylabel near ticks,
    xlabel near ticks,
  xmin=10,xmax=500,xtick={10,100,200,300,400,500},scaled y ticks=base 10:3,
legend columns=3,
legend entries={$\left|\alpha\right|=0.1$,$\left|\alpha\right|=0.3$,$\left|\alpha\right|=0.5$,$\left|\alpha\right|=0.7$,$\left|\alpha\right|=0.9$,$\left|\alpha\right|=0.95$},
legend to name=named4,
    legend style={nodes=right},legend style={nodes={scale=0.8, transform shape}},
      xmajorgrids,
    grid style={dotted},
    ymajorgrids,
   ]
\addplot plot coordinates {
(10,0.040611)(50,0.040549)(100,0.040526)(150,0.040547)(200,0.040602)(250,0.040671)(300,0.040735)(350,0.040778)(400,0.040788)(450,0.040763)(500,0.040704)
};
\addplot plot coordinates {
(10,0.039966)(50,0.039911)(100,0.039878)(150,0.039866)(200,0.039878)(250,0.039921)(300,0.039993)(350,0.040075)(400,0.040141)(450,0.040175)(500,0.040182)
};
     \addplot plot coordinates {
(10,0.037926)(50,0.036628)(100,0.035770)(150,0.035386)(200,0.035356)(250,0.035487)(300,0.035634)(350,0.035743)(400,0.035816)(450,0.035864)(500,0.035887)
};
    \addplot plot coordinates {
(10,0.034546)(50,0.029618)(100,0.026221)(150,0.024472)(200,0.023957)(250,0.024029)(300,0.024214)(350,0.024345)(400,0.024428)(450,0.024493)(500,0.024536)
};
     \addplot plot coordinates {
(10,0.030067)(50,0.020152)(100,0.013192)(150,0.009410)(200,0.008027)(250,0.007811)(300,0.007857)(350,0.007837)(400,0.007767)(450,0.007710)(500,0.007655)
};
      \addplot  plot coordinates {
(10,0.029084)(50,0.017903)(100,0.010057)(150,0.005799)(200,0.004246)(250,0.004004)(300,0.004035)(350,0.003975)(400,0.003860)(450,0.003778)(500,0.003712)
};
\end{axis}
\end{tikzpicture}
\end{adjustbox}
\caption{$n_t=10$, $n_s n_s$}
\label{fig:qdansFa}
\end{subfigure}
\begin{subfigure}[b]{0.3\textwidth}
\begin{adjustbox}{width=0.9\linewidth}
\begin{tikzpicture}
\begin{axis}[width=1\linewidth,font=\footnotesize,cycle list name=colors_qda,
    xlabel= $n_s$, ylabel= MSE,
  xmin=10,xmax=500,xtick={10,100,200,300,400,500},ylabel near ticks,scaled y ticks=base 10:2,
    xlabel near ticks,
    legend style={nodes=right},legend style={nodes={scale=1, transform shape}},
      xmajorgrids,
    grid style={dotted},
    ymajorgrids,
   ]
  \addplot plot coordinates{
(10,0.048751)(50,0.069138)(100,0.084184)(150,0.093495)(200,0.098459)(250,0.101169)(300,0.103066)(350,0.104632)(400,0.105915)(450,0.106915)(500,0.107623)
};
 \addplot plot coordinates{
(10,0.042987)(50,0.056060)(100,0.065975)(150,0.072525)(200,0.076436)(250,0.078844)(300,0.080609)(350,0.082132)(400,0.083566)(450,0.084973)(500,0.086330)
};
 \addplot plot coordinates{
(10,0.038354)(50,0.044463)(100,0.049364)(150,0.053001)(200,0.055562)(250,0.057385)(300,0.058812)(350,0.060103)(400,0.061430)(450,0.062866)(500,0.064387)
};
 \addplot plot coordinates{
(10,0.035279)(50,0.035631)(100,0.036224)(150,0.037097)(200,0.038100)(250,0.039036)(300,0.039846)(350,0.040643)(400,0.041601)(450,0.042824)(500,0.044281)
};
 \addplot plot coordinates{
(10,0.033788)(50,0.029858)(100,0.027096)(150,0.025571)(200,0.025038)(250,0.025086)(300,0.025402)(350,0.025889)(400,0.026613)(450,0.027646)(500,0.028973)
};
 \addplot plot coordinates{
(10,0.033602)(50,0.029107)(100,0.025881)(150,0.024000)(200,0.023204)(250,0.023058)(300,0.023217)(350,0.023573)(400,0.024191)(450,0.025152)(500,0.026438)
};
\end{axis}
\end{tikzpicture}
\end{adjustbox}
\caption{$n_t=10$, $n_s n_t$}
\label{fig:qdansFd}
\end{subfigure}
\par\medskip 
\begin{subfigure}[b]{0.3\textwidth}
\begin{adjustbox}{width=0.9\linewidth}
\begin{tikzpicture}
    \begin{axis}[ width=1\linewidth,font=\footnotesize,cycle list name=colors_qda,
    xlabel= $n_s$, ylabel= MSE,ylabel near ticks,
    xlabel near ticks,
  xmin=10,xmax=500,xtick={10,100,200,300,400,500},scaled y ticks=base 10:3,
      xmajorgrids,
    grid style={dotted},
    ymajorgrids,
   ]

  \addplot plot coordinates {
(10,0.020270)(50,0.020392)(100,0.020463)(150,0.020478)(200,0.020460)(250,0.020442)(300,0.020439)(350,0.020444)(400,0.020443)(450,0.020429)(500,0.020404)
};
\addplot plot coordinates {
(10,0.019982)(50,0.020104)(100,0.020190)(150,0.020235)(200,0.020252)(250,0.020267)(300,0.020301)(350,0.020358)(400,0.020418)(450,0.020462)(500,0.020495)
};
     \addplot plot coordinates {
(10,0.019124)(50,0.018887)(100,0.018731)(150,0.018660)(200,0.018661)(250,0.018711)(300,0.018792)(350,0.018875)(400,0.018928)(450,0.018930)(500,0.018886)
};
    \addplot plot coordinates {
(10,0.017943)(50,0.016224)(100,0.015033)(150,0.014410)(200,0.014222)(250,0.014260)(300,0.014363)(350,0.014469)(400,0.014563)(450,0.014650)(500,0.014728)
};
     \addplot plot coordinates {
(10,0.016812)(50,0.012286)(100,0.009057)(150,0.007213)(200,0.006448)(250,0.006265)(300,0.006274)(350,0.006302)(400,0.006315)(450,0.006320)(500,0.006320)
};
      \addplot  plot coordinates {
(10,0.016552)(50,0.011146)(100,0.007262)(150,0.005002)(200,0.004015)(250,0.003723)(300,0.003663)(350,0.003627)(400,0.003582)(450,0.003544)(500,0.003514)
};
\end{axis}
\end{tikzpicture}
\end{adjustbox}
\caption{$n_t=30$, $n_s n_s$}
\label{fig:qdansFb}
\end{subfigure}
\begin{subfigure}[b]{0.3\textwidth}
\begin{adjustbox}{width=0.9\linewidth}
\begin{tikzpicture}
\begin{axis}[width=1\linewidth,font=\footnotesize,cycle list name=colors_qda,
    xlabel= $n_s$, ylabel= MSE,
  xmin=10,xmax=500,xtick={10,100,200,300,400,500},ylabel near ticks,scaled y ticks=base 10:2,
    xlabel near ticks,
legend style={nodes=right},legend style={nodes={scale=1, transform shape}},
      xmajorgrids,
    grid style={dotted},
    ymajorgrids,
   ]
  \addplot plot coordinates{
(10,0.030660)(50,0.055798)(100,0.074537)(150,0.086438)(200,0.093044)(250,0.096729)(300,0.099207)(350,0.101154)(400,0.102727)(450,0.103971)(500,0.104863)
};
 \addplot plot coordinates{
(10,0.027043)(50,0.044809)(100,0.058144)(150,0.066741)(200,0.071671)(250,0.074603)(300,0.076767)(350,0.078670)(400,0.080432)(450,0.082076)(500,0.083585)
};
 \addplot plot coordinates{
(10,0.023946)(50,0.034696)(100,0.042891)(150,0.048368)(200,0.051693)(250,0.053792)(300,0.055415)(350,0.056946)(400,0.058522)(450,0.060180)(500,0.061902)
};
 \addplot plot coordinates{
(10,0.022099)(50,0.027090)(100,0.030949)(150,0.033601)(200,0.035303)(250,0.036469)(300,0.037438)(350,0.038402)(400,0.039465)(450,0.040673)(500,0.042010)
};
 \addplot plot coordinates{
(10,0.021270)(50,0.022294)(100,0.022976)(150,0.023268)(200,0.023337)(250,0.023421)(300,0.023676)(350,0.024171)(400,0.024957)(450,0.026074)(500,0.027509)
};
 \addplot plot coordinates{
(10,0.021236)(50,0.021832)(100,0.022098)(150,0.021975)(200,0.021683)(250,0.021520)(300,0.021643)(350,0.022067)(400,0.022787)(450,0.023819)(500,0.025157)
};
\end{axis}
\end{tikzpicture}
\end{adjustbox}
\caption{$n_t=30$, $n_s n_t$}
\label{fig:qdansFe}
\end{subfigure}
\par\medskip 
\begin{subfigure}[b]{0.3\textwidth}
\begin{adjustbox}{width=0.9\linewidth}
\begin{tikzpicture}
\begin{axis}[width=1\linewidth,font=\footnotesize,cycle list name=colors_qda,
    xlabel= $n_s$, ylabel= MSE,
  xmin=10,xmax=500,xtick={10,100,200,300,400,500},ylabel near ticks,scaled y ticks=base 10:3,
    xlabel near ticks,
    legend style={nodes=right},legend style={nodes={scale=1, transform shape}},
      xmajorgrids,
    grid style={dotted},
    ymajorgrids,
   ]
  \addplot plot coordinates {
(10,0.011367)(50,0.011433)(100,0.011480)(150,0.011508)(200,0.011518)(250,0.011518)(300,0.011516)(350,0.011517)(400,0.011520)(450,0.011524)(500,0.011528)
};
  \addplot plot coordinates {
(10,0.011247)(50,0.011391)(100,0.011488)(150,0.011535)(200,0.011548)(250,0.011550)(300,0.011560)(350,0.011578)(400,0.011591)(450,0.011590)(500,0.011578)
};
  \addplot plot coordinates {
(10,0.010836)(50,0.010799)(100,0.010792)(150,0.010817)(200,0.010864)(250,0.010917)(300,0.010973)(350,0.011029)(400,0.011073)(450,0.011092)(500,0.011091)
};
  \addplot plot coordinates {
(10,0.010281)(50,0.009630)(100,0.009189)(150,0.008973)(200,0.008931)(250,0.008981)(300,0.009061)(350,0.009140)(400,0.009206)(450,0.009254)(500,0.009285)
};
  \addplot plot coordinates {
(10,0.009994)(50,0.007782)(100,0.006255)(150,0.005371)(200,0.004997)(250,0.004911)(300,0.004936)(350,0.004980)(400,0.005017)(450,0.005042)(500,0.005058)
};
  \addplot plot coordinates {
(10,0.009910)(50,0.007254)(100,0.005254)(150,0.004037)(200,0.003447)(250,0.003227)(300,0.003164)(350,0.003149)(400,0.003146)(450,0.003152)(500,0.003169)
};
\end{axis}
\end{tikzpicture}
\end{adjustbox}
\caption{$n_t=50$, $n_s n_s$}
\label{fig:qdansFc}
\end{subfigure}
\begin{subfigure}[b]{0.3\textwidth}
\begin{adjustbox}{width=0.9\linewidth}
\begin{tikzpicture}
\begin{axis}[ width=1\linewidth,font=\footnotesize,cycle list name=colors_qda,
    xlabel= $n_s$, ylabel= MSE,
  xmin=10,xmax=500,xtick={10,100,200,300,400,500},ylabel near ticks,scaled y ticks=base 10:2,
    xlabel near ticks,
    legend style={nodes=right},legend style={nodes={scale=1, transform shape}},
      xmajorgrids,
    grid style={dotted},
    ymajorgrids,
   ]
 \addplot plot coordinates{
(10,0.019864)(50,0.045338)(100,0.064702)(150,0.077557)(200,0.085296)(250,0.090074)(300,0.093468)(350,0.096133)(400,0.098242)(450,0.099850)(500,0.100929)
};
 \addplot plot coordinates{
(10,0.017364)(50,0.035982)(100,0.050194)(150,0.059720)(200,0.065532)(250,0.069170)(300,0.071826)(350,0.074072)(400,0.076123)(450,0.078053)(500,0.079832)
};
 \addplot plot coordinates{
(10,0.015348)(50,0.027331)(100,0.036577)(150,0.042916)(200,0.046934)(250,0.049581)(300,0.051631)(350,0.053482)(400,0.055289)(450,0.057093)(500,0.058877)
};
 \addplot plot coordinates{
(10,0.014319)(50,0.021020)(100,0.026150)(150,0.029596)(200,0.031755)(250,0.033247)(300,0.034528)(350,0.035784)(400,0.037050)(450,0.038329)(500,0.039617)
};
 \addplot plot coordinates{
(10,0.013797)(50,0.017081)(100,0.019378)(150,0.020589)(200,0.021057)(250,0.021292)(300,0.021644)(350,0.022242)(400,0.023118)(450,0.024291)(500,0.025752)
};
 \addplot plot coordinates{
(10,0.013735)(50,0.016641)(100,0.018579)(150,0.019440)(200,0.019610)(250,0.019650)(300,0.019905)(350,0.020443)(400,0.021235)(450,0.022268)(500,0.023543)};
\end{axis}
\end{tikzpicture}
\end{adjustbox}
\caption{$n_t=50$, $n_s n_t$}
\label{fig:qdansFf}
\end{subfigure}

\caption{MSE deviation from true error
with respect to source sample size. The source class means are flipped with respect to target classes ($\mathbf{m}_{s}^{y} = \mathbf{m}_{t}^{1-y}$, for $y\in \left\{0,1\right\}$). In the first row, the source datasets are correctly considered as source samples. In the second row, the source datasets are intentionally considered as target samples. The Bayes error is fixed at 0.2 and $d=5$.}
\label{fig:qdansFlipped}
\end{center}
\end{figure}
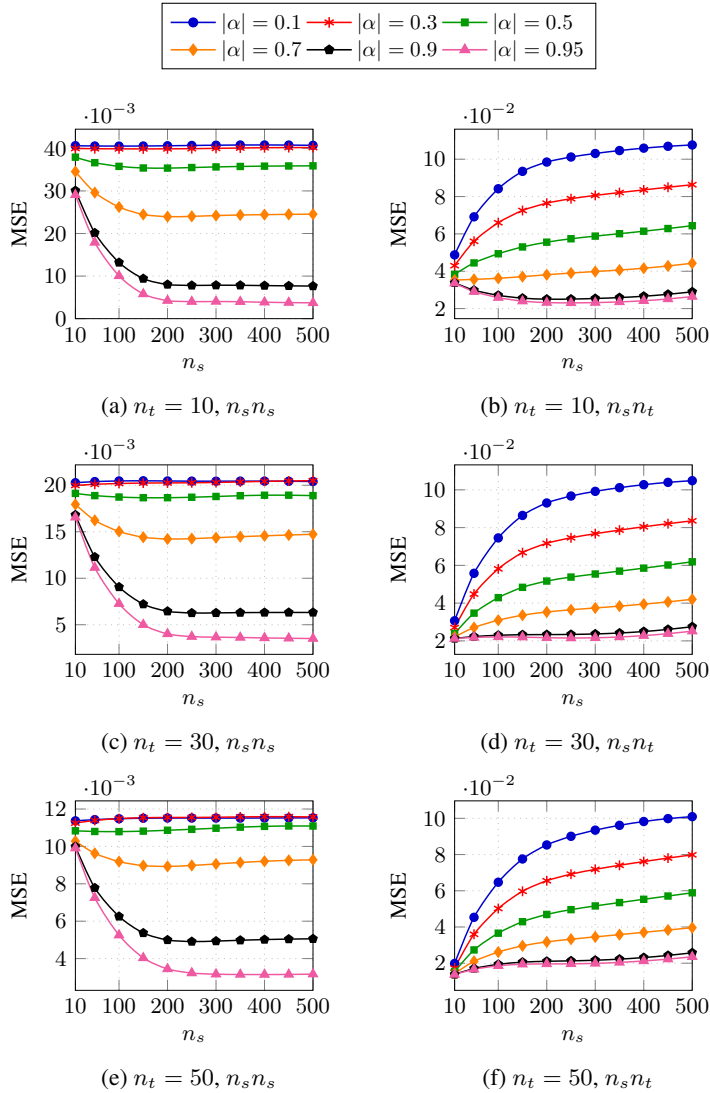

%% file: figures/LDA/LDA_n_s_Bayes.tex
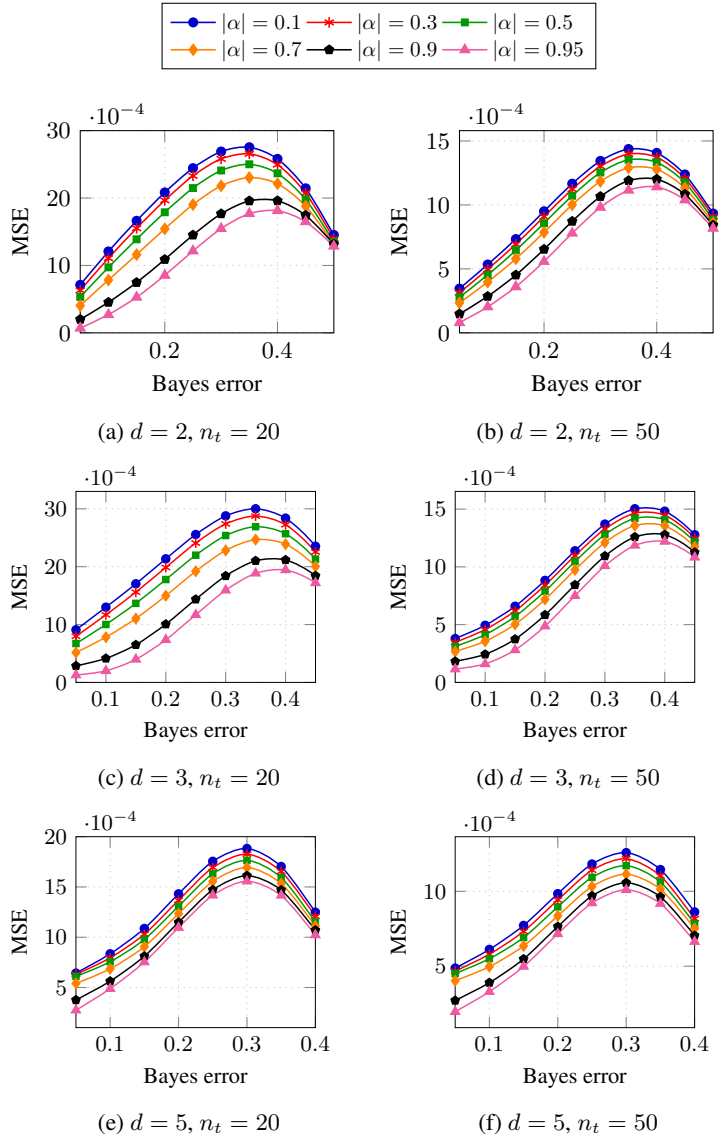
\begin{figure}[h!]
\begin{center}
\ref{named6}
\par
\bigskip
\begin{subfigure}[b]{0.3\textwidth}
\begin{adjustbox}{width=0.9\linewidth}
\begin{tikzpicture}
    \begin{axis}[ width=1\linewidth,font=\footnotesize,cycle list name=colors_qda,
    xlabel= Bayes error, ylabel= MSE,ylabel near ticks,
    xlabel near ticks,
  xmin=5.000000e-02,xmax=5.000000e-01,ymin=0,ymax=3e-03,scaled y ticks=base 10:4,
legend columns=3,
legend entries={$\left|\alpha\right|=0.1$,$\left|\alpha\right|=0.3$,$\left|\alpha\right|=0.5$,$\left|\alpha\right|=0.7$,$\left|\alpha\right|=0.9$,$\left|\alpha\right|=0.95$},
legend to name=named6,
    legend style={nodes=right},legend style={nodes={scale=0.8, transform shape}},
      xmajorgrids,
    grid style={dotted},
    ymajorgrids,
   ]

  \addplot plot coordinates {
(5.00e-02,0.000712)(1.00e-01,0.001208)(1.50e-01,0.001662)(2.00e-01,0.002083)(2.50e-01,0.002444)(3.00e-01,0.002690)(3.50e-01,0.002754)(4.00e-01,0.002581)(4.50e-01,0.002144)(5.00e-01,0.001452)
};
\addplot plot coordinates {
(5.00e-02,0.000630)(1.00e-01,0.001108)(1.50e-01,0.001551)(2.00e-01,0.001966)(2.50e-01,0.002329)(3.00e-01,0.002581)(3.50e-01,0.002656)(4.00e-01,0.002495)(4.50e-01,0.002074)(5.00e-01,0.001399)
};
     \addplot plot coordinates {
(5.00e-02,0.000533)(1.00e-01,0.000972)(1.50e-01,0.001388)(2.00e-01,0.001789)(2.50e-01,0.002148)(3.00e-01,0.002409)(3.50e-01,0.002501)(4.00e-01,0.002367)(4.50e-01,0.001980)(5.00e-01,0.001349)
};
    \addplot plot coordinates {
(5.00e-02,0.000405)(1.00e-01,0.000784)(1.50e-01,0.001161)(2.00e-01,0.001543)(2.50e-01,0.001903)(3.00e-01,0.002181)(3.50e-01,0.002304)(4.00e-01,0.002213)(4.50e-01,0.001880)(5.00e-01,0.001314)
};
     \addplot plot coordinates {
(5.00e-02,0.000200)(1.00e-01,0.000453)(1.50e-01,0.000747)(2.00e-01,0.001089)(2.50e-01,0.001451)(3.00e-01,0.001767)(3.50e-01,0.001958)(4.00e-01,0.001960)(4.50e-01,0.001744)(5.00e-01,0.001317)
};
      \addplot  plot coordinates {
(5.00e-02,0.000070)(1.00e-01,0.000270)(1.50e-01,0.000528)(2.00e-01,0.000853)(2.50e-01,0.001215)(3.00e-01,0.001547)(3.50e-01,0.001769)(4.00e-01,0.001813)(4.50e-01,0.001649)(5.00e-01,0.001286)
};
\end{axis}
\end{tikzpicture}
\end{adjustbox}
\caption{$d=2$, $n_t=20$}
\label{fig:ldansBa}
\end{subfigure}
\begin{subfigure}[b]{0.3\textwidth}
\begin{adjustbox}{width=0.9\linewidth}
\begin{tikzpicture}
\begin{axis}[width=1\linewidth,font=\footnotesize,cycle list name=colors_qda,
     xlabel= Bayes error, ylabel= MSE,
  xmin=5.000000e-02,xmax=5.000000e-01,ymin=0,scaled y ticks=base 10:4,
     ylabel near ticks,
    xlabel near ticks,
    legend style={nodes=right},legend style={nodes={scale=1, transform shape}},
      xmajorgrids,
    grid style={dotted},
    ymajorgrids,
   ]
  \addplot plot coordinates{
(5.00e-02,0.000346)(1.00e-01,0.000534)(1.50e-01,0.000733)(2.00e-01,0.000950)(2.50e-01,0.001166)(3.00e-01,0.001344)(3.50e-01,0.001437)(4.00e-01,0.001407)(4.50e-01,0.001238)(5.00e-01,0.000933)
};
 \addplot plot coordinates{
(5.00e-02,0.000314)(1.00e-01,0.000497)(1.50e-01,0.000693)(2.00e-01,0.000908)(2.50e-01,0.001123)(3.00e-01,0.001302)(3.50e-01,0.001398)(4.00e-01,0.001373)(4.50e-01,0.001208)(5.00e-01,0.000911)
};
 \addplot plot coordinates{
(5.00e-02,0.000278)(1.00e-01,0.000453)(1.50e-01,0.000645)(2.00e-01,0.000857)(2.50e-01,0.001072)(3.00e-01,0.001253)(3.50e-01,0.001353)(4.00e-01,0.001333)(4.50e-01,0.001175)(5.00e-01,0.000885)
};
 \addplot plot coordinates{
(5.00e-02,0.000235)(1.00e-01,0.000397)(1.50e-01,0.000579)(2.00e-01,0.000786)(2.50e-01,0.001001)(3.00e-01,0.001184)(3.50e-01,0.001290)(4.00e-01,0.001279)(4.50e-01,0.001133)(5.00e-01,0.000859)
};
 \addplot plot coordinates{
(5.00e-02,0.000148)(1.00e-01,0.000285)(1.50e-01,0.000452)(2.00e-01,0.000653)(2.50e-01,0.000872)(3.00e-01,0.001066)(3.50e-01,0.001190)(4.00e-01,0.001201)(4.50e-01,0.001083)(5.00e-01,0.000841)
};
 \addplot plot coordinates{
(5.00e-02,0.000080)(1.00e-01,0.000203)(1.50e-01,0.000360)(2.00e-01,0.000558)(2.50e-01,0.000779)(3.00e-01,0.000981)(3.50e-01,0.001115)(4.00e-01,0.001141)(4.50e-01,0.001039)(5.00e-01,0.000817)
};
\end{axis}
\end{tikzpicture}
\end{adjustbox}
\caption{$d=2$, $n_t=50$}
\label{fig:ldansBd}
\end{subfigure}
\par\medskip 
\begin{subfigure}[b]{0.3\textwidth}
\begin{adjustbox}{width=.85\linewidth}
\begin{tikzpicture}
    \begin{axis}[width=1\linewidth,font=\footnotesize,cycle list name=colors_qda,
    xlabel= Bayes error,xtick={0.1,0.2,0.3,0.4}, ylabel= MSE,ylabel near ticks, scaled y ticks=base 10:4,
    xlabel near ticks,
 xmin=5.000000e-02,xmax=4.500000e-01,ymin=0,
        xmajorgrids,
    grid style={dotted},
    ymajorgrids,
   ]

  \addplot plot coordinates {
(5.00e-02,0.000909)(1.00e-01,0.001300)(1.50e-01,0.001704)(2.00e-01,0.002136)(2.50e-01,0.002556)(3.00e-01,0.002878)(3.50e-01,0.003001)(4.00e-01,0.002839)(4.50e-01,0.002353)(5.00e-01,0.001555)
};
\addplot plot coordinates {
(5.00e-02,0.000798)(1.00e-01,0.001168)(1.50e-01,0.001560)(2.00e-01,0.001987)(2.50e-01,0.002409)(3.00e-01,0.002740)(3.50e-01,0.002875)(4.00e-01,0.002726)(4.50e-01,0.002254)(5.00e-01,0.001471)
};
     \addplot plot coordinates {
(5.00e-02,0.000672)(1.00e-01,0.001001)(1.50e-01,0.001366)(2.00e-01,0.001778)(2.50e-01,0.002199)(3.00e-01,0.002539)(3.50e-01,0.002692)(4.00e-01,0.002570)(4.50e-01,0.002131)(5.00e-01,0.001390)
};
    \addplot plot coordinates {
(5.00e-02,0.000516)(1.00e-01,0.000781)(1.50e-01,0.001104)(2.00e-01,0.001498)(2.50e-01,0.001921)(3.00e-01,0.002282)(3.50e-01,0.002469)(4.00e-01,0.002390)(4.50e-01,0.002004)(5.00e-01,0.001322)
};
     \addplot plot coordinates {
(5.00e-02,0.000285)(1.00e-01,0.000415)(1.50e-01,0.000650)(2.00e-01,0.001006)(2.50e-01,0.001436)(3.00e-01,0.001841)(3.50e-01,0.002102)(4.00e-01,0.002117)(4.50e-01,0.001843)(5.00e-01,0.001292)
};
      \addplot  plot coordinates {
(5.00e-02,0.000127)(1.00e-01,0.000200)(1.50e-01,0.000398)(2.00e-01,0.000736)(2.50e-01,0.001169)(3.00e-01,0.001593)(3.50e-01,0.001886)(4.00e-01,0.001944)(4.50e-01,0.001720)(5.00e-01,0.001229)
};
\end{axis}
\end{tikzpicture}
\end{adjustbox}
\caption{$d=3$, $n_t=20$}
\label{fig:ldansBb}
\end{subfigure}
\begin{subfigure}[b]{0.3\textwidth}
\begin{adjustbox}{width=.85\linewidth}
\begin{tikzpicture}
\begin{axis}[width=1\linewidth,font=\footnotesize,cycle list name=colors_qda,
    xlabel= Bayes error, ylabel= MSE,
  xmin=5.000000e-02,xmax=4.500000e-01,ylabel near ticks,ymin=0,xtick={0.1,0.2,0.3,0.4}, scaled y ticks=base 10:4,
    xlabel near ticks,
legend style={nodes=right},legend style={nodes={scale=1, transform shape}},
      xmajorgrids,
    grid style={dotted},
    ymajorgrids,
   ]
  \addplot plot coordinates{
(5.00e-02,0.000379)(1.00e-01,0.000494)(1.50e-01,0.000659)(2.00e-01,0.000883)(2.50e-01,0.001139)(3.00e-01,0.001370)(3.50e-01,0.001503)(4.00e-01,0.001482)(4.50e-01,0.001278)(5.00e-01,0.000901)
};
 \addplot plot coordinates{
(5.00e-02,0.000347)(1.00e-01,0.000457)(1.50e-01,0.000619)(2.00e-01,0.000841)(2.50e-01,0.001097)(3.00e-01,0.001328)(3.50e-01,0.001465)(4.00e-01,0.001447)(4.50e-01,0.001249)(5.00e-01,0.000878)
};
 \addplot plot coordinates{
(5.00e-02,0.000311)(1.00e-01,0.000413)(1.50e-01,0.000570)(2.00e-01,0.000790)(2.50e-01,0.001046)(3.00e-01,0.001280)(3.50e-01,0.001420)(4.00e-01,0.001408)(4.50e-01,0.001216)(5.00e-01,0.000853)
};
 \addplot plot coordinates{
(5.00e-02,0.000268)(1.00e-01,0.000356)(1.50e-01,0.000503)(2.00e-01,0.000718)(2.50e-01,0.000974)(3.00e-01,0.001211)(3.50e-01,0.001358)(4.00e-01,0.001355)(4.50e-01,0.001174)(5.00e-01,0.000826)
};
 \addplot plot coordinates{
(5.00e-02,0.000182)(1.00e-01,0.000243)(1.50e-01,0.000374)(2.00e-01,0.000584)(2.50e-01,0.000844)(3.00e-01,0.001094)(3.50e-01,0.001259)(4.00e-01,0.001279)(4.50e-01,0.001126)(5.00e-01,0.000807)
};
 \addplot plot coordinates{
(5.00e-02,0.000115)(1.00e-01,0.000160)(1.50e-01,0.000281)(2.00e-01,0.000487)(2.50e-01,0.000750)(3.00e-01,0.001009)(3.50e-01,0.001186)(4.00e-01,0.001220)(4.50e-01,0.001083)(5.00e-01,0.000782)
};
\end{axis}
\end{tikzpicture}
\end{adjustbox}
\caption{$d=3$, $n_t=50$}
\label{fig:ldansBe}
\end{subfigure}
\par\medskip 
\begin{subfigure}[b]{0.3\textwidth}
\begin{adjustbox}{width=0.9\linewidth}
\begin{tikzpicture}
\begin{axis}[width=1\linewidth,font=\footnotesize,cycle list name=colors_qda,
    xlabel= Bayes error, ylabel= MSE,
  xmin=5.000000e-02,xmax=4.000000e-01,scaled y ticks=base 10:4,ymax=2e-03,
 ylabel near ticks,xtick={0.1,0.2,0.3,0.4},
    xlabel near ticks,
    legend style={nodes=right},legend style={nodes={scale=1, transform shape}},
      xmajorgrids,
    grid style={dotted},
    ymajorgrids,
   ]
  \addplot plot coordinates {
(5.00e-02,0.000643)(1.00e-01,0.000835)(1.50e-01,0.001087)(2.00e-01,0.001431)(2.50e-01,0.001755)(3.00e-01,0.001882)(3.50e-01,0.001703)(4.00e-01,0.001249)
};
  \addplot plot coordinates {
(5.00e-02,0.000630)(1.00e-01,0.000795)(1.50e-01,0.001030)(2.00e-01,0.001367)(2.50e-01,0.001693)(3.00e-01,0.001826)(3.50e-01,0.001654)(4.00e-01,0.001209)
};
  \addplot plot coordinates {
(5.00e-02,0.000610)(1.00e-01,0.000756)(1.50e-01,0.000978)(2.00e-01,0.001308)(2.50e-01,0.001631)(3.00e-01,0.001764)(3.50e-01,0.001594)(4.00e-01,0.001152)
};
  \addplot plot coordinates {
(5.00e-02,0.000540)(1.00e-01,0.000686)(1.50e-01,0.000906)(2.00e-01,0.001234)(2.50e-01,0.001555)(3.00e-01,0.001693)(3.50e-01,0.001534)(4.00e-01,0.001112)
};
  \addplot plot coordinates {
(5.00e-02,0.000376)(1.00e-01,0.000563)(1.50e-01,0.000811)(2.00e-01,0.001148)(2.50e-01,0.001471)(3.00e-01,0.001612)(3.50e-01,0.001469)(4.00e-01,0.001070)
};
  \addplot plot coordinates {
(5.00e-02,0.000274)(1.00e-01,0.000488)(1.50e-01,0.000753)(2.00e-01,0.001096)(2.50e-01,0.001417)(3.00e-01,0.001557)(3.50e-01,0.001416)(4.00e-01,0.001022)
};
\end{axis}
\end{tikzpicture}
\end{adjustbox}
\caption{$d=5$, $n_t=20$}
\label{fig:ldansBc}
\end{subfigure}
\begin{subfigure}[b]{0.3\textwidth}
\begin{adjustbox}{width=0.9\linewidth}
\begin{tikzpicture}
\begin{axis}[ width=1\linewidth,font=\footnotesize,cycle list name=colors_qda,
    xlabel= Bayes error, ylabel= MSE,
  xmin=5.000000e-02,xmax=4.000000e-01, xtick={0.1,0.2,0.3,0.4},scaled y ticks=base 10:4,
   ylabel near ticks,
    xlabel near ticks,
    legend style={nodes=right},legend style={nodes={scale=1, transform shape}},
      xmajorgrids,
    grid style={dotted},
    ymajorgrids,
   ]
 \addplot plot coordinates{
(5.00e-02,0.000487)(1.00e-01,0.000612)(1.50e-01,0.000772)(2.00e-01,0.000984)(2.50e-01,0.001182)(3.00e-01,0.001258)(3.50e-01,0.001145)(4.00e-01,0.000862)
};
 \addplot plot coordinates{
(5.00e-02,0.000467)(1.00e-01,0.000580)(1.50e-01,0.000732)(2.00e-01,0.000942)(2.50e-01,0.001141)(3.00e-01,0.001217)(3.50e-01,0.001103)(4.00e-01,0.000817)
};
 \addplot plot coordinates{
(5.00e-02,0.000451)(1.00e-01,0.000550)(1.50e-01,0.000691)(2.00e-01,0.000896)(2.50e-01,0.001093)(3.00e-01,0.001171)(3.50e-01,0.001062)(4.00e-01,0.000784)
};
 \addplot plot coordinates{
(5.00e-02,0.000403)(1.00e-01,0.000497)(1.50e-01,0.000635)(2.00e-01,0.000836)(2.50e-01,0.001032)(3.00e-01,0.001114)(3.50e-01,0.001014)(4.00e-01,0.000752)
};
 \addplot plot coordinates{
(5.00e-02,0.000271)(1.00e-01,0.000390)(1.50e-01,0.000548)(2.00e-01,0.000764)(2.50e-01,0.000969)(3.00e-01,0.001057)(3.50e-01,0.000963)(4.00e-01,0.000705)
};
 \addplot plot coordinates{
(5.00e-02,0.000196)(1.00e-01,0.000330)(1.50e-01,0.000498)(2.00e-01,0.000716)(2.50e-01,0.000922)(3.00e-01,0.001010)(3.50e-01,0.000918)(4.00e-01,0.000664)
};
\end{axis}
\end{tikzpicture}
\end{adjustbox}
\caption{$d=5$, $n_t=50$}
\label{fig:ldansBf}
\end{subfigure}

\caption{MSE deviation from LDA true error for Gaussian distributions with respect to Bayes error. Source sample size is set to $n_{s}=200$ in all figures.}
\label{fig:ldansBayes}
\end{center}
\end{figure}

%% file: figures/OBTL/standard_compare.tex
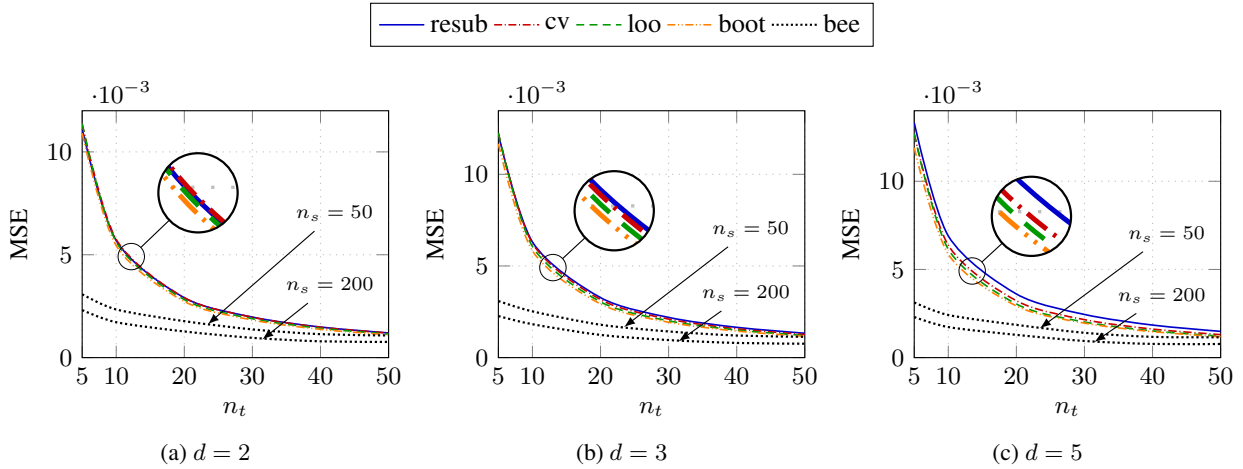
\begin{figure}[h!]
\begin{center}
\ref{named8}
\par
\bigskip
\begin{subfigure}[b]{0.33\textwidth}
\begin{adjustbox}{width=\linewidth}
\begin{tikzpicture}[
        spy using outlines={
            circle,
            magnification=3,
            connect spies,
            size=1cm,
            black,
        },
    ]
    \begin{axis}[width=1\linewidth,font=\footnotesize,cycle list name=colors_obtl,
    xlabel= $n_t$, ylabel= MSE,ylabel near ticks,scaled y ticks=base 10:3,ymin=0,
    xlabel near ticks,
  xmin=5,xmax=50,xtick={5,10,20,30,40,50},ymax=12e-03,
legend columns=-1,
legend entries={resub, cv, loo, boot, bee},
legend to name=named8,
    legend style={nodes=right},legend style={nodes={scale=1, transform shape}},
      xmajorgrids,
    grid style={dotted},
    ymajorgrids,
   ]

  \addplot plot coordinates {
(5,0.011126)(10,0.005686)(20,0.002909)(30,0.001970)(40,0.001485)(50,0.001199)
};
\addplot plot coordinates {
(5,0.011359)(10,0.005747)(20,0.002914)(30,0.001963)(40,0.001477)(50,0.001192)
};
     \addplot plot coordinates {
(5,0.011359)(10,0.005626)(20,0.002817)(30,0.001890)(40,0.001423)(50,0.001145)
};
    \addplot plot coordinates {
(5,0.010911)(10,0.005447)(20,0.002730)(30,0.001828)(40,0.001373)(50,0.001110)
};
     \addplot plot coordinates {
(5,0.003080)(10,0.002341)(20,0.001772)(30,0.001371)(40,0.001139)(50,0.001078)
};
     \addplot  plot coordinates {
(5,0.002311)(10,0.001721)(20,0.001271)(30,0.000961)(40,0.000792)(50,0.000763)
};

\node[anchor=west] (source) at (axis cs:35,7e-03){\scriptsize{$n_s=50$}};
       \node (destination) at (axis cs:22,0.0012){};
       \draw[->](source)--(destination);
%
%
\node[anchor=west] (source) at (axis cs:33.5,3.5e-03){\scriptsize{$n_s=200$}};
\node (destination) at (axis cs:30,0.0005){};
\draw[->](source)--(destination);

\coordinate (point) at (axis cs:12.2,4.9e-03);
\coordinate (spy point) at (axis cs:22,8e-03);
\spy on (point) in node [fill=white]  at (spy point);
\end{axis}
\end{tikzpicture}
\end{adjustbox}
\caption{$d=2$}
\label{fig:obtlnta}
\end{subfigure}
\begin{subfigure}[b]{0.33\textwidth}
\begin{adjustbox}{width=\linewidth}
\begin{tikzpicture}[
        spy using outlines={
            circle,
            magnification=3,
            connect spies,
            size=1cm,
            black,
        },
    ]
\begin{axis}[width=1\linewidth,font=\footnotesize,cycle list name=colors_obtl,
    xlabel= $n_t$, ylabel= MSE,
  xmin=5,xmax=50,xtick={5,10,20,30,40,50},ylabel near ticks,scaled y ticks=base 10:3,ymin=0,
    xlabel near ticks,
    legend style={nodes=right},legend style={nodes={scale=1, transform shape}},
      xmajorgrids,
    grid style={dotted},
    ymajorgrids,
  ]
  \addplot plot coordinates{
(5,0.012170)(10,0.006313)(20,0.003274)(30,0.002194)(40,0.001652)(50,0.001330)
};
 \addplot plot coordinates{
(5,0.012244)(10,0.006238)(20,0.003167)(30,0.002103)(40,0.001573)(50,0.001261)
};
 \addplot plot coordinates{
(5,0.012244)(10,0.006067)(20,0.003043)(30,0.002009)(40,0.001504)(50,0.001206)
};
 \addplot plot coordinates{
(5,0.011662)(10,0.005829)(20,0.002929)(30,0.001931)(40,0.001448)(50,0.001162)
};
 \addplot plot coordinates{
(5,0.003086)(10,0.002523)(20,0.001793)(30,0.001427)(40,0.001205)(50,0.001136)
};
 \addplot plot coordinates{
(5,0.002267)(10,0.001821)(20,0.001238)(30,0.000957)(40,0.000800)(50,0.000759)
};

\node[anchor=west] (source) at (axis cs:35,7e-03){\scriptsize{$n_s=50$}};
      \node (destination) at (axis cs:22,0.00125){};
      \draw[->](source)--(destination);
%
%
\node[anchor=west] (source) at (axis cs:33.5,3.5e-03){\scriptsize{$n_s=200$}};
\node (destination) at (axis cs:30,0.0005){};
\draw[->](source)--(destination);

\coordinate (point) at (axis cs:13,4.9e-03);
\coordinate (spy point) at (axis cs:22,8e-03);
\spy on (point) in node [fill=white]  at (spy point);

\end{axis}
\end{tikzpicture}
\end{adjustbox}
\caption{$d=3$}
\label{fig:obtlntb}
\end{subfigure}
\begin{subfigure}[b]{0.33\textwidth}
\begin{adjustbox}{width=\linewidth}
\begin{tikzpicture}[
        spy using outlines={
            circle,
            magnification=3,
            connect spies,
            size=1cm,
            black,
        },
    ]
\begin{axis}[width=1\linewidth,font=\footnotesize,cycle list name=colors_obtl,
    xlabel= $n_t$, ylabel= MSE,
  xmin=5,xmax=50,xtick={5,10,20,30,40,50},ylabel near ticks,scaled y ticks=base 10:3,ymin=0,ymax=14e-03,
    xlabel near ticks,
    legend style={nodes=right},legend style={nodes={scale=1, transform shape}},
      xmajorgrids,
    grid style={dotted},
    ymajorgrids,
   ]
  \addplot plot coordinates{
(5,0.013317)(10,0.006903)(20,0.003609)(30,0.002448)(40,0.001845)(50,0.001488)
};
 \addplot plot coordinates{
(5,0.012686)(10,0.006373)(20,0.003225)(30,0.002153)(40,0.001617)(50,0.001303)
};
 \addplot plot coordinates{
(5,0.012686)(10,0.006148)(20,0.003041)(30,0.002021)(40,0.001515)(50,0.001219)
};
 \addplot plot coordinates{
(5,0.011904)(10,0.005866)(20,0.002928)(30,0.001957)(40,0.001469)(50,0.001183)
};
 \addplot plot coordinates{
(5,0.003128)(10,0.002418)(20,0.001857)(30,0.001407)(40,0.001176)(50,0.001153)
};
 \addplot  plot coordinates{
(5,0.002301)(10,0.001733)(20,0.001291)(30,0.000943)(40,0.000775)(50,0.000773)
};

\node[anchor=west] (source) at (axis cs:35,7e-03){\scriptsize{$n_s=50$}};
       \node (destination) at (axis cs:22,0.00125){};
       \draw[->](source)--(destination);
%
%
\node[anchor=west] (source) at (axis cs:33.5,3.5e-03){\scriptsize{$n_s=200$}};
\node (destination) at (axis cs:30,0.0005){};
\draw[->](source)--(destination);

\coordinate (point) at (axis cs:13.5,4.9e-03);
\coordinate (spy point) at (axis cs:22.2,8e-03);
\spy on (point) in node [fill=white]  at (spy point);

\end{axis}
\end{tikzpicture}
\end{adjustbox}
\caption{$d=5$}
\label{fig:obtlntc}
\end{subfigure}

\caption{MSE deviation from true error with respect to target data size. The proposed TL-based BEE is compared with other widely used estimators. In all figures, the Bayes error is fixed at $0.2$, and $\left|\alpha\right|=0.9$.}
\label{fig:obtlnt}
\end{center}
\vspace{-0.3in}
\end{figure}

%% file: figures/QDA/scz_alpha.tex
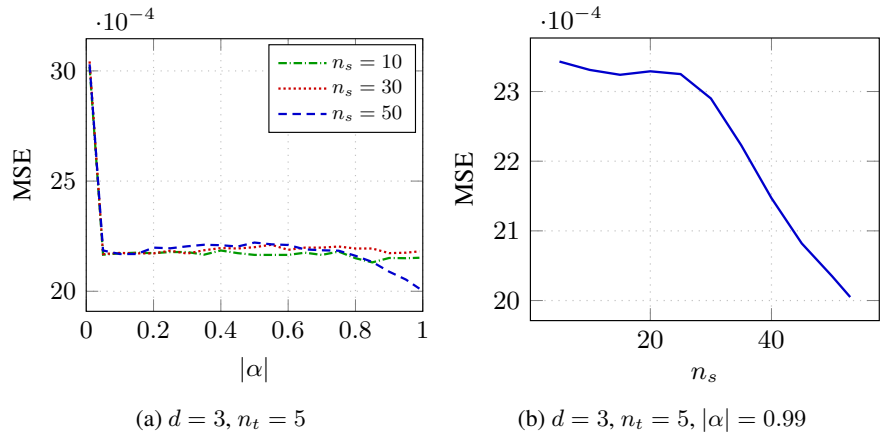
\begin{figure}[h!]
\begin{center}
\begin{subfigure}[b]{0.35\textwidth}
\begin{adjustbox}{width=\linewidth}
\begin{tikzpicture}
    \begin{axis}[width=1\linewidth,font=\footnotesize,cycle list name=colors_scz,
    xlabel= $\left|{\alpha}\right|$, ylabel= MSE,ylabel near ticks,xmin=0,xmax=1,xtick={0,0.2,0.4,0.6,0.8,1},
    xlabel near ticks,
  scaled y ticks=base 10:4,
    legend style={nodes=right},legend style={nodes={scale=0.8, transform shape}},
      xmajorgrids,
    grid style={dotted},
    ymajorgrids,
   ]

  \addplot plot coordinates {
(1.000000e-02,0.003018)(5.000000e-02,0.002166)(1.000000e-01,0.002172)(1.500000e-01,0.002175)(2.000000e-01,0.002173)(2.500000e-01,0.002178)(3.000000e-01,0.002178)(3.500000e-01,0.002166)(4.000000e-01,0.002185)(4.500000e-01,0.002173)(5.000000e-01,0.002165)(5.500000e-01,0.002165)(6.000000e-01,0.002165)(6.500000e-01,0.002175)(7.000000e-01,0.002165)(7.500000e-01,0.002180)(8.000000e-01,0.002150)(8.500000e-01,0.002131)(9.000000e-01,0.002151)(9.500000e-01,0.002150)(9.900000e-01,0.002152)
};
\addplot plot coordinates {
(1.000000e-02,0.003043)(5.000000e-02,0.002169)(1.000000e-01,0.002174)(1.500000e-01,0.002172)(2.000000e-01,0.002172)(2.500000e-01,0.002184)(3.000000e-01,0.002171)(3.500000e-01,0.002186)(4.000000e-01,0.002196)(4.500000e-01,0.002194)(5.000000e-01,0.002200)(5.500000e-01,0.002211)(6.000000e-01,0.002188)(6.500000e-01,0.002198)(7.000000e-01,0.002198)(7.500000e-01,0.002203)(8.000000e-01,0.002194)(8.500000e-01,0.002194)(9.000000e-01,0.002173)(9.500000e-01,0.002175)(9.900000e-01,0.002181)
};
     \addplot plot coordinates {
(1.000000e-02,0.003030)(5.000000e-02,0.002185)(1.000000e-01,0.002169)(1.500000e-01,0.002169)(2.000000e-01,0.002198)(2.500000e-01,0.002195)(3.000000e-01,0.002204)(3.500000e-01,0.002211)(4.000000e-01,0.002209)(4.500000e-01,0.002203)(5.000000e-01,0.002221)(5.500000e-01,0.002213)(6.000000e-01,0.002210)(6.500000e-01,0.002190)(7.000000e-01,0.002186)(7.500000e-01,0.002184)(8.000000e-01,0.002161)(8.500000e-01,0.002133)(9.000000e-01,0.002089)(9.500000e-01,0.002053)(9.900000e-01,0.002012)
};
\legend{$n_s=10$,$n_s=30$,$n_s=50$};
\end{axis}
\end{tikzpicture}
\end{adjustbox}
\caption{$d=3$, $n_t=5$}
\label{fig:sczalpha}
\end{subfigure}
\begin{subfigure}[b]{0.35\textwidth}
\begin{adjustbox}{width=\linewidth}
\begin{tikzpicture}
    \begin{axis}[ width=1\linewidth,font=\footnotesize,
    xlabel= $n_s$, ylabel= MSE,ylabel near ticks,
    xlabel near ticks,
    scaled y ticks=base 10:4,
      xmajorgrids,
    grid style={dotted},
    ymajorgrids,
   ]

  \addplot [blue!80!black,solid, thick]plot coordinates {
(5,0.002343)(10,0.002331)(15,0.002324)(20,0.002329)(25,0.002325)(30,0.002290)(35,0.002223)(40,0.002147)(45,0.002082)(50,0.002035)(53,0.002005)
};

\end{axis}
\end{tikzpicture}
\end{adjustbox}
\caption{$d=3$, $n_t=5$, $\left|{\alpha}\right|=0.99$}
\label{fig:sczalphans}
\end{subfigure}

\caption{MSE deviation from QDA true error for normally distributed brain gene expression data with respect to $\left|\alpha\right|$ and $n_{s}$. \textbf{a}. Gene features from the FC brain region demonstrate high relatedness with those from DPLFC area ($\left|\alpha\right|=0.99$). \textbf{b}. Utilizing the data from source domain significantly reduces the MSE of the TL-based BEE in the target domain.}
\end{center}
\end{figure}

%% file: figures/hg.tex
\begin{figure}[ht!]
\begin{center}
\begin{subfigure}[b]{0.35\textwidth}
\begin{adjustbox}{width=.85\linewidth}
\begin{tikzpicture}
    \begin{axis}[font=\footnotesize,ymode=log,
    xlabel= $\tau$, ylabel= Function Value,ylabel near ticks,
    xlabel near ticks,xmin=0,xmax=0.8,ymin=1,ytick={1,10},xtick={0,0.2,0.4,0.6,0.8},
    legend style={nodes=right},legend style={nodes={scale=1, transform shape}}, legend pos= north west,
      xmajorgrids,
    grid style={dotted},
    ymajorgrids,
   ]
  \addplot [smooth,blue,thick,mark=o,mark options={solid},every mark/.append style={solid, fill=white},densely dashed] plot coordinates {
(0.01,1.038216)(0.10,1.455577)(0.20,2.120418)(0.30,3.091436)(0.40,4.510798)(0.50,6.587197)(0.60,9.627127)(0.70,14.080649)(0.80,20.616102)
};
\addplot [smooth, thick,mark=star,red,only marks]plot coordinates {
(0.01,1.038215)(0.10,1.455435)(0.20,2.119685)(0.30,3.089331)(0.40,4.506094)(0.50,6.578162)(0.60,9.611720)(0.70,14.057622)(0.80,20.580530)
};
\legend{Exact confluent HG function,Laplace approximation};
\end{axis}
\end{tikzpicture}
\end{adjustbox}
\caption{$d=5$, $a=3$, $b=4$.}
\label{fig:1F1tau}
\end{subfigure}
\begin{subfigure}[b]{0.35\textwidth}
\begin{adjustbox}{width=.85\linewidth}
\begin{tikzpicture}
    \begin{axis}[ font=\footnotesize,ymode=log,
    xlabel= $\tau$, xmin=0,xmax=0.8,ymin=1,ymax=1e07,ylabel= Function Value,ylabel near ticks,xtick={0,0.2,0.4,0.6,0.8},
    xlabel near ticks,
    legend style={nodes=right},legend style={nodes={scale=1, transform shape}},legend pos= north west,
      xmajorgrids,
    grid style={dotted},
    ymajorgrids,
   ]
  \addplot [smooth,blue,thick,mark=o,mark options={solid},every mark/.append style={solid, fill=white},densely dashed] plot coordinates {
(0.01,1.105677)(0.10,2.853537)(0.20,9.100906)(0.30,33.414575)(0.40,146.855970)(0.50,823.829511)(0.60,6529.758470)(0.70,88258.605218)(0.80,3090062.939003)
};
\addplot [smooth, thick,mark=star,red,only marks]plot coordinates {
(0.01,1.105681)(0.10,2.854737)(0.20,9.121994)(0.30,33.573664)(0.40,148.342927)(0.50,838.204441)(0.60,6720.837600)(0.70,92521.186192)(0.80,3308710.172375)
};
\legend{Exact Gauss HG function,Laplace approximation};
\end{axis}
\end{tikzpicture}
\end{adjustbox}
\caption{$d=5$, $a=3$, $b=4$, $c=6$.}
\label{fig:2F1tau}
\end{subfigure}
\par\medskip 
\begin{subfigure}[b]{0.35\textwidth}
\begin{adjustbox}{width=.85\linewidth}
\begin{tikzpicture}
 \begin{axis}[ font=\footnotesize,ymode=log,
 scaled ticks=false,ytick={1.05,1.1,1.15,1.2,1.25,1.3},yticklabels={1.05,1.1,1.15,1.2,1.25,1.3},
    xlabel= $b$, xmin=10,xmax=50,ymin=1.05,ylabel= Function Value,ylabel near ticks,  
    scaled y ticks = false,
    xlabel near ticks,
    legend style={nodes=right},legend style={nodes={scale=1, transform shape}},
      xmajorgrids,
    grid style={dotted},
    ymajorgrids,
   ]
  \addplot [smooth,blue,thick,mark=o,mark options={solid},every mark/.append style={solid, fill=white},densely dashed] plot coordinates {
(10,1.349665)(14,1.238909)(16,1.206189)(18,1.181335)(22,1.146089)(24,1.133143)(26,1.122302)(28,1.113092)(32,1.098286)(34,1.092246)(36,1.086906)(40,1.077886)(44,1.070562)(46,1.067393)(48,1.064497)(50,1.061839)
};
\addplot [smooth, thick,mark=star,red,only marks]plot coordinates {
(10,1.349676)(14,1.238911)(16,1.206190)(18,1.181335)(22,1.146089)(24,1.133143)(26,1.122302)(28,1.113092)(32,1.098286)(34,1.092246)(36,1.086906)(40,1.077886)(44,1.070562)(46,1.067393)(48,1.064497)(50,1.061839)
};
\legend{Exact confluent HG function,Laplace approximation};
\end{axis}
\end{tikzpicture}
\end{adjustbox}
\caption{$d=10$, $a=30$, $\tau=0.01$.}
\label{fig:1F1b}
\end{subfigure}
\begin{subfigure}[b]{0.35\textwidth}
\begin{adjustbox}{width=.85\linewidth}
\begin{tikzpicture}
 \begin{axis}[ font=\footnotesize,ymode=log,
    xlabel= $c$,xmin=10,xmax=50,ymin=10, ylabel= Function Value,ylabel near ticks,
    xlabel near ticks,
    legend style={nodes=right},legend style={nodes={scale=1, transform shape}},
      xmajorgrids,
    grid style={dotted},
    ymajorgrids,
   ]
  \addplot [smooth,blue,thick,mark=o,mark options={solid},every mark/.append style={solid, fill=white},densely dashed] plot coordinates {
(10,2641842.965813)(14,43093.276379)(16,11657.441384)(18,4190.325648)(22,935.017369)(24,531.062619)(26,328.656250)(28,217.650254)(32,111.250430)(34,84.352589)(36,65.943318)(40,43.375611)(44,30.781226)(46,26.515311)(48,23.125791)(50,20.384023)
};
\addplot [smooth, thick,mark=star,red,only marks]plot coordinates {
(10,2601772.429410)(14,43078.774929)(16,11667.009952)(18,4191.637248)(22,935.082292)(24,531.080434)(26,328.661122)(28,217.651351)(32,111.250193)(34,84.352341)(36,65.943121)(40,43.375516)(44,30.781190)(46,26.515291)(48,23.125783)(50,20.391145)
};
\legend{Exact Gauss HG function,Laplace approximation};
\end{axis}
\end{tikzpicture}
\end{adjustbox}
\caption{$d=10$, $a=30$, $b=50$, $\tau=0.01$.}
\label{fig:2F1c}
\end{subfigure}
\caption{Laplace approximations and exact evaluations of confluent and Gauss hypergeometric functions of matrix argument}
\label{fig:hypergeoFvalues}
\end{center}
\end{figure}
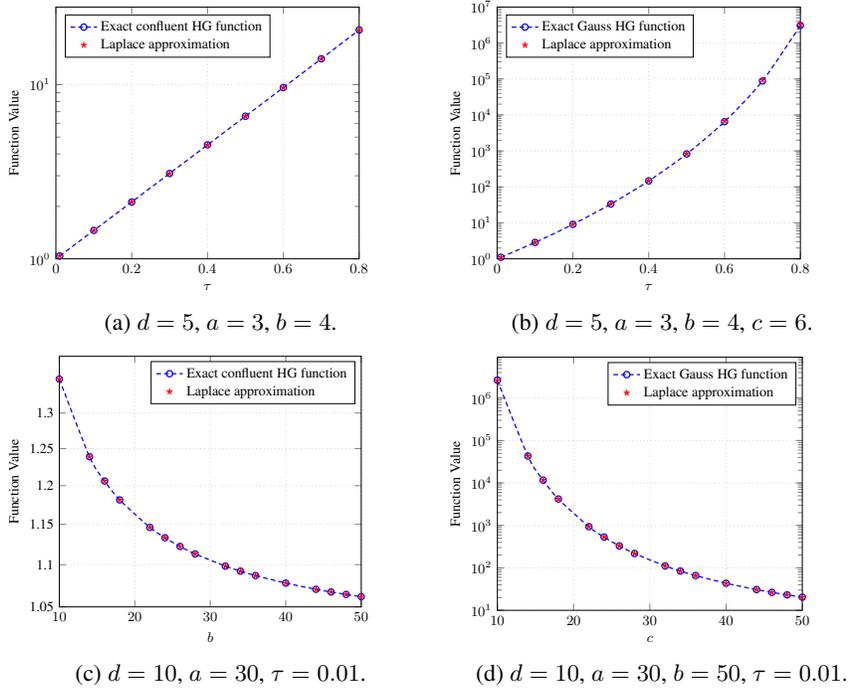

%% file: figures/LDA/LDA_n_s.tex
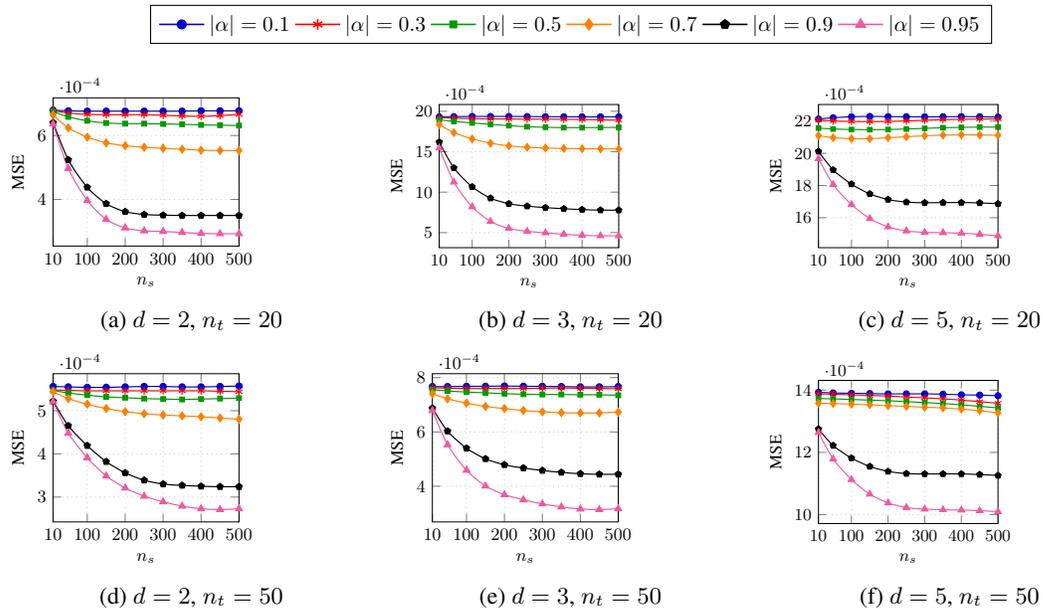
\begin{figure*}[h!]
\begin{center}
\ref{named5}
\par
\bigskip
\begin{subfigure}[b]{0.3\textwidth}
\begin{adjustbox}{width=0.68\linewidth}
\begin{tikzpicture}
    \begin{axis}[ width=1\linewidth,font=\footnotesize,cycle list name=colors_qda,
    xlabel= $n_s$, ylabel= MSE,ylabel near ticks,
    xlabel near ticks,
  xmin=10,xmax=500,xtick={10,100,200,300,400,500},
legend columns=-1,
legend entries={$\left|\alpha\right|=0.1$,$\left|\alpha\right|=0.3$,$\left|\alpha\right|=0.5$,$\left|\alpha\right|=0.7$,$\left|\alpha\right|=0.9$,$\left|\alpha\right|=0.95$},
legend to name=named5,
    legend style={nodes=right},legend style={nodes={scale=0.8, transform shape}},
      xmajorgrids,
    grid style={dotted},
    ymajorgrids,
   ]

  \addplot plot coordinates {
(10,0.000679)(50,0.000677)(100,0.000676)(150,0.000676)(200,0.000676)(250,0.000676)(300,0.000676)(350,0.000676)(400,0.000677)(450,0.000677)(500,0.000677)
};
\addplot plot coordinates {
(10,0.000678)(50,0.000670)(100,0.000666)(150,0.000665)(200,0.000665)(250,0.000665)(300,0.000663)(350,0.000661)(400,0.000660)(450,0.000662)(500,0.000666)
};
     \addplot plot coordinates {
(10,0.000675)(50,0.000658)(100,0.000646)(150,0.000639)(200,0.000637)(250,0.000637)(300,0.000636)(350,0.000635)(400,0.000633)(450,0.000632)(500,0.000631)
};
    \addplot plot coordinates {
(10,0.000664)(50,0.000624)(100,0.000595)(150,0.000577)(200,0.000568)(250,0.000563)(300,0.000560)(350,0.000557)(400,0.000554)(450,0.000553)(500,0.000553)
};
     \addplot plot coordinates {
(10,0.000640)(50,0.000524)(100,0.000438)(150,0.000387)(200,0.000362)(250,0.000353)(300,0.000351)(350,0.000350)(400,0.000350)(450,0.000350)(500,0.000350)
};
      \addplot  plot coordinates {
(10,0.000636)(50,0.000497)(100,0.000397)(150,0.000338)(200,0.000311)(250,0.000302)(300,0.000300)(350,0.000297)(400,0.000294)(450,0.000293)(500,0.000293)
};
\end{axis}
\end{tikzpicture}
\end{adjustbox}
\caption{$d=2$, $n_t=20$}
\label{fig:ldansa}
\end{subfigure}
\begin{subfigure}[b]{0.3\textwidth}
\begin{adjustbox}{width=0.68\linewidth}
\begin{tikzpicture}
    \begin{axis}[ width=1\linewidth,font=\footnotesize,cycle list name=colors_qda,
    xlabel= $n_s$, ylabel= MSE,ylabel near ticks,scaled y ticks=base 10:4,
    xlabel near ticks,
  xmin=10,xmax=500,xtick={10,100,200,300,400,500},
      xmajorgrids,
    grid style={dotted},
    ymajorgrids,
   ]

  \addplot plot coordinates {
(10,0.001931)(50,0.001934)(100,0.001936)(150,0.001936)(200,0.001936)(250,0.001935)(300,0.001933)(350,0.001932)(400,0.001931)(450,0.001931)(500,0.001932)
};
\addplot plot coordinates {
(10,0.001922)(50,0.001915)(100,0.001909)(150,0.001906)(200,0.001904)(250,0.001903)(300,0.001902)(350,0.001901)(400,0.001898)(450,0.001894)(500,0.001889)
};
     \addplot plot coordinates {
(10,0.001895)(50,0.001875)(100,0.001856)(150,0.001838)(200,0.001822)(250,0.001809)(300,0.001802)(350,0.001798)(400,0.001798)(450,0.001799)(500,0.001801)
};
    \addplot plot coordinates {
(10,0.001836)(50,0.001736)(100,0.001658)(150,0.001605)(200,0.001573)(250,0.001555)(300,0.001546)(350,0.001540)(400,0.001537)(450,0.001535)(500,0.001534)
};
     \addplot plot coordinates {
(10,0.001619)(50,0.001299)(100,0.001066)(150,0.000926)(200,0.000857)(250,0.000827)(300,0.000809)(350,0.000795)(400,0.000784)(450,0.000778)(500,0.000777)
};
      \addplot  plot coordinates {
(10,0.001546)(50,0.001125)(100,0.000819)(150,0.000638)(200,0.000552)(250,0.000516)(300,0.000497)(350,0.000480)(400,0.000467)(450,0.000461)(500,0.000461)
};
\end{axis}
\end{tikzpicture}
\end{adjustbox}
\caption{$d=3$, $n_t=20$}
\label{fig:ldansb}
\end{subfigure}
\begin{subfigure}[b]{0.3\textwidth}
\begin{adjustbox}{width=0.68\linewidth}
\begin{tikzpicture}
\begin{axis}[ width=1\linewidth,font=\footnotesize,cycle list name=colors_qda,
    xlabel= $n_s$, ylabel= MSE,
  xmin=10,xmax=500,xtick={10,100,200,300,400,500},ylabel near ticks,scaled y ticks=base 10:4,
    xlabel near ticks,
    legend style={nodes=right},legend style={nodes={scale=1, transform shape}},
      xmajorgrids,
    grid style={dotted},
    ymajorgrids,
   ]
  \addplot plot coordinates {
(10,0.002213)(50,0.002221)(100,0.002227)(150,0.002229)(200,0.002228)(250,0.002227)(300,0.002227)(350,0.002227)(400,0.002228)(450,0.002227)(500,0.002225)
};
  \addplot plot coordinates {
(10,0.002207)(50,0.002201)(100,0.002198)(150,0.002197)(200,0.002198)(250,0.002201)(300,0.002205)(350,0.002209)(400,0.002211)(450,0.002213)(500,0.002213)
};
  \addplot plot coordinates {
(10,0.002157)(50,0.002151)(100,0.002148)(150,0.002147)(200,0.002148)(250,0.002151)(300,0.002155)(350,0.002159)(400,0.002161)(450,0.002163)(500,0.002163)
};
  \addplot plot coordinates {
(10,0.002110)(50,0.002097)(100,0.002091)(150,0.002091)(200,0.002097)(250,0.002103)(300,0.002109)(350,0.002113)(400,0.002115)(450,0.002114)(500,0.002112)
};
  \addplot plot coordinates {
(10,0.002012)(50,0.001897)(100,0.001809)(150,0.001748)(200,0.001713)(250,0.001697)(300,0.001693)(350,0.001694)(400,0.001694)(450,0.001692)(500,0.001687)
};
  \addplot plot coordinates {
(10,0.001967)(50,0.001806)(100,0.001682)(150,0.001595)(200,0.001543)(250,0.001518)(300,0.001509)(350,0.001507)(400,0.001504)(450,0.001497)(500,0.001488)
};
\end{axis}
\end{tikzpicture}
\end{adjustbox}
\caption{$d=5$, $n_t=20$}
\label{fig:ldansc}
\end{subfigure}
\par\medskip 
\begin{subfigure}[b]{0.3\textwidth}
\begin{adjustbox}{width=0.68\linewidth}
\begin{tikzpicture}
\begin{axis}[width=1\linewidth,font=\footnotesize,cycle list name=colors_qda,
    xlabel= $n_s$, ylabel= MSE,
  xmin=10,xmax=500,xtick={10,100,200,300,400,500},ylabel near ticks,
    xlabel near ticks,
    legend style={nodes=right},legend style={nodes={scale=1, transform shape}},
      xmajorgrids,
    grid style={dotted},
    ymajorgrids,
   ]
  \addplot plot coordinates{
(10,0.000556)(50,0.000555)(100,0.000554)(150,0.000554)(200,0.000555)(250,0.000556)(300,0.000556)(350,0.000555)(400,0.000555)(450,0.000556)(500,0.000557)
};
 \addplot plot coordinates{
(10,0.000548)(50,0.000547)(100,0.000546)(150,0.000546)(200,0.000546)(250,0.000546)(300,0.000546)(350,0.000546)(400,0.000546)(450,0.000545)(500,0.000544)
};
 \addplot plot coordinates{
(10,0.000547)(50,0.000541)(100,0.000536)(150,0.000532)(200,0.000530)(250,0.000528)(300,0.000527)(350,0.000526)(400,0.000527)(450,0.000528)(500,0.000529)
};
 \addplot plot coordinates{
(10,0.000544)(50,0.000528)(100,0.000515)(150,0.000505)(200,0.000498)(250,0.000493)(300,0.000490)(350,0.000488)(400,0.000486)(450,0.000483)(500,0.000480)
};
 \addplot plot coordinates{
(10,0.000522)(50,0.000465)(100,0.000419)(150,0.000382)(200,0.000356)(250,0.000339)(300,0.000330)(350,0.000327)(400,0.000325)(450,0.000324)(500,0.000324)
};
 \addplot plot coordinates{
(10,0.000521)(50,0.000448)(100,0.000391)(150,0.000349)(200,0.000321)(250,0.000302)(300,0.000289)(350,0.000279)(400,0.000273)(450,0.000271)(500,0.000273)
};
\end{axis}
\end{tikzpicture}
\end{adjustbox}
\caption{$d=2$, $n_t=50$}
\label{fig:ldansd}
\end{subfigure}
\begin{subfigure}[b]{0.3\textwidth}
\begin{adjustbox}{width=0.68\linewidth}
\begin{tikzpicture}
\begin{axis}[width=1\linewidth,font=\footnotesize,cycle list name=colors_qda,
    xlabel= $n_s$, ylabel= MSE,
  xmin=10,xmax=500,xtick={10,100,200,300,400,500},ylabel near ticks,
    xlabel near ticks,
legend style={nodes=right},legend style={nodes={scale=1, transform shape}},
      xmajorgrids,
    grid style={dotted},
    ymajorgrids,
   ]
  \addplot plot coordinates{
(10,0.000767)(50,0.000767)(100,0.000768)(150,0.000768)(200,0.000769)(250,0.000768)(300,0.000768)(350,0.000767)(400,0.000766)(450,0.000766)(500,0.000767)
};
 \addplot plot coordinates{
(10,0.000763)(50,0.000761)(100,0.000760)(150,0.000760)(200,0.000760)(250,0.000760)(300,0.000761)(350,0.000762)(400,0.000761)(450,0.000760)(500,0.000759)
};
 \addplot plot coordinates{
(10,0.000756)(50,0.000751)(100,0.000747)(150,0.000744)(200,0.000741)(250,0.000739)(300,0.000738)(350,0.000738)(400,0.000737)(450,0.000737)(500,0.000735)
};
 \addplot plot coordinates{
(10,0.000741)(50,0.000722)(100,0.000706)(150,0.000694)(200,0.000686)(250,0.000679)(300,0.000674)(350,0.000671)(400,0.000670)(450,0.000670)(500,0.000673)
};
 \addplot plot coordinates{
(10,0.000687)(50,0.000603)(100,0.000540)(150,0.000501)(200,0.000479)(250,0.000467)(300,0.000458)(350,0.000451)(400,0.000446)(450,0.000444)(500,0.000444)
};
 \addplot plot coordinates{
(10,0.000679)(50,0.000553)(100,0.000459)(150,0.000401)(200,0.000369)(250,0.000350)(300,0.000335)(350,0.000324)(400,0.000316)(450,0.000314)(500,0.000317)
};
\end{axis}
\end{tikzpicture}
\end{adjustbox}
\caption{$d=3$, $n_t=50$}
\label{fig:ldanse}
\end{subfigure}
\begin{subfigure}[b]{0.3\textwidth}
\begin{adjustbox}{width=0.68\linewidth}
\begin{tikzpicture}
\begin{axis}[width=1\linewidth, font=\footnotesize,cycle list name=colors_qda,
    xlabel= $n_s$, ylabel= MSE,
  xmin=10,xmax=500,xtick={10,100,200,300,400,500},ylabel near ticks,scaled y ticks=base 10:4,
    xlabel near ticks,
    legend style={nodes=right},legend style={nodes={scale=1, transform shape}},
      xmajorgrids,
    grid style={dotted},
    ymajorgrids,
   ]
 \addplot plot coordinates{
(10,0.001393)(50,0.001391)(100,0.001390)(150,0.001389)(200,0.001388)(250,0.001388)(300,0.001388)(350,0.001387)(400,0.001385)(450,0.001384)(500,0.001382)
};
 \addplot plot coordinates{
(10,0.001388)(50,0.001387)(100,0.001385)(150,0.001383)(200,0.001381)(250,0.001378)(300,0.001375)(350,0.001372)(400,0.001368)(450,0.001363)(500,0.001358)
};
 \addplot plot coordinates{
(10,0.001373)(50,0.001372)(100,0.001370)(150,0.001368)(200,0.001366)(250,0.001363)(300,0.001360)(350,0.001357)(400,0.001353)(450,0.001348)(500,0.001343)
};
 \addplot plot coordinates{
(10,0.001358)(50,0.001357)(100,0.001355)(150,0.001353)(200,0.001350)(250,0.001348)(300,0.001345)(350,0.001343)(400,0.001339)(450,0.001334)(500,0.001327)
};
 \addplot plot coordinates{
(10,0.001275)(50,0.001222)(100,0.001182)(150,0.001155)(200,0.001139)(250,0.001132)(300,0.001131)(350,0.001131)(400,0.001131)(450,0.001129)(500,0.001126)
};
 \addplot plot coordinates{
(10,0.001264)(50,0.001179)(100,0.001113)(150,0.001066)(200,0.001038)(250,0.001023)(300,0.001018)(350,0.001016)(400,0.001015)(450,0.001013)(500,0.001010)};
\end{axis}
\end{tikzpicture}
\end{adjustbox}
\caption{$d=5$, $n_t=50$}
\label{fig:ldansf}
\end{subfigure}

\caption{MSE deviation from true error for Gaussian distributions with respect to source sample size. The Bayes error is fixed at 0.2 in all figures.}
\label{fig:ldans}
\end{center}
\end{figure*}

%% file: figures/LDA/LDA_n_t.tex
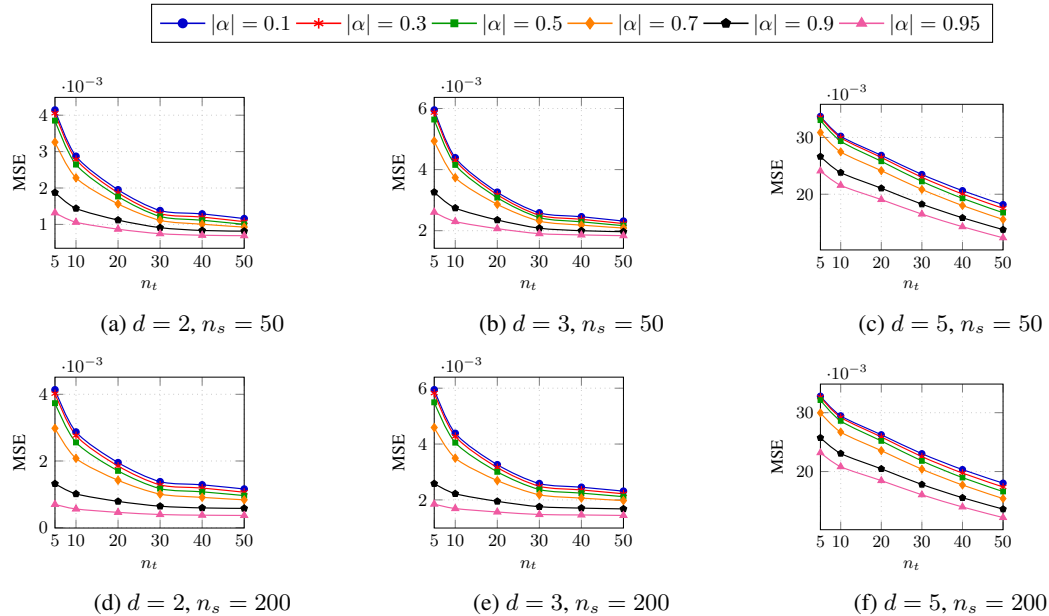
\begin{figure*}[h!]
\begin{center}
\ref{named7}
\par
\bigskip
\begin{subfigure}[b]{0.3\textwidth}
\begin{adjustbox}{width=0.68\linewidth}
\begin{tikzpicture}
    \begin{axis}[ width=1\linewidth,font=\footnotesize,cycle list name=colors_qda,
    xlabel= $n_t$, ylabel= MSE,ylabel near ticks,
    xlabel near ticks,
  xmin=5,xmax=50,xtick={5,10,20,30,40,50}, 
legend columns=-1,
legend entries={$\left|\alpha\right|=0.1$,$\left|\alpha\right|=0.3$,$\left|\alpha\right|=0.5$,$\left|\alpha\right|=0.7$,$\left|\alpha\right|=0.9$,$\left|\alpha\right|=0.95$},
legend to name=named7,
    legend style={nodes=right},legend style={nodes={scale=0.8, transform shape}},
      xmajorgrids,
    grid style={dotted},
    ymajorgrids,
   ]

  \addplot plot coordinates {
(5,0.004144)(10,0.002873)(20,0.001952)(30,0.001381)(40,0.001289)(50,0.001160)
};
\addplot plot coordinates {
(5,0.004056)(10,0.002787)(20,0.001867)(30,0.001296)(40,0.001202)(50,0.001075)
};
     \addplot plot coordinates {
(5,0.003851)(10,0.002644)(20,0.001766)(30,0.001219)(40,0.001115)(50,0.001002)
};
    \addplot plot coordinates {
(5,0.003260)(10,0.002279)(20,0.001563)(30,0.001112)(40,0.001005)(50,0.000926)
};
     \addplot plot coordinates {
(5,0.001875)(10,0.001440)(20,0.001119)(30,0.000911)(40,0.000833)(50,0.000815)
};
      \addplot  plot coordinates {
(5,0.001319)(10,0.001060)(20,0.000869)(30,0.000745)(40,0.000702)(50,0.000690)
};
\end{axis}
\end{tikzpicture}
\end{adjustbox}
\caption{$d=2$, $n_s=50$}
\label{fig:ldanta}
\end{subfigure}
\begin{subfigure}[b]{0.3\textwidth}
\begin{adjustbox}{width=0.68\linewidth}
\begin{tikzpicture}
    \begin{axis}[ width=1\linewidth,font=\footnotesize,cycle list name=colors_qda,
    xlabel= $n_t$, ylabel= MSE,ylabel near ticks,
    xlabel near ticks,
  xmin=5,xmax=50,xtick={5,10,20,30,40,50},
      xmajorgrids,
    grid style={dotted},
    ymajorgrids,
   ]

  \addplot plot coordinates {
(5,0.005952)(10,0.004388)(20,0.003261)(30,0.002585)(40,0.002457)(50,0.002314)
};
\addplot plot coordinates {
(5,0.005868)(10,0.004304)(20,0.003179)(30,0.002505)(40,0.002373)(50,0.002233)
};
     \addplot plot coordinates {
(5,0.005639)(10,0.004154)(20,0.003081)(30,0.002430)(40,0.002290)(50,0.002164)
};
    \addplot plot coordinates {
(5,0.004935)(10,0.003737)(20,0.002862)(30,0.002314)(40,0.002182)(50,0.002087)
};
     \addplot plot coordinates {
(5,0.003262)(10,0.002742)(20,0.002351)(30,0.002082)(40,0.001999)(50,0.001967)
};
      \addplot  plot coordinates {
(5,0.002606)(10,0.002300)(20,0.002067)(30,0.001903)(40,0.001861)(50,0.001836)
};
\end{axis}
\end{tikzpicture}
\end{adjustbox}
\caption{$d=3$, $n_s=50$}
\label{fig:ldantb}
\end{subfigure}
\begin{subfigure}[b]{0.3\textwidth}
\begin{adjustbox}{width=0.68\linewidth}
\begin{tikzpicture}
\begin{axis}[ width=1\linewidth,font=\footnotesize,cycle list name=colors_qda,
    xlabel= $n_t$, ylabel= MSE,
  xmin=5,xmax=50,xtick={5,10,20,30,40,50},ylabel near ticks,scaled y ticks=base 10:3,
    xlabel near ticks,
    legend style={nodes=right},legend style={nodes={scale=1, transform shape}},
      xmajorgrids,
    grid style={dotted},
    ymajorgrids,
   ]
  \addplot plot coordinates {
(5,0.03370)(10,0.03022)(20,0.02685)(30,0.02348)(40,0.02063)(50,0.01818)
};
  \addplot plot coordinates {
(5,0.03355)(10,0.02996)(20,0.02648)(30,0.02298)(40,0.02004)(50,0.01753)
};
  \addplot plot coordinates {
(5,0.03304)(10,0.02935)(20,0.02580)(30,0.02225)(40,0.01927)(50,0.01675)
};
  \addplot plot coordinates {
(5,0.03087)(10,0.02746)(20,0.02415)(30,0.02084)(40,0.01801)(50,0.01556)
};
  \addplot plot coordinates {
(5,0.02664)(10,0.02382)(20,0.02106)(30,0.01825)(40,0.01585)(50,0.01377)
};
  \addplot plot coordinates {
(5,0.02410)(10,0.02157)(20,0.01906)(30,0.01650)(40,0.01428)(50,0.01233)
};
\end{axis}
\end{tikzpicture}
\end{adjustbox}
\caption{$d=5$, $n_s=50$}
\label{fig:ldantc}
\end{subfigure}
\par\medskip 
\begin{subfigure}[b]{0.3\textwidth}
\begin{adjustbox}{width=0.68\linewidth}
\begin{tikzpicture}
\begin{axis}[width=1\linewidth,font=\footnotesize,cycle list name=colors_qda,
    xlabel= $n_t$, ylabel= MSE,
  xmin=5,xmax=50,xtick={5,10,20,30,40,50},ylabel near ticks,  
    xlabel near ticks,
    legend style={nodes=right},legend style={nodes={scale=1, transform shape}},
      xmajorgrids,
    grid style={dotted},
    ymajorgrids,
   ]
  \addplot plot coordinates{
(5,0.004137)(10,0.002870)(20,0.001952)(30,0.001381)(40,0.001286)(50,0.001160)
};
 \addplot plot coordinates{
(5,0.004022)(10,0.002763)(20,0.001850)(30,0.001284)(40,0.001189)(50,0.001063)
};
 \addplot plot coordinates{
(5,0.003734)(10,0.002557)(20,0.001704)(30,0.001172)(40,0.001076)(50,0.000963)
};
 \addplot plot coordinates{
(5,0.002980)(10,0.002079)(20,0.001422)(30,0.001007)(40,0.000909)(50,0.000836)
};
 \addplot plot coordinates{
(5,0.001319)(10,0.001014)(20,0.000789)(30,0.000644)(40,0.000593)(50,0.000579)
};
 \addplot plot coordinates{
(5,0.000699)(10,0.000561)(20,0.000460)(30,0.000395)(40,0.000375)(50,0.000367)
};
\end{axis}
\end{tikzpicture}
\end{adjustbox}
\caption{$d=2$, $n_s=200$}
\label{fig:ldantd}
\end{subfigure}
\begin{subfigure}[b]{0.3\textwidth}
\begin{adjustbox}{width=0.68\linewidth}
\begin{tikzpicture}
\begin{axis}[width=1\linewidth,font=\footnotesize,cycle list name=colors_qda,
    xlabel= $n_t$, ylabel= MSE,
  xmin=5,xmax=50,xtick={5,10,20,30,40,50},ylabel near ticks,
    xlabel near ticks,
legend style={nodes=right},legend style={nodes={scale=1, transform shape}},
      xmajorgrids,
    grid style={dotted},
    ymajorgrids,
   ]
  \addplot plot coordinates{
(5,0.005945)(10,0.004385)(20,0.003261)(30,0.002586)(40,0.002454)(50,0.002314)
};
 \addplot plot coordinates{
(5,0.005826)(10,0.004275)(20,0.003159)(30,0.002489)(40,0.002357)(50,0.002219)
};
 \addplot plot coordinates{
(5,0.005494)(10,0.004050)(20,0.003006)(30,0.002372)(40,0.002243)(50,0.002116)
};
 \addplot plot coordinates{
(5,0.004592)(10,0.003495)(20,0.002692)(30,0.002184)(40,0.002065)(50,0.001976)
};
 \addplot plot coordinates{
(5,0.002583)(10,0.002222)(20,0.001949)(30,0.001756)(40,0.001706)(50,0.001678)
};
 \addplot plot coordinates{
(5,0.001849)(10,0.001690)(20,0.001567)(30,0.001477)(40,0.001461)(50,0.001443)
};
\end{axis}
\end{tikzpicture}
\end{adjustbox}
\caption{$d=3$, $n_s=200$}
\label{fig:ldante}
\end{subfigure}
\begin{subfigure}[b]{0.3\textwidth}
\begin{adjustbox}{width=0.68\linewidth}
\begin{tikzpicture}
\begin{axis}[width=1\linewidth, font=\footnotesize,cycle list name=colors_qda,
    xlabel= $n_t$, ylabel= MSE,
   xmin=5,xmax=50,xtick={5,10,20,30,40,50},ylabel near ticks,scaled y ticks=base 10:3,
    xlabel near ticks,
    legend style={nodes=right},legend style={nodes={scale=1, transform shape}},
      xmajorgrids,
    grid style={dotted},
    ymajorgrids,
   ]
 \addplot plot coordinates{
(5,0.03280)(10,0.02947)(20,0.02625)(30,0.02303)(40,0.02033)(50,0.01803)
};
 \addplot plot coordinates{
(5,0.03265)(10,0.02921)(20,0.02588)(30,0.02253)(40,0.01974)(50,0.01738)
};
 \addplot plot coordinates{
(5,0.03214)(10,0.02860)(20,0.02520)(30,0.02180)(40,0.01897)(50,0.01660)
};
 \addplot plot coordinates{
(5,0.02997)(10,0.02671)(20,0.02355)(30,0.02039)(40,0.01771)(50,0.01541)
};
 \addplot plot coordinates{
(5,0.02574)(10,0.02307)(20,0.02046)(30,0.01780)(40,0.01555)(50,0.01362)
};
 \addplot plot coordinates{
(5,0.02320)(10,0.02082)(20,0.01846)(30,0.01605)(40,0.01398)(50,0.01218)};
\end{axis}
\end{tikzpicture}
\end{adjustbox}
\caption{$d=5$, $n_s=200$}
\label{fig:ldantf}
\end{subfigure}

\caption{MSE deviation from true error for Gaussian distributions with respect to target sample size. The Bayes error is fixed at 0.2 in all figures.}
\label{fig:ldant}
\end{center}
\end{figure*}